\documentclass{article}

\usepackage{arxiv}

\usepackage[utf8]{inputenc} 
\usepackage[T1]{fontenc}    
\usepackage{hyperref}       
\usepackage{url}            
\usepackage{booktabs}       
\usepackage{amsfonts}       
\usepackage{nicefrac}       
\usepackage{microtype}      
\usepackage{xcolor}         

\usepackage{amsmath}
\usepackage{amsthm}
\usepackage{algorithm}
\usepackage{algpseudocode}
\usepackage{graphicx}
\usepackage{subcaption}
\usepackage{bbm}

\newtheorem{theorem}{Theorem}

\newcommand\blfootnote[1]{%
  \begingroup
  \renewcommand\thefootnote{}\footnotetext{#1}%
  \endgroup
}

\title{Non-Asymptotic Uncertainty Quantification in High-Dimensional Learning}

\author{%
  Frederik Hoppe$^*$ \\
  RWTH Aachen University\\
  \texttt{hoppe@mathc.rwth-aachen.de} \\
  \And
  Claudio Mayrink Verdun$^*$ \\
  Harvard University \\
  \texttt{claudioverdun@seas.harvard.edu} \\
  \AND
  Hannah Laus$^*$ \\
  TU Munich $\&$ MCML\\
  \texttt{hannah.laus@tum.de}
  \And
  Felix Krahmer \\
  TU Munich $\&$ MCML\\
  \texttt{felix.krahmer@tum.de}
  \And
  Holger Rauhut \\
  LMU Munich $\&$ MCML\\
  \texttt{rauhut@math.lmu.de}
}

\begin{document}
\blfootnote{$*$: Equal contribution. Correspondence to \texttt{hoppe@mathc.rwth-aachen.de}.}

\maketitle

\begin{abstract}

Uncertainty quantification (UQ) is a crucial but challenging task in many high-dimensional regression or learning problems to increase the confidence of a given predictor. We develop a new data-driven approach for UQ in regression that applies both to classical regression approaches such as the LASSO as well as to neural networks. One of the most notable UQ techniques is the debiased LASSO, which modifies the LASSO to allow for the construction of asymptotic confidence intervals by decomposing the estimation error into a Gaussian and an asymptotically vanishing bias component. However, in real-world problems with finite-dimensional data, the bias term is often too significant to be neglected, resulting in overly narrow confidence intervals. Our work rigorously addresses this issue and derives a data-driven adjustment that corrects the confidence intervals for a large class of predictors by estimating the means and variances of the bias terms from training data, exploiting high-dimensional concentration phenomena. This gives rise to non-asymptotic confidence intervals, which can help avoid overestimating uncertainty in critical applications such as MRI diagnosis. Importantly, our analysis extends beyond sparse regression to data-driven predictors like neural networks, enhancing the reliability of model-based deep learning. Our findings bridge the gap between established theory and the practical applicability of such debiased methods.
\end{abstract}

\section{Introduction}

The past few years have witnessed remarkable advances in high-dimensional statistical models, inverse problems, and learning methods for solving them. In particular, we have seen a surge of new methodologies and algorithms that have revolutionized our ability to extract insights from complex, high-dimensional data \cite{wainwright2019high,giraud2021introduction,wright2022high}. Also, the theoretical underpinnings of the techniques in these fields have achieved tremendous success. However, the development of rigorous methods for quantifying uncertainty associated with their estimates, such as constructing confidence intervals for a given solution, has lagged behind, with much of the underlying theory remaining elusive.

In high-dimensional statistics, for example, even for classical regularized estimators such as the LASSO \cite{chen1995examples,tibshirani1996regression,hastie2009elements}, it was shown that a closed-form characterization of the probability distribution of the estimator in simple terms is not possible, e.g., \cite[Theorem 5.1]{Fu2000}. This, in turn, implies that it is very challenging to establish rigorous confidence intervals that would quantify the uncertainty of such estimated parameters. To overcome this, a series of papers \cite{Zhang.2014,Javanmard.2014,vandeGeer.2014} proposed and analyzed the \emph{debiased LASSO}, also known as the desparsified LASSO, a procedure to fix the bias introduced by the $\ell_1$ penalty in the LASSO; see \cite[Corollary 11]{Javanmard.2014} and  \cite{zhang2008sparsity} for a discussion on the bias induced by the $\ell_1$ regularizer. The debiased estimator derived in the aforementioned works has established a principled framework for obtaining sharp confidence intervals for the LASSO, initiating a statistical inference approach with UQ guarantees for high-dimensional regression problems where the number of predictors significantly exceeds the number of observations. Recently, this estimator was also extended in several directions beyond $\ell_1$-minimization which include, for example, deep unrolled algorithms \cite{hoppe2023uncertainty,bellec2024uncertainty} and it has been applied to fields like magnetic resonance image both with high-dimensional regression techniques as well as learning ones \cite{hoppe2023high,hoppe2023uncertainty}; see the paragraph \emph{related works} below. 

\begin{figure}
    \centering
    \begin{subfigure}[T]{0.19\linewidth}
        \includegraphics[width=\textwidth]{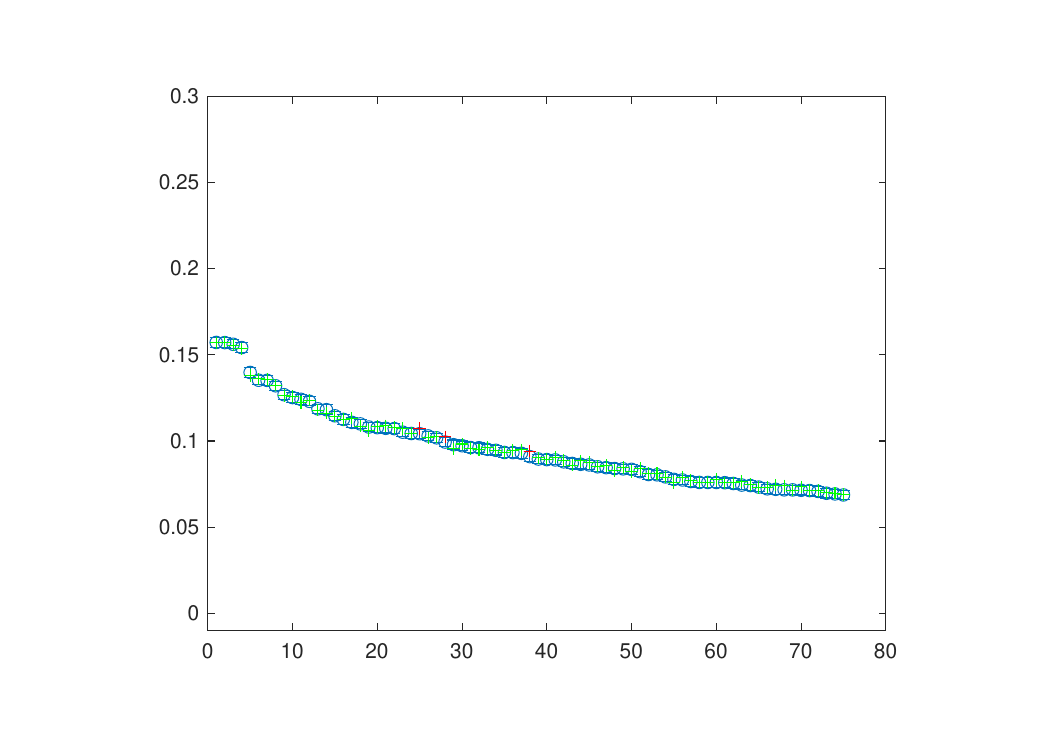}
        \caption{}
        \label{subfig:CI_comp_old}
    \end{subfigure}
    \begin{subfigure}[T]{0.19\linewidth}
        \includegraphics[width=\textwidth]{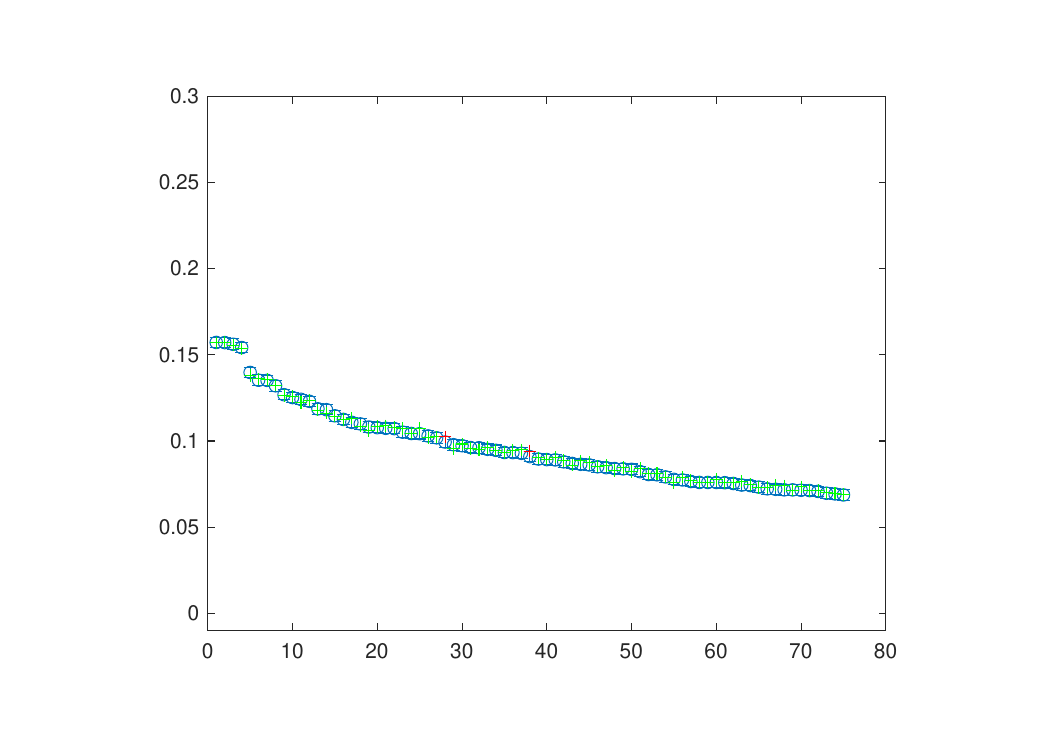}
        \caption{}
        \label{subfig:CI_comp_gauss}
    \end{subfigure}
    \begin{subfigure}[T]{0.19\linewidth}
        \includegraphics[width=\textwidth]{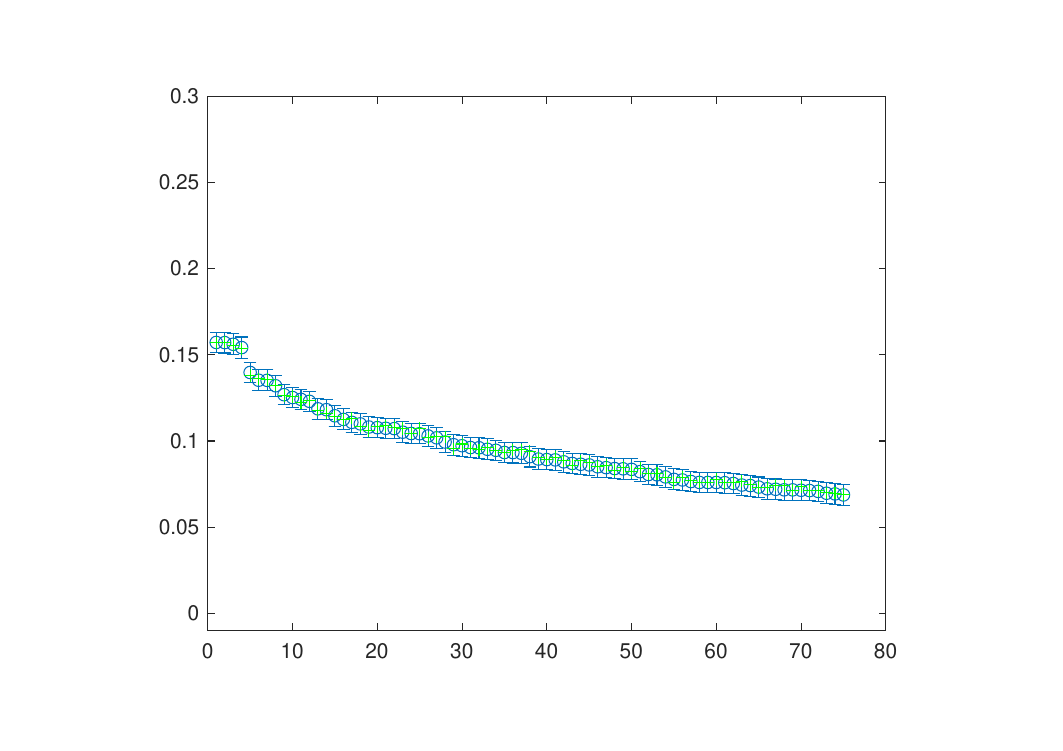}
        \caption{}
        \label{subfig:CI_comp_new}
    \end{subfigure}
    \begin{subfigure}[T]{0.19\linewidth}
        \includegraphics[width=\textwidth]{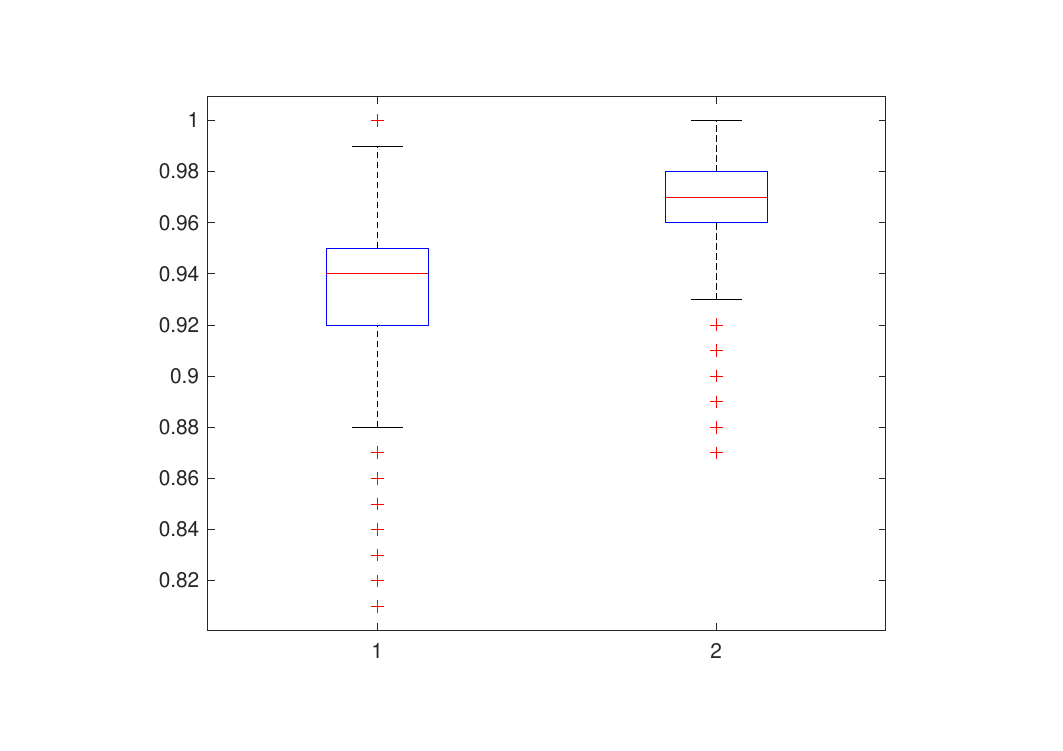}
        \caption{}
        \label{subfig:CI_comp_box_all_old}
    \end{subfigure}
    \begin{subfigure}[T]{0.19\linewidth}
        \includegraphics[width=\textwidth]{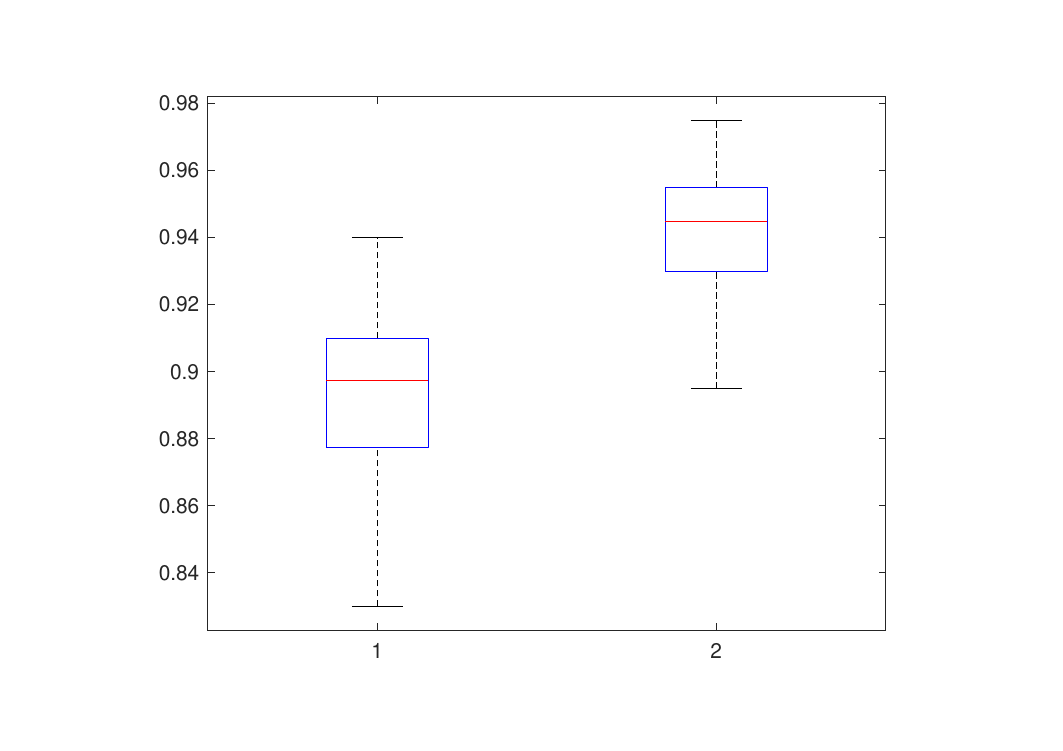}
        \caption{}
        \label{subfig:CI_comp_box_S_old}
    \end{subfigure}
    \caption{Illustration of the confidence interval correction. Figs. \ref{subfig:CI_comp_old}, \ref{subfig:CI_comp_gauss}, \ref{subfig:CI_comp_new} show the construction of CIs with standard debiased techniques (w/o data adjustment) and with our proposed method (w/ Gaussian adjustment - Thm. \ref{thm:remainder_dist_gaussian} - in Fig. \ref{subfig:CI_comp_gauss} and data adjustment - Thm. \ref{thm:main_stat_result} - in Fig. \ref{subfig:CI_comp_new}), respectively. The red points represent the entries that are not captured by the CIs. Additionally, Fig. \ref{subfig:CI_comp_box_all_old} shows box plots of coverage over all components, and Fig. \ref{subfig:CI_comp_box_S_old} shows them on the support. In the last two plots, the left box refers to the asymptotic and the right to the non-asymptotic CI based on Gaussian adjustment of $500$ feature vectors. We solve a sparse regression problem $y=Ax+\varepsilon$ via the LASSO, where $A \in \mathbb{C}^{4000 \times 10000}$, $x \in \mathbb{C}^N$ is 200-sparse, and the noise level is $\approx 10\%$. The averaged coverage over $250$ vectors with significance level $\alpha=0.05$ of the asymptotic confidence intervals is $h^W(0.05)=0.9353$ and on the support $h^W_S(0.05)=0.8941$. Confidence intervals built with our proposed method yield for Gaussian adjustment $h^G(0.05)=0.9684$ and on the support $h_S^G(0.05)=0.9421$, and for data-driven adjustment $h(0.05)=h_S(0.05)=1$. For more details, cf. Section \ref{subsec:classic regression} and Appendix \ref{sec:further_numerical}.}
    \label{fig:comparison_CI}
\end{figure}

The idea of the debiased LASSO is that its estimation error, i.e., the difference between the debiased estimator and the ground truth, can be decomposed into a \textbf{Gaussian} and a \textbf{remainder/bias} component. It has been shown in certain cases that the $\ell_{\infty}$ norm of the remainder component vanishes with high probability, assuming an \emph{asymptotic} setting, i.e., when the dimensions of the problem grow. In this case, the estimator is proven to be \emph{approximately Gaussian} from which the confidence intervals are derived. However, in practice, one needs to be in a very high-dimensional regime with enough data for these assumptions to kick in. In many applications with a finite set of observations, the \emph{remainder term does not vanish}; it can rather be substantially large, and the confidence intervals constructed solely based on the Gaussian component \emph{fail to account for the entire estimation error}. Consequently, the derived confidence intervals are narrower, resulting in an overestimation of certainty. This issue is particularly problematic in applications where it is crucial to estimate the magnitude of a vector coefficient with a high degree of confidence, such as in medical imaging applications.

Moreover, according to the standard theory of debiased estimators, the estimation of how small the \textbf{remainer term} is depends on how well one can quantify the $\ell_2$ and $\ell_1$ bounds for the corresponding \emph{biased estimator}, e.g., the LASSO \cite{vandeGeer.2014,Javanmard.2018}. Although sharp oracle inequalities exist for such classical regression estimators, cf. \emph{related works}, the same cannot be said about when unrolled algorithms are employed. For the latter, generalization bounds are usually not sharp or do not exist.

In this paper, we tackle the challenge of constructing valid confidence intervals around debiased estimators used in high-dimensional regression. The key difficulty lies in accounting for the remainder term in the estimation error decomposition, which hinders the development of finite-sample confidence intervals. We propose a \textbf{novel non-asymptotic theory that explicitly characterizes the remainder term}, enabling us to construct reliable confidence intervals in the finite-sample regime. Furthermore, we extend our framework to quantify uncertainty for model-based neural networks used for solving inverse problems, which paves the way to a rigorous theory of data-driven UQ for modern deep learning techniques. We state an informal version of our main result, discussed in detail in Section \ref{data-driven_CI}.

\begin{theorem}[Informal Version]\label{thm:informal}
    Let $x^{(1)},\hdots, x^{(l)}\in\mathbb{C}^N$ be i.i.d. data. Let $b^{(i)}=Ax^{(i)}+\varepsilon^{(i)}$ be a high-dimensional regression model with noise $\varepsilon^{(i)} \sim \mathcal{CN}(0, \sigma^2 I_{N \times N})$. With the data, derive, for a significance level $\alpha$, a confidence radius $r_j(\alpha)$ for a new sample's component $x^{(l+1)}_j$. Let $(\hat{x}^u)^{(l+1)}_j$ be the debiased estimator based on a (learned) high-dimensional regression estimator $\hat{x}^{(i)}_j$. Then, it holds that
    \begin{equation*}
        \mathbb{P}\left( \left\vert (\hat{x}^u)^{(l+1)}_j - x^{(l+1)}_j \right\vert \leq r_j(\alpha) \right) \geq 1-\alpha.
    \end{equation*}
\end{theorem}

Theorem \ref{thm:informal} has far-reaching implications that transcend the classical regularized high-dimensional regression setting. For example, it enables the establishment of rigorous confidence intervals for learning algorithms such as unrolled networks \cite{monga2021algorithm}. To our knowledge, obtaining rigorous UQ results for neural networks without relying on non-scalable Monte Carlo methods remains a challenging problem \cite{gawlikowski2023survey}. To address this and quantify uncertainty, our approach combines model-based prior knowledge with data-driven statistical techniques.
The model-based component harnesses the Gaussian distribution of the noise to quantify the uncertainty arising from the noisy data itself. We note that the Gaussian assumption for the noise is not a limitation, and extensions to non-Gaussian distributions are also possible, as clarified by \cite{vandeGeer.2014}. We make a Gaussian noise assumption here for the sake of clarity. Complementing this, the data-driven component is imperative for quantifying the uncertainty inherent in the estimator's performance. Moreover, \emph{our approach does not require any assumptions regarding the convergence or quality properties of the estimator}. This flexibility enables the debiased method to apply to a wide range of estimators.

\textbf{Contributions.} The key contributions in this work are threefold  \footnote{ The code for our findings is available on GitHub : \url{https://github.com/frederikhoppe/UQ_high_dim_learning} }
\begin{enumerate}
    \item We solve the problem illustrated in Fig. \ref{fig:comparison_CI} by \textbf{developing a non-asymptotic theory for constructing confidence intervals around the debiased LASSO estimator}. Unlike existing approaches that rely on asymptotic arguments and ignore the remainder term, \emph{our finite-sample analysis explicitly accounts for the remainder}, clarifying an important theoretical gap and providing rigorous guarantees without appealing to asymptotic regimes.
    \item We establish a general framework that \textbf{extends the debiasing techniques to model-based deep learning approaches for high-dimensional regression}. Our results enable the principled measurement of uncertainty for estimators learned by neural networks, a capability crucial for reliable decision-making in safety-critical applications. We test our approach with state-of-the-art unrolled networks such as the It-Net \cite{genzel2022near}.
    \item For real-world medical imaging tasks, we demonstrate that the remainder term in the debiased LASSO estimation error can be accurately modeled as a Gaussian distribution. Leveraging this finding, we derive \textbf{Gaussian adjusted CIs that provide sharper uncertainty estimates than previous methods}, enhancing the practical utility of debiased estimators in high-stakes medical domains.
\end{enumerate}

\section{Background and Problem Formulation}
In numerous real-world applications, we encounter high-dimensional regression problems where the number of features far exceeds the number of observations. This scenario, known as high-dimensional regression, arises when we aim to estimate $N$ features, described by $x^0 \in \mathbb{C}^N$ from only a few $m$ target measurements $b \in \mathbb{C}^m$, where $m \ll N$. Mathematically, this can be expressed as a linear model $b = Ax^0 + \varepsilon$, where $A \in \mathbb{C}^{m \times N}$ is the measurement matrix and $\varepsilon \sim \mathcal{CN}(0, \sigma^2 I_{N \times N})$ is additive Gaussian noise with variance $\sigma^2$.
In the presence of sparsity, where the feature vector $x^0$ has only $s$ non-zero entries ($s \ll N$), a popular approach is to solve the LASSO, which gives an estimator $\hat{x}$ obtained by solving the following $\ell_1$-regularized optimization problem:
\begin{equation}\label{eq:LASSO}
\min\limits_{x \in \mathbb{C}^N} \frac{1}{2m} \Vert Ax - b \Vert_2^2 + \lambda \Vert x \Vert_1.
\end{equation}

However, the LASSO estimator is known to exhibit a systematic bias, and its distribution is intractable, posing challenges for uncertainty quantification \cite{Fu2000}. To address this limitation, debiasing techniques have been developed in recent years \cite{Zhang.2014,Javanmard.2014,vandeGeer.2014}. The debiased LASSO estimator, $\hat{x}^u$, is defined as:
\begin{equation}
\hat{x}^u = \hat{x} + \frac{1}{m} MA^*(A\hat{x} - b),
\end{equation}
where $M$ is a correction matrix that could be chosen such that $\max_{i,j\in\{1,\hdots,N\}}\vert (M\hat{\Sigma}-I_{N\times N})_{ij}\vert$ is small. Here, $\hat{\Sigma}=\frac{A^*A}{m}$. We refer to \cite{Javanmard.2018} for a more detailed description of how to choose $M$. Remarkably, the estimation error
\begin{equation}\label{eq:decomp}
    \hat{x}^u - x^0=\underbrace{MA^*\varepsilon/m}_{=:W} + \underbrace{(M\hat{\Sigma} - I_{N\times N})(x^0-\hat{x})}_{=:R},
\end{equation}
can be decomposed into a Gaussian component $W \sim \mathcal{CN}(0, \frac{\sigma^2}{m}\hat{\Sigma})$ and a remainder term $R$ that vanishes asymptotically with high probability \cite[Theorem 3.8]{Javanmard.2018}, assuming a Gaussian measurement matrix $A$. Such a result was extended to matrices associated to a bounded orthonormal system like a subsampled Fourier matrix, allowing for extending the debiased LASSO to MRI \cite{hoppe2022uncertainty}. The decomposition \eqref{eq:decomp} and the asymptotic behavior of $R$ enable the construction of asymptotically valid CIs for the debiased LASSO estimate, providing principled UQ for high-dimensional sparse regression problems.

However, in real-world applications involving finite data regimes, the remainder term can be significant, rendering the asymptotic confidence intervals imprecise or even misleading, as illustrated in Fig. \ref{fig:comparison_CI}. This issue is particularly pronounced in high-stakes domains like medical imaging, where reliable UQ is crucial for accurate diagnosis and treatment planning. Second, the debiasing techniques have thus far been restricted to estimators whose error is well quantifiable, leaving the \textbf{challenge of how they would behave for deep learning architectures open}. In such cases, the behavior of the remainder term is largely unknown, precluding the direct application of existing debiasing methods and hindering the deployment of these methods in risk-sensitive applications.

A prominent example for solving the LASSO problem in \eqref{eq:LASSO} with an unrolled algorithm is the ISTA \cite{daubechies2004iterative,ISTAfirst}:
\begin{equation*}
x^{k+1} = \mathcal{S}_\lambda\left((I_{N\times N} - \frac{1}{\mu}A^TA)x^k + \frac{1}{\mu}A^Tb\right), \qquad k \geq 0.
\end{equation*}
Here, $\mu > 0$ is a step-size parameter, and $\mathcal{S}_\lambda(x)$ is the soft-thresholding operator. The work \cite{gregor2010learning} interpreted each ISTA iteration as a layer of a recurrent neural network (RNN). The Learned ISTA (LISTA) approach learns the parameters $W_1^k, W_2^k, \lambda^k$ instead of using the fixed ISTA updates:
\begin{equation*}
x^{k+1} = \mathcal{S}_{\lambda^k}(W_2^k x^k + W_1^k b).
\end{equation*}
In this formulation, LISTA unrolls $K$ iterations into $K$ layers, with learnable parameters $(W^k, \lambda^k)$ per layer. The parameters are learned by minimizing the reconstruction error $\min_{\lambda, W}\frac{1}{l} \sum_{i=1}^l \lVert x_i^k(\lambda, W, b^{(i)}, x^{(i)})-x^{(i)} \rVert_2^2$ on training data $(x^{(i)},b^{(i)})$. Unrolled neural networks like LISTA have shown promise as model-based deep learning solutions for inverse problems, leveraging domain knowledge for improved performance. Such iterative end-to-end network schemes provide state-of-the-art reconstructions for inverse problems \cite{genzel2022near}. Recently, the work \cite{hoppe2023uncertainty} proposes a framework based on the debiasing step to derive confidence intervals specifically for the unrolled LISTA estimator. However, similar to the previously mentioned debiased LASSO literature, \emph{it only handles the asymptotic setting}.

\textbf{Related Works.} \textit{High-dimensional regression.} High-dimensional regression and sparse recovery is now a well-established theory, see \cite{Foucart.2013,wainwright2019high,wright2022high}. In this context, several extensions of the LASSO have been proposed such as the elastic net \cite{zou2005regularization}, the group LASSO \cite{groupLasso}, the LASSO with a nuclear norm penalization \cite{lassoNuclearNorm}, the Sorted L-One Penalized Estimation (SLOPE) \cite{bogdan2015slope} which adapts the $\ell_1$-norm to control the false discovery rate. In addition to convex penalty functions, concave penalties have been explored to address some limitations of the LASSO, e.g., the Smoothly Clipped Absolute Deviation (SCAD) penalty \cite{fan2001variable} and the Minimax Concave Penalty (MCP) \cite{zhang2010nearly}. Non-convex variants of the LASSO for $\ell_p$-norm ($p<1$) minimization were also studied \cite{zheng2017does,rakotomamonjy2022convergent} as well as noise-bling variants such as the square-root LASSO \cite{belloni2011square,verdun2024fast}. Scalable and fast algorithms for solving the LASSO and its variants include semi-smooth Newton methods \cite{li2018highly} and IRLS \cite{kummerle2021iteratively}.

\textit{LASSO theory.} Several works have established oracle inequalities for the LASSO \cite{bunea2007sparsity,koltchinskii2009sparsity,ye2010rate,raskutti2011minimax,dalalyan2017prediction}. Another key theoretical result is the consistency of the LASSO in terms of variable selection. \cite{zhao2006model} and \cite{wainwright2009sharp} established the consistency of the LASSO while \cite{foucart2023sparsity} analyzed the sparsity behavior of the LASSO when the design matrices satisfy the Restricted Isometry Property.

\textit{Debiased estimators.} After the first papers about the debiased LASSO \cite{Zhang.2014,vandeGeer.2014,Javanmard.2014}, some works have focused on improving its finite-sample performance and computational efficiency \cite{Javanmard.2018, li2020debiasing}. The size of the confidence intervals derived for the debiased LASSO has been proven to be sharp in the minimax sense \cite{cai2017confidence}. Debiased estimators have been extended in several directions, e.g., \cite{li2020debiasing,hoppe2022uncertainty,guo2022doubly,bellec2022biasing}. Recently, \cite{bellec2023debiasing} established asymptotic normality results for a debiased estimator of convex regularizers beyond the $\ell_1$-norm. In the context of MR images, \cite{hoppe2024tv} explored a debiased estimator for inverse problems with a total variation regularizer.
Debiased estimators have also been recently extended to unrolled estimators -- see discussion in the next paragraph -- in \cite{bellec2024uncertainty, hoppe2023uncertainty}.

\textit{Algorithm unrolling and model-based deep learning for inverse problems.} The idea of unfolding the iterative steps of classical algorithms into a deep neural network architecture dates back to \cite{gregor2010learning}, which proposed the Learned ISTA (LISTA) to fast approximate the solution of sparse coding problems. Several works have extended and improved upon the original LISTA framework \cite{ito2019trainable,wu2019sparse,liu2019alista,chen2021hyperparameter,aberdam2021ada,zheng2022hybrid}.
\cite{adler2018learned} proposed the Learned Primal-Dual algorithm, unrolling the primal-dual hybrid gradient method for tomographic reconstruction. \cite{schlemper2017deep} proposed the Deep Cascade of Convolutional Neural Networks (DC-CNN) for dynamic MRI reconstruction. \cite{cherkaoui2020learning} unfolded proximal gradient descent solvers to learn their parameters for 1D TV regularized problems.
\cite{monga2021algorithm} introduced a general framework for algorithm unrolling. \cite{aggarwal2018modl} developed MoDL, a model-based deep learning approach for MRI reconstruction that unrolls the ADMM algorithm. \cite{liu2019deep} proposed a proximal alternating direction network (PADNet) to unroll nonconvex optimization. See also the surveys for more information about unrolling and also the connection with physics-inspired methods \cite{zhang2023physics,arridge2019solving}. \cite{genzel2022near,genzel2022solving} developed the It-Net, an unrolled proximal gradient descent scheme where the proximal operator is replaced by a U-Net. This scheme won the AAPM Challenge 2021 \cite{sidky2022report} whose goal was \emph{to identify the state-of-the-art in solving the CT inverse problem with data-driven techniques}. A generalization of the previous paradigm is the learning to optimize framework that develops an optimization method by training, i.e., learning from its performance on sample problem \cite{li2016learning,chen2022learning}. 

\textit{Uncertainty Quantification.} There have been a few attempts to quantify uncertainty on a pixel level for unrolled networks, e.g., \cite{ekmekci2022uncertainty}. However, such approaches are based on Bayesian networks and MC dropout \cite{gal2016dropout}, which requires significant inference time paired with a loss of reconstruction performance since the dropout for UQ is a strong regularizer in the neural network. Unlike prior work, our contribution focuses on a scalable data-driven method that is easily implementable in the data reconstruction pipeline.

\section{Data-Driven Confidence Intervals}\label{data-driven_CI}
We now introduce our data-driven approach to correct the CIs. Instead of deriving asymptotic CIs from the decomposition $\hat{x}^u-x^0 = W + R$, by assuming that $R$ asymptotically vanishes, we utilize data $\left(b^{(i)}, x^{(i)}\right)_{i=1}^l$ along with concentration techniques to estimate the size of the bias component $R$. We continue to leverage the prior knowledge of the Gaussian component $W$ while extending the CIs' applicability to a broad class of estimators, including neural networks. Our method is summarized in Algorithm \ref{alg:data_estimation}, where the data is used to estimate the radii of the CIs, and in Algorithm \ref{alg3}, which constructs the estimator around which the CIs are centered. The following main result proves the validity of our method.

\begin{theorem}\label{thm:main_stat_result}
    Let $x^{(1)},\hdots, x^{(l)}\in\mathbb{C}^N$ be i.i.d. complex random variables representing ground truth data drawn from an unknown distribution $\mathbb{Q}$. Suppose, that $\varepsilon^{(i)}\sim\mathcal{CN}(0,\sigma^2 I_{m\times m})$ is noise in the high-dimensional models $b^{(i)} = A x^{(i)} + \varepsilon^{(i)}$, where $A\in\mathbb{C}^{m\times N}$, and independent of the $x^{(i)}$'s. Let $\hat{X}:\mathbb{C}^m\to\mathbb{C}^N$ be a (learned) estimation function that maps the data $b^{(i)}$ to $\hat{x}^{(i)}$, which is an estimate for $x^{(i)}$. Set $\vert R^{(i)}_j\vert = \vert e_j^T (M\hat{\Sigma}-I_{N\times N})(\hat{x}^{(i)}-x^{(i)})\vert$ for fixed $A$ and $M$. For $j=1,\hdots, N$, we denote the true but unknown mean with $\mu_j=\mathbb{E} [\vert R^{(1)}_j\vert]$ and the unknown variance with $(\sigma_R^2)_j:=\mathbb{E}[(\vert R^{(1)}_j\vert-\mu_j)^2]$, respectively. Let $\hat{S}_j = \frac{1}{l}\sum\limits_{i=1}^l \vert \ R^{(i)}_j\vert$ be the unbiased sample mean estimator and $(\hat{\sigma}_R^2)_j = \frac{1}{l-1}\sum\limits_{i=1}^l(\vert R^{(i)}_j\vert - \hat{S}_j)^2$ the unbiased variance estimator. Let $\alpha\in(0,1)$ and $\gamma\in\left(0,1-\frac{1}{l\alpha}\right)$.
    Furthermore, set the confidence regions for the sample $x^{(l+1)}\sim\mathbb{Q}$ in the model $b^{(l+1)} = Ax^{(l+1)} + \varepsilon^{(l+1)}$ as $C_j(\alpha) = \{ z \in\mathbb{C}: \vert (\hat{x}^u)^{(l+1)}_j - z\vert \leq r_j(\alpha)\}$ with radius
    \begin{equation}
        r_j(\alpha) = \frac{\sigma (M\hat{\Sigma}M^*)_{jj}^{1/2}}{\sqrt{m}}\sqrt{\log\left(\frac{1}{\gamma_j\alpha}\right)} + c_l\left(\alpha\right)\cdot (\hat{\sigma}_R)_j + \hat{S}_j,
        \qquad c_l(\alpha) := \sqrt{\frac{l^2-1}{l^2(1-\gamma_j)\alpha - l }}.
    \end{equation}
    Then, it holds that
    \begin{equation}\label{eq:prob_CI}
        \mathbb{P}\left(x^{(l+1)}_j \in C_j(\alpha)\right) \geq 1-\alpha.
    \end{equation}
\end{theorem}
Theorem \ref{thm:main_stat_result} achieves conservative confidence intervals that are proven to be valid, i.e., are proven to contain the true parameter with a probability of $1-\alpha$. Its main advantage is that there are \emph{no assumptions} on the distribution $\mathbb{Q}$ (except that $\sigma_R^2$ exists), making it widely applicable. Hence, Theorem \ref{thm:main_stat_result} includes the worst-case distribution showing a way to quantify uncertainty even in such an ill-posed setting. Especially in medical imaging, such certainty guarantees are crucial for accurate diagnosis. The proof exploits the Gaussianity of the component $W$ as well as an empirical version of Chebyshev's inequality, which is tight when there is no information on the underlying distribution. The detailed proof can be found in Appendix \ref{sec:proof_main}. For a thorough discussion on Theorem \ref{thm:main_stat_result} including practical simplifications, we refer to Appendix \ref{sec:discussion}.

More certainty comes with the price of larger confidence intervals. If there is additional information on the distribution of $R$, like the ability to be approximated by a Gaussian distribution, then the confidence intervals become tighter. This case, which includes relevant settings such as MRI, is discussed in Section \ref{sec:gaussian_remainder}.

\begin{algorithm}[t]
\caption{Estimation of Confidence Radius}\label{alg:data_estimation}
\begin{algorithmic}[1]
  \State \textbf{Input:} Estimation function $\hat{X}$, dictionary matrix $A$, correction matrix $M$, data points $\left(b^{(i)}, x^{(i)}\right)_{i=1}^l$, significance level $\alpha$
  \For{$i=1,\hdots,l$}
  \State Compute $\hat{x}^{(i)},R^{(i)}\in\mathbb{C}^N$ via $\hat{x}^{(i)}=\hat{X}(b^{(i)})$ and $ R^{(i)}_j = e_j^T (M\hat{\Sigma}-I_{N\times N})(\hat{x}^{(i)}-x^{(i)})$.
  \EndFor
  \For{$j=1,\hdots,N$}
  \State Estimate $\hat{S}_j = \frac{1}{l}\sum\limits_{i=1}^l \vert R^{(i)}_j\vert$ and $(\hat{\sigma}_R^2)_j = \frac{1}{l-1}\sum\limits_{i=1}^l(\vert R^{(i)}_j\vert - \hat{S}_j)^2$
  \State Solve $r_j(\alpha) = \min\limits_{\gamma\in\left(0,1-\frac{1}{l\alpha}\right)} \frac{\sigma (M\hat{\Sigma}M^*)_{jj}^{1/2}}{\sqrt{m}}\sqrt{\log\left(\frac{1}{\gamma\alpha}\right)} + c_l\left((1-\gamma)\alpha\right)\cdot (\hat{\sigma}_R)_j + \hat{S}_j$
  \EndFor
  \State \textbf{Output:} Radii of confidence regions $\left(r_j(\alpha)\right)_{j=1}^N$
\end{algorithmic}
\end{algorithm}

\begin{algorithm}[t]
\caption{Construction of Confidence Regions}\label{alg3}
\begin{algorithmic}[1]
  \State \textbf{Input:} Estimation function $\hat{X}$, dictionary matrix $A$, correction matrix $M$, measurement $b$, radii $\left(r_j(\alpha)\right)_{j=1}^N$ (derived from Algorithm \ref{alg:data_estimation} or Theorem \ref{thm:remainder_dist_gaussian})
  \State Compute estimator $\hat{x}=\hat{X}(b)$.
  \State Construct debiased estimator via $\hat{x}^u = \hat{x} + \frac{1}{m}MA^*(b-A\hat{x})$
  \For{$j=1,\hdots,N$}
  \State Construct confidence region $C_j(\alpha) = \{z\in\mathbb{C}\mid \vert \hat{x}^u_j-z\vert\leq r_j(\alpha)\}$
  \EndFor
  \State \textbf{Output:} Debiased estimator $x^u$ and confidence regions $\left(C_1(\alpha)\right)_{j=1}^N$
\end{algorithmic}
\end{algorithm}

\section{Confidence Intervals for Gaussian Remainders}\label{sec:gaussian_remainder}

Valid confidence intervals can be derived most straightforwardly when the distribution of the remainder term is known and easily characterized. In such cases, more informative distributional assumptions lead to potentially tighter confidence intervals compared to Theorem \ref{thm:main_stat_result}, which makes no assumptions about the remainder component. In this section, we derive non-asymptotic confidence intervals assuming the remainder term to be approximated by a Gaussian distribution.

\begin{theorem}\label{thm:remainder_dist_gaussian}
    Let $\hat{x}^u\in\mathbb{C}^N$ be a debiased estimator for $x\in\mathbb{C}^N$ with a remainder term $R\sim\mathcal{CN}(0,\Sigma_R/m)$. Then, $C_j(\alpha)=\{ z \in\mathbb{C}\mid \vert z-\hat{x}^u_j\vert \leq r_j(\alpha)\}$ with radius
    \begin{equation}
        r^G_j(\alpha) = \frac{(\sigma^2(M\hat{\Sigma}M^*)_{jj}+(\Sigma_R)_{jj})^{1/2}}{\sqrt{m}}\sqrt{\log\left(\frac{1}{\alpha}\right)}.
    \end{equation}
    is valid, i.e. $\mathbb{P}\left( x_j \in C_j(\alpha)\right)\geq 1-\alpha$.
\end{theorem}
For the proof, we refer to Appendix \ref{sec:proof_main}. In Appendix \ref{sec:gaussian_remainder_density}, we demonstrate empirically that the Gaussian assumption for the remainder term holds in a wide range of relevant practical settings. This validation enables the application of the proposed confidence intervals derived under this assumption. These confidence intervals strike a careful balance between non-asymptotic reliability, ensuring valid coverage even in finite-sample regimes, and tightness, providing informative and precise uncertainty estimates. By leveraging the Gaussian approximation, which becomes increasingly accurate in higher dimensions as illustrated in Figure \ref{fig:densities_sparse_regr}, our framework offers a principled and computationally efficient approach to quantifying uncertainty in high-dimensional prediction problems. The variance of $R$ can be estimated with the given data using, e.g., the unbiased estimator for the variance as in Theorem \ref{thm:main_stat_result}.

\section{Numerical Experiments}
\label{sec:numerical_exp}
We evaluate the performance of our non-asymptotic confidence intervals through extensive numerical experiments across two settings: (i.) the \emph{classical debiased LASSO framework} to contrast our non-asymptotic confidence intervals against the asymptotic ones. (ii.) the \emph{learned framework} where we employ learned estimators, specifically the U-net \cite{ronneberger2015u} as well as the It-Net \cite{genzel2022near}, to reconstruct real-world MR images and quantify uncertainty. Our experiments demonstrate the importance of properly accounting for the remainder term in practical, non-asymptotic regimes. Each experiment follows the same structure:

\begin{enumerate}
    \item Data Generation and Management: We fix the forward operator $A$ and generate $n>2$ feature vectors ${x^{(i)}}_{i=1}^n$ and noise vectors ${\varepsilon^{(i)}}_{i=1}^n$ with $\varepsilon^{(i)}\sim\mathcal{CN}(0,\sigma^2I_{m\times m})$. We obtain observations ${b^{(i)}}_{i=1}^n$ via $b^{(i)} = A x^{(i)} + \varepsilon^{(i)}$. We split the data ${(b^{(i)}, x^{(i)})}_{i=1}^n$ into an \emph{estimation} dataset of size $l$ and a test dataset of size $k$ ($l+k=n$). If we learn an estimator, we further split the data into training, estimation, and test sets.
    \item Reconstruction: Depending on the experiment, we obtain a reconstruction function $\hat{X}$ in one of the following ways: for the classical LASSO setting, we use the LASSO; for the learned estimator experiment, we train a U-Net  \cite{ronneberger2015u} or It-net \cite{genzel2022near} on the training data to serve as the reconstruction function $\hat{X}$.
    \item Estimation of Confidence Radii: We run Algorithm \ref{alg:data_estimation} with $A, \hat{X}, M$ (that is chosen according to \cite{Javanmard.2018}), the estimation data ${(b^{(i)}, x^{(i)})}_{i=1}^l$, and a predefined significance level $\alpha\in(0,1)$ to obtain radii ${r_j(\alpha)}_{j=1}^N$. To construct the final confidence intervals, the radii need to be centered according to the debiased estimator. For every new measurement $b$, we run Algorithm \ref{alg3} to obtain tailored confidence intervals for the feature vector $x$ corresponding to $b$. In addition, we compute the CI for the Gaussian adjustment based on Theorem \ref{thm:remainder_dist_gaussian} using the estimation set to quantify the variance of $R$ with the unbiased estimator for the variance as before.
    \item Evaluation: We use the test dataset ${(b^{(i)}, x^{(i)})}_{i=l+1}^{k}$ to evaluate our adjustments. For each $b^{(i)}$, we run Algorithm \ref{alg3} to obtain confidence intervals ${C_j^{(i)}(\alpha)}_{j=1}^N$ for $x^{(i)}$. We estimate $\mathbb{P}(x_j^{(i)}\in C_j(\alpha))$ by $h_j(\alpha) = \frac{1}{k}\sum_{i=l+1}^k\mathbbm{1}_{\{x_j^{(i)}\in C_j(\alpha)\}}$ and average over all components $h(\alpha)=\frac{1}{N}\sum_{j=1}^N h_j(\alpha)$. Since performance on the support $S$ is crucial, we define the hit rate on $S$ as $h_S^{(i)} = \frac{1}{|S|}\sum_{j=1}^N\mathbbm{1}_{\{x_j^{(i)}\in C_j(\alpha)\}}$ and average $h_S(\alpha) = \frac{1}{l}\sum_{i=1}^lh_S^{(i)}$. Note that the support may change with $i$. Moreover, we do the same for the CI based on the Gaussian-adjusted radii.
\end{enumerate}

\subsection{UQ for Classical Model-Based Regression}\label{subsec:classic regression}
We consider a setting aligned with existing debiased LASSO literature, e.g., \cite{Javanmard.2014} to demonstrate our approach's extension of current UQ methods. The forward operator is a complex Gaussian matrix $A\in\mathbb{C}^{m\times N}$ with dimensions $N=10000$, $m=0.6N$, and $A_{ij}\sim\mathcal{CN}(0,1)$. We generate $n=750$ $s=0.1N$-sparse features $x^{(i)}$ by randomly selecting $m$ distinct indices from ${1,\hdots,N}$ and drawing magnitudes from $\mathcal{CN}(0,1)$. With relative noise $\frac{\Vert \varepsilon^{(i)}\Vert}{\Vert Ax^{(i)}\Vert}\approx 0.2$, we split the data ${(b^{(i)}, x^{(i)})}_{i=1}^n$ into $l=500$ estimation and $k=250$ test data. For reconstruction, we solve the LASSO $\hat{X}(b):=\operatorname{argmin}_{x\in\mathbb{C}^N} \frac{1}{m}\Vert Ax-b\Vert + \lambda \Vert x\Vert_1$ with $\lambda = 10\frac{\sigma}{\sqrt{m}}(2+\sqrt{12\log(N)})$ following \cite{hoppe2022uncertainty}.

With significance level $\alpha = 0.05$, we run Algorithm \ref{alg:data_estimation} to obtain confidence radii, choosing $M=I_{N\times N}$ \cite{Javanmard.2018} and exploiting the relaxation \eqref{eq:one_gamma}. Averaged over the $l$ estimation data points, the $\ell_2$ and $\ell_\infty$ norm ratios are: $\frac{\Vert R\Vert_2}{\Vert W\Vert_2} = 0.9993$ and $\frac{\Vert R\Vert_\infty}{\Vert W\Vert_\infty}=1.1581$. In existing literature, the $\ell_\infty$ norm is typically measured when the remainder term vanishes, as it is relevant for pixel-wise confidence intervals. Here, the remainder term is of comparable order as the Gaussian term and hence, too significant to neglect in confidence intervals derivation.

\begin{figure}
    \centering
    \begin{subfigure}[T]{0.19\linewidth}
        \includegraphics[width=\linewidth]{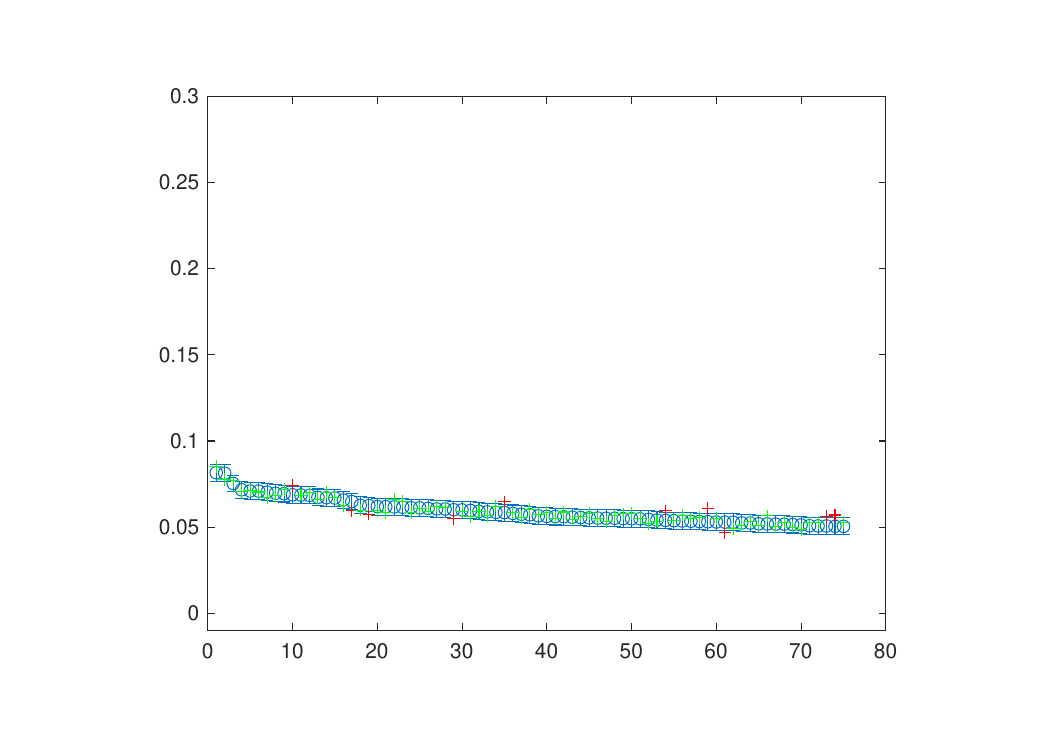}
        \caption{}
        \label{subfig:3comparison_old}
    \end{subfigure}
    \begin{subfigure}[T]{0.19\linewidth}
        \includegraphics[width=\linewidth]{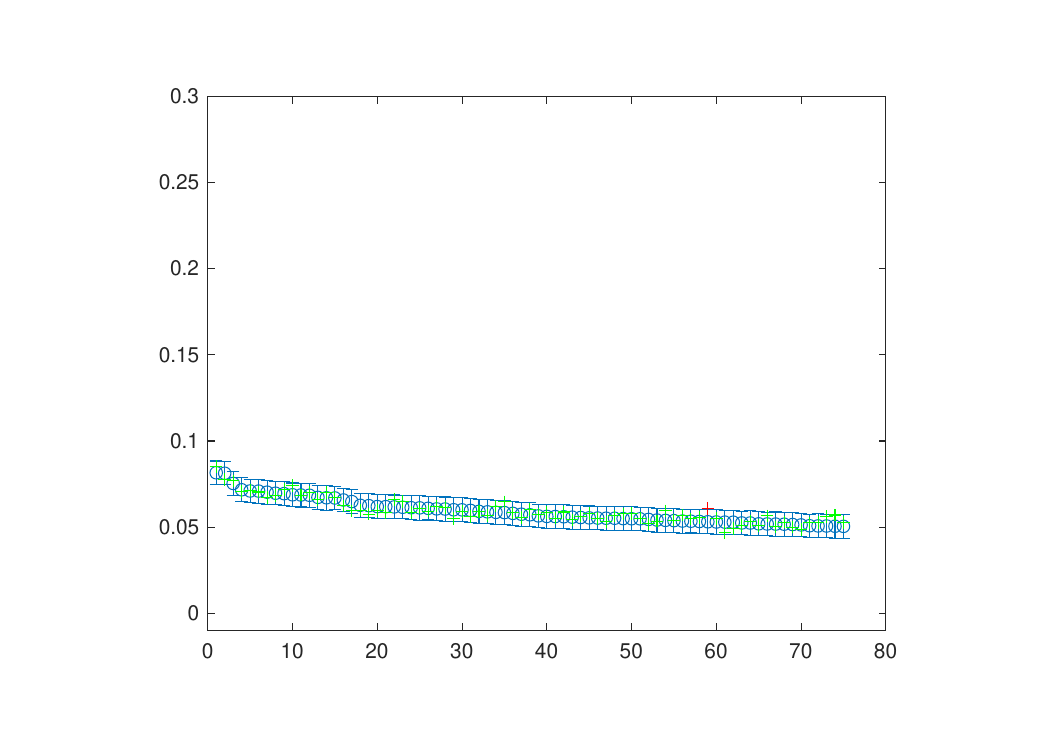}
        \caption{}
        \label{subfig:3comparison_gauss}
    \end{subfigure}
    \begin{subfigure}[T]{0.19\linewidth}
        \includegraphics[width=\linewidth]{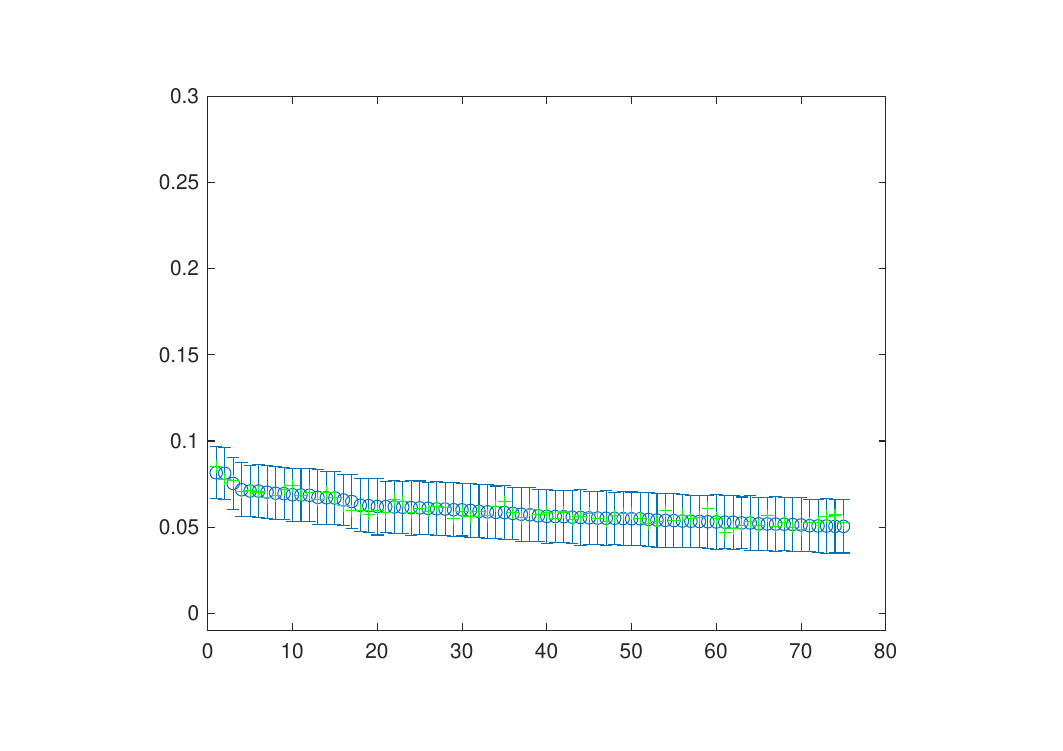}
        \caption{}
        \label{subfig:3comparision_new}
    \end{subfigure}
    \begin{subfigure}[T]{0.19\linewidth}
        \includegraphics[width=\linewidth]{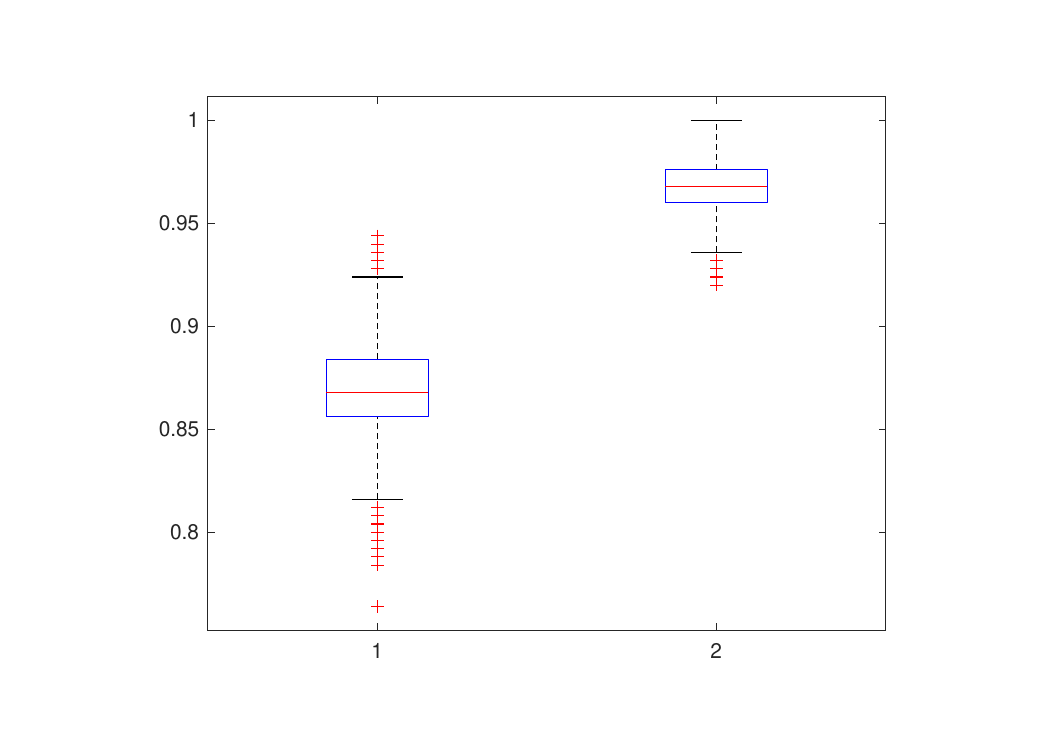}
        \caption{}
        \label{subfig:gauss_regr_boxplot_all}
    \end{subfigure}
    \begin{subfigure}[T]{0.19\linewidth}
        \includegraphics[width=\linewidth]{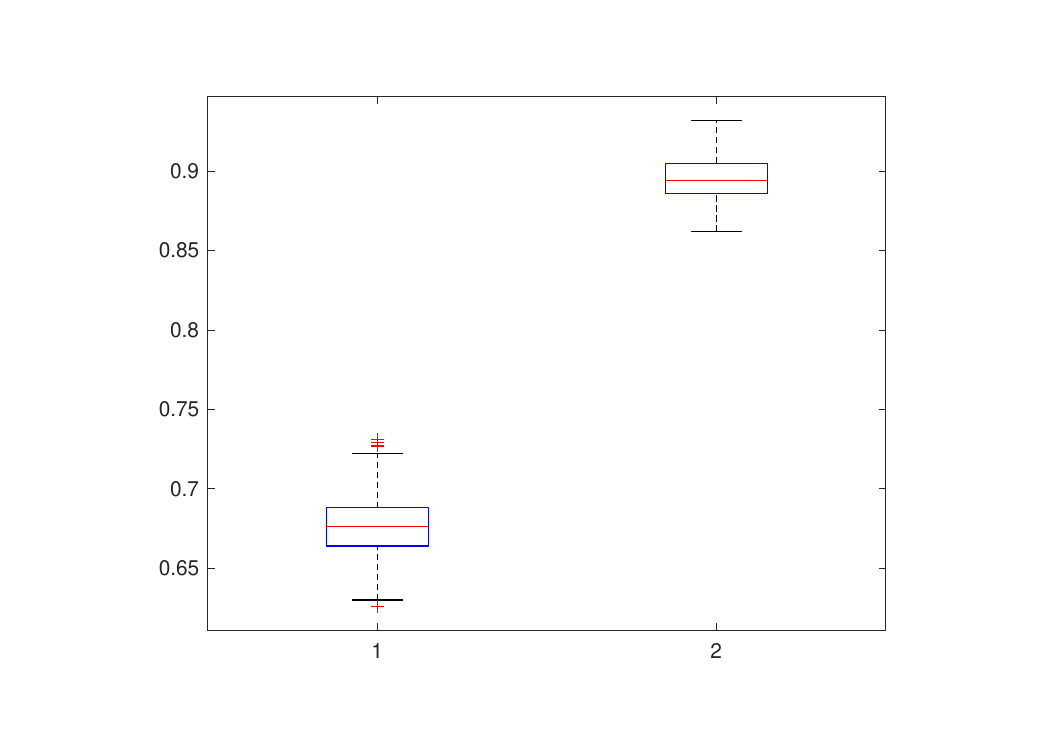}
        \caption{}
        \label{subfig:gauss_regr_boxplot_s}
    \end{subfigure}
    \caption{Confidence intervals of asymptotic type \ref{subfig:3comparison_old}, with Gaussian adjustment \ref{subfig:3comparison_gauss} and data-driven adjustment \ref{subfig:3comparision_new} for one evaluation feature vector in the sparse regression setting described in Section \ref{subsec:classic regression}. Box plots of hit rates $h_j(0.05)$ and $h_j^G(0.05)$, $j=1,\hdots,N$, averaged over feature vectors $x^{(1)},\hdots, x^{(500)}$ \ref{subfig:gauss_regr_boxplot_all} and hit rates $h_S(0.05)^{(i)}$ and $(h_S^G)^{(i)}(0.05)$, $i=1,\hdots,k$, averaged over components $j$ \ref{subfig:gauss_regr_boxplot_s}.}
    \label{fig:gaussian_boxplots}
\end{figure}

Evaluating it on the remaining $k=250$ data points, the data-driven and Gaussian-adjusted averaged hit rates are $h(0.05) = 1$, $h_S(0.05)=1$ and $h^G(0.05)=0.9691$, $h^G_S(0.05) = 0.8948$, respectively. Neglecting the remainder term yields $h^W(0.05) = 0.8692$ and $h^W_S(0.05) = 0.6783$, which is substantially lower and violates the specified $0.05$ significance level. Fig. \ref{fig:gaussian_boxplots} presents confidence intervals of each type for one data point $x^{(i)}$ and a detailed visualization of $h_j(0.05)$, $h_S^{(i)}(0.05)$,  $h_j^G(0.05)$, and $(h^G_S)^{(i)}(0.05)$. Further experiments with different sparse regression settings, including subsampled Fourier matrices, are presented in Appendix \ref{sec:further_numerical}.

\subsection{UQ for MRI Reconstruction with Neural Networks}
\label{subsec:UQNN}
We extend the debiasing approach to model-based deep learning for MRI reconstruction using the U-Net and It-Net on single-coil knee images from the NYU fastMRI dataset \footnote{We obtained the data, which we used for conducting the experiments in this paper from the NYU fastMRI Initiative database (fastmri.med.nyu.edu) \cite{zbontar2019fastmri, fastMRIdataset}. The data was only obtained from the NYU fastMRI investigators, but they did not contribute any ideas, analysis, or writing to this paper. The list of the NYU fastMRI investigators, can be found at fastmri.med.nyu.edu, it is subject to updates.} \cite{zbontar2019fastmri, fastMRIdataset}. Here, the forward operator is the undersampled Fourier operator $\mathcal{P} \mathcal{F}\in\mathbb{C}^{m\times N}$ with $N=320\times 320$, $m=0.6N$, the Fourier matrix $\mathcal{F}$ and a radial mask $\mathcal{P}$, see Figure \ref{subfig:mask}. The noise level $\sigma$ is chosen such that the relative noise is approximately $0.1$. The data is split into training (33370 slices), validation (5346 slices), estimation (1372 slices), and test (100 slices) datasets.

We then train an It-Net \cite{genzel2022near} with $8$ layers, a combination of MS-SSIM \cite{wang2003multiscale} and $\ell_1$-losses and Adam optimizer with learning rate $5e^{-5}$ for $15$ epochs to obtain our reconstruction function $\hat{X}$.

\begin{figure}[t]
 \centering
 \begin{subfigure}[T]{0.26\linewidth}
 \rotatebox[origin=c]{180}
      {\includegraphics[width=\linewidth]{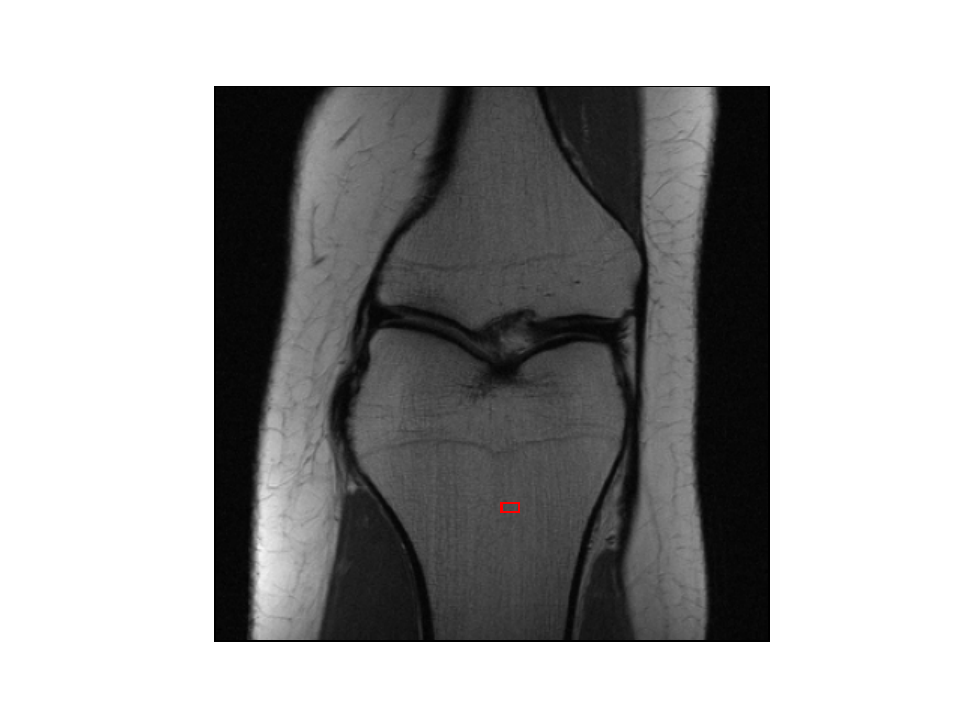}}
      \caption{}
      \label{subfig:imagewithbox}
  \end{subfigure}
  \hspace{-1cm}
  \begin{subfigure}[T]{0.26\linewidth}
      \includegraphics[width=\linewidth]{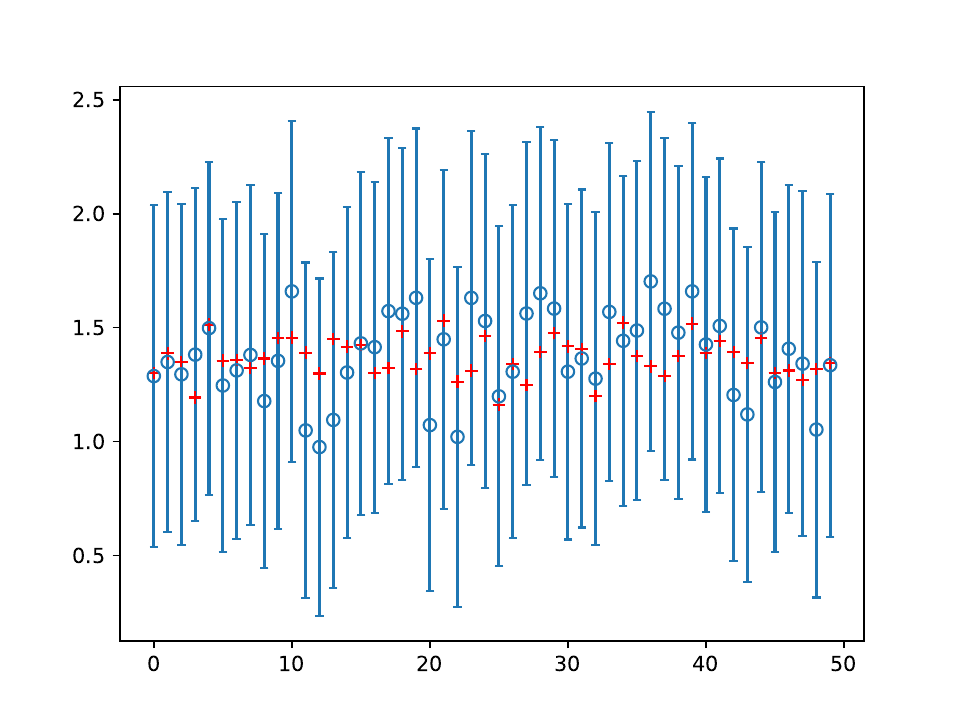}
      \caption{}
      \label{subfig:CIsnew}
  \end{subfigure}
   \hspace{-0.5cm}
   \begin{subfigure}[T]{0.26\linewidth}
   \includegraphics[width=\linewidth]{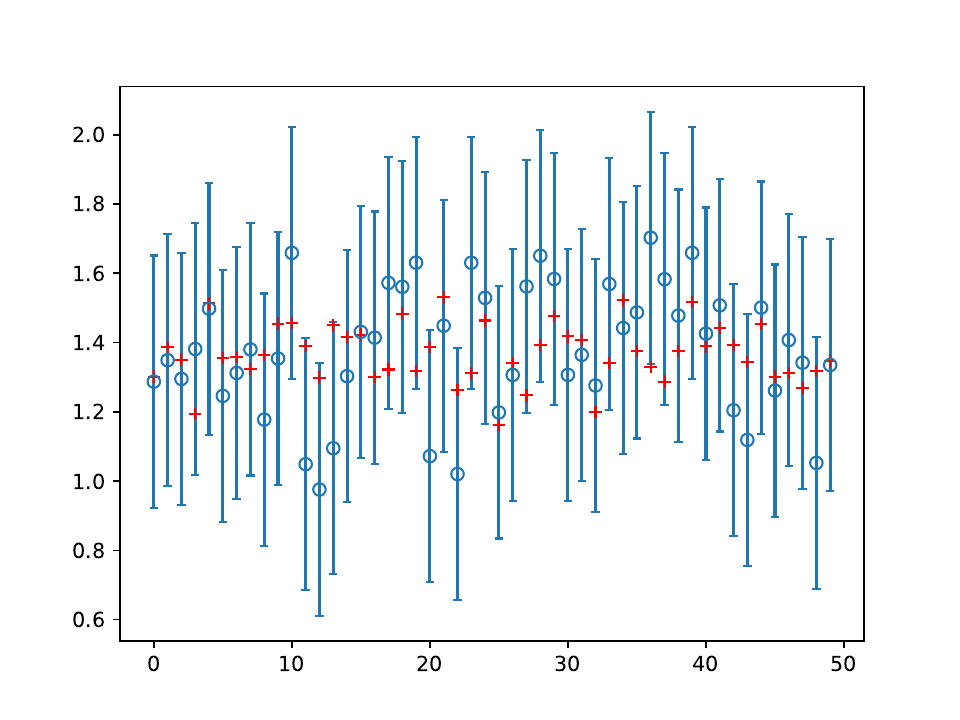}
   \caption{}
   \label{subfig:CIsold}
    \end{subfigure}
    \begin{subfigure}[T]{0.27\linewidth}
   \includegraphics[width=\linewidth]{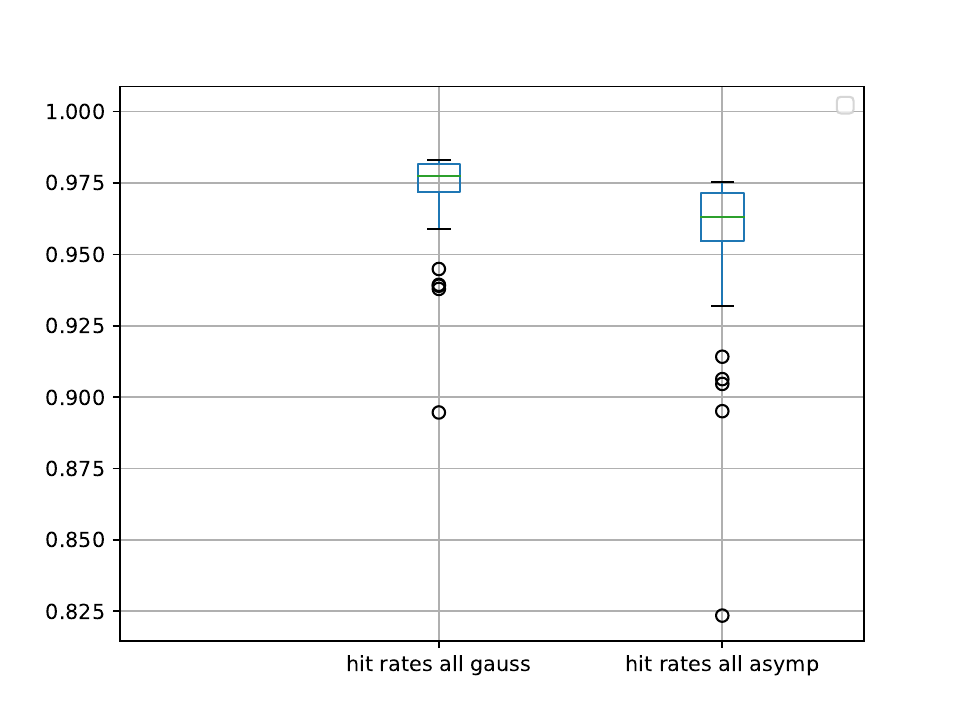}
   \caption{}
   \label{subfig:boxplot}
    \end{subfigure}
   \caption{Reconstruction obtained with the It-Net as described in \ref{subsec:UQNN}. Data-driven adjustment confidence intervals \ref{subfig:CIsnew} and asymptotic confidence intervals \ref{subfig:CIsold} for the region (50 pixels) in 320x320 knee image \ref{subfig:imagewithbox}; Box plots of hit rates \ref{subfig:boxplot} for $90 \%$ confidence level for the Gaussian adjusted and asymptotic confidence intervals.}
   \label{fig:mriCI}
\end{figure}

With significance level $\alpha = 0.1$, we run Algorithm \ref{alg:data_estimation} to construct confidence radii, choosing $M=I_{N\times N}$ \cite{Javanmard.2018} and exploiting the relaxation \eqref{eq:one_gamma}. Averaged over the $l$ estimation data points, we have $\frac{\Vert R\Vert_2}{\Vert W\Vert_2} = 0.38$ and $\frac{\Vert R\Vert_{\infty}}{\Vert W\Vert_{\infty}}= 0.49$, which indicates that the remainder term is significant and cannot be neglected. Evaluating the test data, the averages of the data-driven adjustment hit rates are $h(0.1) = 0.9999$, $h_S(0.1)=0.9998$, and the averages of the Gaussian adjusted hit rates are $h^G(0.1)=0.9752$, $h^G_S(0.1) = 0.98$. Neglecting the remainder term, the hit rates of the asymptotic CIs are $h^W(0.1) = 0.9502$ and $h^W_S(0.1) = 0.87$. As in the sparse regression setting, they are significantly lower. Fig. \ref{fig:mriCI} presents confidence intervals based on the data-driven adjustment and the asymptotic confidence intervals for a region in one image $x^{(i)}$. In addition, it contains a box plot showing the distribution of the hit rates based on the Gaussian adjustment and the asymptotic hit rates. More experiments for UQ for MRI reconstruction can be found in Appendix \ref{sec:further_numerical} and Tables \ref{tab:results_itnet} and \ref{tab:results_unet}.

\section{Final Remarks}
\label{sec:finalremarks}
In this work, we proposed a data-driven uncertainty quantification method that derives non-asymptotic confidence intervals based on debiased estimators. Our approach corrects asymptotic confidence intervals by incorporating an estimate of the remainder component and has solid theoretical foundations. While the correction can be based on prior knowledge, e.g., a Gaussian distribution of the remainder term, we also derive CI based on a data-driven adjustment without further information. This data-driven nature enhances its applicability to a wide range of estimators, including model-based deep-learning techniques. We conducted experiments that confirm our theoretical findings, demonstrating that even in classical sparse regression settings, the remainder term is too significant to be neglected. Furthermore, we applied the proposed method to MRI, achieving significantly better rates on the image support.

While our method corrects for the remainder term, larger remainder terms necessitate greater corrections, resulting in wider confidence intervals. Therefore, it is crucial to achieve a small remainder term to avoid excessively large confidence intervals. Additionally, the accuracy of our method depends on the quality of the estimates for the mean and variance of the remainder term, which improves with more available data. Additionally, the length of the intervals can be minimized over a larger parameter set, provided that more data is available. We leave as a future direction to study the sharpness of the proposed confidence intervals and radii for a given amount of data. Moreover, we would like to investigate how the length of the confidence intervals could be improved when estimating higher moments. We believe that our method is applicable to a wide variety of deep learning architectures, including vision transformers in MRI, e.g., \cite{lin2022vision}. Testing the generality of the method with state-of-the-art architectures for different problems would demonstrate its broad usefulness.

\section*{Acknowledgments}
We gratefully acknowledge financial support with funds provided by the German Federal Ministry of Education and Research in the grant \emph{``SparseMRI3D+: Compressive Sensing und Quantifizierung von Unsicherheiten f{\"u}r die beschleunigte multiparametrische quantitative
Magnetresonanztomografie (FZK 05M20WOA)''}.

\bibliographystyle{apalike}
\bibliography{refs}

\begin{thebibliography}{}

\bibitem[Aberdam et~al., 2021]{aberdam2021ada}
Aberdam, A., Golts, A., and Elad, M. (2021).
\newblock {Ada-lista: Learned solvers adaptive to varying models}.
\newblock {\em IEEE Transactions on Pattern Analysis and Machine Intelligence},
  44(12):9222--9235.

\bibitem[Adler and {\"O}ktem, 2018]{adler2018learned}
Adler, J. and {\"O}ktem, O. (2018).
\newblock {Learned primal-dual reconstruction}.
\newblock {\em IEEE transactions on medical imaging}, 37(6):1322--1332.

\bibitem[Aggarwal et~al., 2018]{aggarwal2018modl}
Aggarwal, H.~K., Mani, M.~P., and Jacob, M. (2018).
\newblock {MoDL: Model-based deep learning architecture for inverse problems}.
\newblock {\em IEEE transactions on medical imaging}, 38(2):394--405.

\bibitem[Aja-Fern{\'a}ndez and Vegas-S{\'a}nchez-Ferrero,
  2016]{aja2016statistical}
Aja-Fern{\'a}ndez, S. and Vegas-S{\'a}nchez-Ferrero, G. (2016).
\newblock {Statistical analysis of noise in MRI}.
\newblock {\em Switzerland: Springer International Publishing}.

\bibitem[Arridge et~al., 2019]{arridge2019solving}
Arridge, S., Maass, P., {\"O}ktem, O., and Sch{\"o}nlieb, C.-B. (2019).
\newblock {Solving inverse problems using data-driven models}.
\newblock {\em Acta Numerica}, 28:1--174.

\bibitem[Beck and Teboulle, 2009]{ISTAfirst}
Beck, A. and Teboulle, M. (2009).
\newblock {A Fast Iterative Shrinkage-Thresholding Algorithm for Linear Inverse
  Problems}.
\newblock {\em SIAM Journal on Imaging Sciences}, 2(1):183--202.

\bibitem[Becker et~al., 2011]{becker2011templates}
Becker, S.~R., Cand{\`e}s, E.~J., and Grant, M.~C. (2011).
\newblock {Templates for convex cone problems with applications to sparse
  signal recovery}.
\newblock {\em Mathematical programming computation}, 3(3):165--218.

\bibitem[Bellec and Tan, 2024]{bellec2024uncertainty}
Bellec, P.~C. and Tan, K. (2024).
\newblock {Uncertainty quantification for iterative algorithms in linear models
  with application to early stopping}.
\newblock {\em arXiv preprint arXiv:2404.17856}.

\bibitem[Bellec and Zhang, 2022]{bellec2022biasing}
Bellec, P.~C. and Zhang, C.-H. (2022).
\newblock {De-biasing the lasso with degrees-of-freedom adjustment}.
\newblock {\em Bernoulli}, 28(2):713--743.

\bibitem[Bellec and Zhang, 2023]{bellec2023debiasing}
Bellec, P.~C. and Zhang, C.-H. (2023).
\newblock {Debiasing convex regularized estimators and interval estimation in
  linear models}.
\newblock {\em The Annals of Statistics}, 51(2):391--436.

\bibitem[Belloni et~al., 2011]{belloni2011square}
Belloni, A., Chernozhukov, V., and Wang, L. (2011).
\newblock Square-root lasso: pivotal recovery of sparse signals via conic
  programming.
\newblock {\em Biometrika}, 98(4):791--806.

\bibitem[Bogdan et~al., 2015]{bogdan2015slope}
Bogdan, M., Van Den~Berg, E., Sabatti, C., Su, W., and Cand{\`e}s, E. (2015).
\newblock {SLOPE--Adaptive Variable Selection via Convex Optimization}.
\newblock {\em The Annals of Applied Statistics}, 9(65):1103--1140.

\bibitem[Bunea et~al., 2007]{bunea2007sparsity}
Bunea, F., Tsybakov, A., and Wegkamp, M. (2007).
\newblock {Sparsity oracle inequalities for the Lasso}.
\newblock {\em Electronic Journal of Statistics}, 1:169--194.

\bibitem[Cai and Guo, 2017]{cai2017confidence}
Cai, T.~T. and Guo, Z. (2017).
\newblock {Confidence Intervals for High-Dimensional Linear Regression: Minimax
  Rates and Adaptivity}.
\newblock {\em The Annals of Statistics}, pages 615--646.

\bibitem[Chen and Donoho, 1995]{chen1995examples}
Chen, S. and Donoho, D.~L. (1995).
\newblock {Examples of basis pursuit}.
\newblock In {\em Wavelet Applications in Signal and Image Processing III},
  volume 2569, pages 564--574. SPIE.

\bibitem[Chen et~al., 2022]{chen2022learning}
Chen, T., Chen, X., Chen, W., Heaton, H., Liu, J., Wang, Z., and Yin, W.
  (2022).
\newblock {Learning to optimize: A primer and a benchmark}.
\newblock {\em Journal of Machine Learning Research}, 23(189):1--59.

\bibitem[Chen et~al., 2021]{chen2021hyperparameter}
Chen, X., Liu, J., Wang, Z., and Yin, W. (2021).
\newblock {Hyperparameter tuning is all you need for LISTA}.
\newblock {\em Advances in Neural Information Processing Systems},
  34:11678--11689.

\bibitem[Cherkaoui et~al., 2020]{cherkaoui2020learning}
Cherkaoui, H., Sulam, J., and Moreau, T. (2020).
\newblock {Learning to solve TV regularised problems with unrolled algorithms}.
\newblock {\em Advances in Neural Information Processing Systems},
  33:11513--11524.

\bibitem[Dalalyan et~al., 2017]{dalalyan2017prediction}
Dalalyan, A., Hebiri, M., and Lederer, J.~C. (2017).
\newblock {On the prediction performance of the Lasso}.
\newblock {\em Bernoulli}, 23(1):552--581.

\bibitem[Daubechies et~al., 2004]{daubechies2004iterative}
Daubechies, I., Defrise, M., and De~Mol, C. (2004).
\newblock {An iterative thresholding algorithm for linear inverse problems with
  a sparsity constraint}.
\newblock {\em Communications on Pure and Applied Mathematics: A Journal Issued
  by the Courant Institute of Mathematical Sciences}, 57(11):1413--1457.

\bibitem[Dicker, 2014]{dicker2014variance}
Dicker, L.~H. (2014).
\newblock {Variance estimation in high-dimensional linear models}.
\newblock {\em Biometrika}, 101(2):269--284.

\bibitem[Ekmekci and Cetin, 2022]{ekmekci2022uncertainty}
Ekmekci, C. and Cetin, M. (2022).
\newblock {Uncertainty quantification for deep unrolling-based computational
  imaging}.
\newblock {\em IEEE Transactions on Computational Imaging}, 8:1195--1209.

\bibitem[Fan and Li, 2001]{fan2001variable}
Fan, J. and Li, R. (2001).
\newblock {Variable selection via nonconcave penalized likelihood and its
  oracle properties}.
\newblock {\em Journal of the American statistical Association},
  96(456):1348--1360.

\bibitem[Foucart and Rauhut, 2013]{Foucart.2013}
Foucart, S. and Rauhut, H. (2013).
\newblock {\em {A Mathematical Introduction to Compressive Sensing}}.
\newblock {Springer New York}, New York, NY.

\bibitem[Foucart et~al., 2023]{foucart2023sparsity}
Foucart, S., Tadmor, E., and Zhong, M. (2023).
\newblock {On the sparsity of LASSO minimizers in sparse data recovery}.
\newblock {\em Constructive Approximation}, 57(2):901--919.

\bibitem[Gal and Ghahramani, 2016]{gal2016dropout}
Gal, Y. and Ghahramani, Z. (2016).
\newblock {Dropout as a bayesian approximation: Representing model uncertainty
  in deep learning}.
\newblock In {\em international conference on machine learning}, pages
  1050--1059. PMLR.

\bibitem[Gawlikowski et~al., 2023]{gawlikowski2023survey}
Gawlikowski, J., Tassi, C. R.~N., Ali, M., Lee, J., Humt, M., Feng, J., Kruspe,
  A., Triebel, R., Jung, P., Roscher, R., et~al. (2023).
\newblock {A survey of uncertainty in deep neural networks}.
\newblock {\em Artificial Intelligence Review}, 56(Suppl 1):1513--1589.

\bibitem[Genzel et~al., 2022a]{genzel2022near}
Genzel, M., G{\"u}hring, I., Macdonald, J., and M{\"a}rz, M. (2022a).
\newblock {Near-exact recovery for tomographic inverse problems via deep
  learning}.
\newblock In {\em International Conference on Machine Learning}, pages
  7368--7381. PMLR.

\bibitem[Genzel et~al., 2022b]{genzel2022solving}
Genzel, M., Macdonald, J., and M{\"a}rz, M. (2022b).
\newblock {Solving inverse problems with deep neural networks--robustness
  included?}
\newblock {\em IEEE transactions on pattern analysis and machine intelligence},
  45(1):1119--1134.

\bibitem[Giraud, 2021]{giraud2021introduction}
Giraud, C. (2021).
\newblock {\em {Introduction to high-dimensional statistics}}.
\newblock Chapman and Hall/CRC.

\bibitem[Giraud et~al., 2012]{giraud2012high}
Giraud, C., Huet, S., and Verzelen, N. (2012).
\newblock {High-dimensional regression with unknown variance}.
\newblock {\em Statist. Sci.}, 27(4):500--518.

\bibitem[Gregor and LeCun, 2010]{gregor2010learning}
Gregor, K. and LeCun, Y. (2010).
\newblock {Learning fast approximations of sparse coding}.
\newblock In {\em Proceedings of the 27th international conference on machine
  learning}, pages 399--406.

\bibitem[Guo et~al., 2022]{guo2022doubly}
Guo, Z., {\'C}evid, D., and B{\"u}hlmann, P. (2022).
\newblock {Doubly debiased lasso: High-dimensional inference under hidden
  confounding}.
\newblock {\em Annals of statistics}, 50(3):1320.

\bibitem[Hastie et~al., 2009]{hastie2009elements}
Hastie, T., Tibshirani, R., Friedman, J.~H., and Friedman, J.~H. (2009).
\newblock {\em {The elements of statistical learning: data mining, inference,
  and prediction}}, volume~2.
\newblock Springer.

\bibitem[Hoppe et~al., 2022]{hoppe2022uncertainty}
Hoppe, F., Krahmer, F., Mayrink~Verdun, C., Menzel, M.~I., and Rauhut, H.
  (2022).
\newblock {Uncertainty quantification for sparse Fourier recovery}.
\newblock {\em arXiv preprint arXiv:2212.14864}.

\bibitem[Hoppe et~al., 2023a]{hoppe2023high}
Hoppe, F., Krahmer, F., Mayrink~Verdun, C., Menzel, M.~I., and Rauhut, H.
  (2023a).
\newblock {High-Dimensional Confidence Regions in Sparse MRI}.
\newblock In {\em ICASSP 2023-2023 IEEE International Conference on Acoustics,
  Speech and Signal Processing (ICASSP)}, pages 1--5. IEEE.

\bibitem[Hoppe et~al., 2024]{hoppe2024tv}
Hoppe, F., Mayrink~Verdun, C., Laus, H., Endt, S., Menzel, M.~I., Krahmer, F.,
  and Rauhut, H. (2024).
\newblock {Imaging with Confidence: Uncertainty Quantification for
  High-dimensional Undersampled MR Images}.
\newblock In {\em Proceedings of European Conference on Computer Vision
  (ECCV)}.

\bibitem[Hoppe et~al., 2023b]{hoppe2023uncertainty}
Hoppe, F., Mayrink~Verdun, C., Laus, H., Krahmer, F., and Rauhut, H. (2023b).
\newblock {Uncertainty quantification for learned ISTA}.
\newblock In {\em 2023 IEEE 33rd International Workshop on Machine Learning for
  Signal Processing (MLSP)}, pages 1--6. IEEE.

\bibitem[Ito et~al., 2019]{ito2019trainable}
Ito, D., Takabe, S., and Wadayama, T. (2019).
\newblock {Trainable ISTA for sparse signal recovery}.
\newblock {\em IEEE Transactions on Signal Processing}, 67(12):3113--3125.

\bibitem[Javanmard and Montanari, 2014]{Javanmard.2014}
Javanmard, A. and Montanari, A. (2014).
\newblock {Confidence intervals and hypothesis testing for high-dimensional
  regression}.
\newblock {\em Journal of Machine Learning Research}, 15:2869--2909.

\bibitem[Javanmard and Montanari, 2018]{Javanmard.2018}
Javanmard, A. and Montanari, A. (2018).
\newblock {Debiasing the lasso: Optimal sample size for {Gaussian} designs}.
\newblock {\em The Annals of Statistics}, 46(6A).

\bibitem[Kennedy and Ward, 2020]{kennedy2020greedy}
Kennedy, C. and Ward, R. (2020).
\newblock {Greedy variance estimation for the LASSO}.
\newblock {\em Appl. Math. Optim.}, 82(3):1161--1182.

\bibitem[Knight and Fu, 2000]{Fu2000}
Knight, K. and Fu, W. (2000).
\newblock {Asymptotics for Lasso-Type Estimators}.
\newblock {\em The Annals of Statistics}, 28(5):1356--1378.

\bibitem[Knoll et~al., 2020]{fastMRIdataset}
Knoll, F., Zbontar, J., Sriram, A., Muckley, M.~J., Bruno, M., Defazio, A.,
  Parente, M., Geras, K.~J., Katsnelson, J., Chandarana, H., Zhang, Z.,
  Drozdzalv, M., Romero, A., Rabbat, M., Vincent, P., Pinkerton, J., Wang, D.,
  Yakubova, N., Owens, E., Zitnick, C.~L., Recht, M.~P., Sodickson, D.~K., and
  Lui, Y.~W. (2020).
\newblock {fastMRI: A Publicly Available Raw k-Space and DICOM Dataset of Knee
  Images for Accelerated MR Image Reconstruction Using Machine Learning}.
\newblock {\em Radiology: Artificial Intelligence}, 2(1):e190007.
\newblock PMID: 32076662.

\bibitem[Koltchinskii, 2009]{koltchinskii2009sparsity}
Koltchinskii, V. (2009).
\newblock {Sparsity in penalized empirical risk minimization}.
\newblock In {\em Annales de l'IHP Probabilit{\'e}s et statistiques},
  volume~45, pages 7--57.

\bibitem[Koltchinskii et~al., 2011]{lassoNuclearNorm}
Koltchinskii, V., Lounici, K., and Tsybakov, A.~B. (2011).
\newblock {Nuclear-norm penalization and optimal rates for noisy low-rank
  matrix completion}.
\newblock {\em The Annals of Statistics}, 39(5):2302 -- 2329.

\bibitem[K{\"u}mmerle et~al., 2021]{kummerle2021iteratively}
K{\"u}mmerle, C., Mayrink~Verdun, C., and St{\"o}ger, D. (2021).
\newblock {Iteratively reweighted least squares for basis pursuit with global
  linear convergence rate}.
\newblock {\em Advances in Neural Information Processing Systems},
  34:2873--2886.

\bibitem[Li and Malik, 2016]{li2016learning}
Li, K. and Malik, J. (2016).
\newblock {Learning to Optimize}.
\newblock In {\em International Conference on Learning Representations}.

\bibitem[Li, 2020]{li2020debiasing}
Li, S. (2020).
\newblock {Debiasing the debiased Lasso with bootstrap}.
\newblock {\em Electronic Journal of Statistics}, 14:2298--2337.

\bibitem[Li et~al., 2018]{li2018highly}
Li, X., Sun, D., and Toh, K.-C. (2018).
\newblock {A highly efficient semismooth Newton augmented Lagrangian method for
  solving Lasso problems}.
\newblock {\em SIAM Journal on Optimization}, 28(1):433--458.

\bibitem[Lin and Heckel, 2022]{lin2022vision}
Lin, K. and Heckel, R. (2022).
\newblock Vision transformers enable fast and robust accelerated mri.
\newblock In {\em International Conference on Medical Imaging with Deep
  Learning}, pages 774--795. PMLR.

\bibitem[Liu and Chen, 2019]{liu2019alista}
Liu, J. and Chen, X. (2019).
\newblock {ALISTA: Analytic weights are as good as learned weights in LISTA}.
\newblock In {\em International Conference on Learning Representations (ICLR)}.

\bibitem[Liu et~al., 2019]{liu2019deep}
Liu, R., Cheng, S., Ma, L., Fan, X., and Luo, Z. (2019).
\newblock {Deep proximal unrolling: Algorithmic framework, convergence analysis
  and applications}.
\newblock {\em IEEE Transactions on Image Processing}, 28(10):5013--5026.

\bibitem[Liu et~al., 2020]{liu2020estimation}
Liu, X., Zheng, S., and Feng, X. (2020).
\newblock {Estimation of error variance via ridge regression}.
\newblock {\em Biometrika}, 107(2):481--488.

\bibitem[Mayrink~Verdun et~al., 2024]{verdun2024fast}
Mayrink~Verdun, C., Melnyk, O., Krahmer, F., and Jung, P. (2024).
\newblock Fast, blind, and accurate: Tuning-free sparse regression with global
  linear convergence.
\newblock In {\em The Thirty Seventh Annual Conference on Learning Theory},
  pages 3823--3872. PMLR.

\bibitem[Monga et~al., 2021]{monga2021algorithm}
Monga, V., Li, Y., and Eldar, Y.~C. (2021).
\newblock {Algorithm unrolling: Interpretable, efficient deep learning for
  signal and image processing}.
\newblock {\em IEEE Signal Processing Magazine}, 38(2):18--44.

\bibitem[Rakotomamonjy et~al., 2022]{rakotomamonjy2022convergent}
Rakotomamonjy, A., Flamary, R., Salmon, J., and Gasso, G. (2022).
\newblock {Convergent working set algorithm for lasso with non-convex sparse
  regularizers}.
\newblock In {\em International Conference on Artificial Intelligence and
  Statistics}, pages 5196--5211. PMLR.

\bibitem[Raskutti et~al., 2011]{raskutti2011minimax}
Raskutti, G., Wainwright, M.~J., and Yu, B. (2011).
\newblock {Minimax rates of estimation for high-dimensional linear regression
  over $\ell_q$-balls}.
\newblock {\em IEEE transactions on information theory}, 57(10):6976--6994.

\bibitem[Reid et~al., 2016]{reid2016study}
Reid, S., Tibshirani, R., and Friedman, J. (2016).
\newblock {A study of error variance estimation in lasso regression}.
\newblock {\em Statist. Sinica}, pages 35--67.

\bibitem[Ronneberger et~al., 2015]{ronneberger2015u}
Ronneberger, O., Fischer, P., and Brox, T. (2015).
\newblock {U-net: Convolutional networks for biomedical image segmentation}.
\newblock In {\em Medical image computing and computer-assisted
  intervention--MICCAI 2015: 18th international conference, Munich, Germany,
  October 5-9, 2015, proceedings, part III 18}, pages 234--241. Springer.

\bibitem[Saw et~al., 1984]{Saw1984}
Saw, J.~G., Yang, M. C.~K., and Mo, T.~C. (1984).
\newblock {Chebyshev Inequality with Estimated Mean and Variance}.
\newblock {\em The American Statistician}, 38(2):130--132.

\bibitem[Schlemper et~al., 2017]{schlemper2017deep}
Schlemper, J., Caballero, J., Hajnal, J.~V., Price, A.~N., and Rueckert, D.
  (2017).
\newblock {A deep cascade of convolutional neural networks for dynamic MR image
  reconstruction}.
\newblock {\em IEEE transactions on Medical Imaging}, 37(2):491--503.

\bibitem[Sidky and Pan, 2022]{sidky2022report}
Sidky, E.~Y. and Pan, X. (2022).
\newblock {Report on the AAPM deep-learning sparse-view CT grand challenge}.
\newblock {\em Medical physics}, 49(8):4935--4943.

\bibitem[Stellato et~al., 2017]{Stellato.2017}
Stellato, B., {van Parys}, B. P.~G., and Goulart, P.~J. (2017).
\newblock {Multivariate Chebyshev Inequality With Estimated Mean and Variance}.
\newblock {\em {The American Statistician}}, 71(2):123--127.

\bibitem[Sun and Zhang, 2012]{sun2012scaled}
Sun, T. and Zhang, C.-H. (2012).
\newblock {Scaled sparse linear regression}.
\newblock {\em Biometrika}, 99(4):879--898.

\bibitem[Tibshirani, 1996]{tibshirani1996regression}
Tibshirani, R. (1996).
\newblock {Regression shrinkage and selection via the lasso}.
\newblock {\em Journal of the Royal Statistical Society Series B: Statistical
  Methodology}, 58(1):267--288.

\bibitem[{van de Geer} et~al., 2014]{vandeGeer.2014}
{van de Geer}, S., B{\"u}hlmann, P., Ritov, Y., and Dezeure, R. (2014).
\newblock {On asymptotically optimal confidence regions and tests for
  high-dimensional models}.
\newblock {\em The Annals of Statistics}, 42(3).

\bibitem[Wainwright, 2009]{wainwright2009sharp}
Wainwright, M.~J. (2009).
\newblock {Sharp thresholds for High-Dimensional and noisy sparsity recovery
  using $\ell_1$-Constrained Quadratic Programming (Lasso)}.
\newblock {\em IEEE transactions on information theory}, 55(5):2183--2202.

\bibitem[Wainwright, 2019]{wainwright2019high}
Wainwright, M.~J. (2019).
\newblock {\em {High-dimensional statistics: A non-asymptotic viewpoint}},
  volume~48.
\newblock Cambridge university press.

\bibitem[Wang et~al., 2003]{wang2003multiscale}
Wang, Z., Simoncelli, E.~P., and Bovik, A.~C. (2003).
\newblock {Multiscale structural similarity for image quality assessment}.
\newblock In {\em The Thirty-Seventh Asilomar Conference on Signals, Systems \&
  Computers, 2003}, volume~2, pages 1398--1402. Ieee.

\bibitem[Wright and Ma, 2022]{wright2022high}
Wright, J. and Ma, Y. (2022).
\newblock {\em {High-dimensional data analysis with low-dimensional models:
  Principles, computation, and applications}}.
\newblock Cambridge University Press.

\bibitem[Wu et~al., 2019]{wu2019sparse}
Wu, K., Guo, Y., Li, Z., and Zhang, C. (2019).
\newblock {Sparse coding with gated learned ISTA}.
\newblock In {\em International conference on learning representations}.

\bibitem[Ye and Zhang, 2010]{ye2010rate}
Ye, F. and Zhang, C.-H. (2010).
\newblock {Rate minimaxity of the Lasso and Dantzig selector for the $\ell_q$
  loss in $\ell_r$ balls}.
\newblock {\em The Journal of Machine Learning Research}, 11:3519--3540.

\bibitem[Yu and Bien, 2019]{yu2019estimating}
Yu, G. and Bien, J. (2019).
\newblock {Estimating the error variance in a high-dimensional linear model}.
\newblock {\em Biometrika}, 106(3):533--546.

\bibitem[Yuan and Lin, 2005]{groupLasso}
Yuan, M. and Lin, Y. (2005).
\newblock {Model Selection and Estimation in Regression with Grouped
  Variables}.
\newblock {\em Journal of the Royal Statistical Society Series B: Statistical
  Methodology}, 68(1):49--67.

\bibitem[Zbontar et~al., 2019]{zbontar2019fastmri}
Zbontar, J., Knoll, F., Sriram, A., Murrell, T., Huang, Z., Muckley, M.~J.,
  Defazio, A., Stern, R., Johnson, P., Bruno, M., Parente, M., Geras, K.~J.,
  Katsnelson, J., Chandarana, H., Zhang, Z., Drozdzal, M., Romero, A., Rabbat,
  M., Vincent, P., Yakubova, N., Pinkerton, J., Wang, D., Owens, E., Zitnick,
  C.~L., Recht, M.~P., Sodickson, D.~K., and Lui, Y.~W. (2019).
\newblock {fastMRI: An Open Dataset and Benchmarks for Accelerated MRI}.

\bibitem[Zhang, 2010]{zhang2010nearly}
Zhang, C.~H. (2010).
\newblock {Nearly unbiased variable selection under minimax concave penalty}.
\newblock {\em Annals of Statistics}, 38(2):894--942.

\bibitem[Zhang and Huang, 2008]{zhang2008sparsity}
Zhang, C.~H. and Huang, J. (2008).
\newblock {The sparsity and bias of the lasso selection in high-dimensional
  linear regression}.
\newblock {\em Annals of Statistics}, 36(4):1567--1594.

\bibitem[Zhang and Zhang, 2014]{Zhang.2014}
Zhang, C.-H. and Zhang, S.~S. (2014).
\newblock {Confidence intervals for low dimensional parameters in high
  dimensional linear models}.
\newblock {\em Journal of the Royal Statistical Society: Series B (Statistical
  Methodology)}, 76(1):217--242.

\bibitem[Zhang et~al., 2023]{zhang2023physics}
Zhang, J., Chen, B., Xiong, R., and Zhang, Y. (2023).
\newblock {Physics-inspired compressive sensing: Beyond deep unrolling}.
\newblock {\em IEEE Signal Processing Magazine}, 40(1):58--72.

\bibitem[Zhao and Yu, 2006]{zhao2006model}
Zhao, P. and Yu, B. (2006).
\newblock {On model selection consistency of Lasso}.
\newblock {\em The Journal of Machine Learning Research}, 7:2541--2563.

\bibitem[Zheng et~al., 2017]{zheng2017does}
Zheng, L., Maleki, A., Weng, H., Wang, X., and Long, T. (2017).
\newblock {Does $\ell_p$-Minimization Outperform $\ell_1$-Minimization?}
\newblock {\em IEEE Transactions on Information Theory}, 63(11):6896--6935.

\bibitem[Zheng et~al., 2022]{zheng2022hybrid}
Zheng, Z., Dai, W., Xue, D., Li, C., Zou, J., and Xiong, H. (2022).
\newblock {Hybrid ISTA: Unfolding ISTA with convergence guarantees using
  free-form deep neural networks}.
\newblock {\em IEEE Transactions on Pattern Analysis and Machine Intelligence},
  45(3):3226--3244.

\bibitem[Zou and Hastie, 2005]{zou2005regularization}
Zou, H. and Hastie, T. (2005).
\newblock {Regularization and variable selection via the elastic net}.
\newblock {\em Journal of the Royal Statistical Society Series B: Statistical
  Methodology}, 67(2):301--320.

\end{thebibliography}

\newpage

\appendix

\textbf{Supplementary material to the paper \textit{Non-Asymptotic Uncertainty Quantification in
High-Dimensional Learning}.}

In this supplement to the paper, we present in Section \ref{sec:discussion} a detailed discussion about aspects of the main result that are not mentioned in the main body of the paper. Moreover, Section \ref{sec:proof_main} presents the proof Theorem \ref{thm:main_stat_result} and Theorem \ref{thm:remainder_dist_gaussian}. The former establishes data-driven confidence intervals, while the latter assumes the remainder component to be approximated by a Gaussian distribution. In Section \ref{sec:further_numerical}, we confirm our theoretical findings with several numerical experiments for classical high-dimensional regression as well as model-based neural networks. In Section \ref{sec:gaussian_remainder_density}, we visualize the approximate Gaussian distribution of the remainder terms, demonstrating the applicability of Theorem \ref{thm:remainder_dist_gaussian} in relevant settings.

\section{Further Discussion of Main Result}\label{sec:discussion}

\textbf{Length of radius:} To minimize the length of the radius, $\gamma_j\in\left(0,1-\frac{1}{l\alpha}\right)$ should be chosen as the minimizer of the problem
\begin{equation}\label{eq:min_lengh_CI}
    \min\limits_{\gamma_j\in\left(0,1-\frac{1}{l\alpha}\right)} \frac{\sigma (M\hat{\Sigma}M^*)_{jj}^{1/2}}{\sqrt{m}}\sqrt{\log\left(\frac{1}{\gamma_j\alpha}\right)} + \sqrt{\frac{l^2-1}{l^2(1-\gamma_j)\alpha - l }}\cdot (\hat{\sigma}_R)_j,
\end{equation}
In order to minimize over a large set for a given significance level $\alpha$, a large number of data $l$ is needed. For fixed estimates $\hat{S}_j$ and $(\hat{\sigma}_R^2)_j$, more data leads to a potentially smaller confidence interval length.
If we assume $R_j=0$, it follows that $\hat{S}_j = 0$ and $(\hat{\sigma}_R^2)_j = 0$. Then, $\gamma = 1$ is a valid choice, for which the function $\frac{\sigma (M\hat{\Sigma}M^*)_{jj}^{1/2}}{\sqrt{m}}\sqrt{\log\left(\frac{1}{\gamma_j\alpha}\right)}$ is well-defined and is minimized. In this case, the radius coincides with the asymptotic radius derived in \cite{Javanmard.2014, Javanmard.2018, vandeGeer.2014} (except for that these works handle the real case) and the ones in \cite{hoppe2023uncertainty} with $M=I_{N\times N}$. In this sense, the asymptotic confidence intervals can be seen as a special case of the proposed method.

The significance level $\alpha$ depends on $\gamma_j$ and $l$ to assure $c_l(\,\cdot\,)$ to be well-defined. For a large dataset $x^{(1)},\hdots,x^{(l)}$, i.e. if $l$ is large, then it holds that $    \lim\limits_{l\to\infty} c_l(\alpha) = \lim\limits_{l\to\infty}\sqrt{\frac{1-\frac{1}{l^2}}{(1-\gamma_j)\alpha - \frac{1}{l}}} = \frac{1}{\sqrt{(1-\gamma_j)\alpha}}$.

\textbf{Probabilistic discussion:} The probability in \eqref{eq:prob_CI} is over the randomness of the noise as well as $\mathbb{Q}$. The confidence circles $C_j(\alpha)$ consist of two random variables, the debiased estimator $\hat{x}^u_j$ and the radius $r_j(\alpha)$. The former depends on the random noise and potentially on training data, while the latter depends on the estimators $\hat{S}_j$ and $\hat{\sigma}_R$, which in turn depend on both the noise and the data $x^{(1)},\hdots, x^{(l)}$.

A crucial requirement of applying the empirical version of Chebyshev's inequality \cite{Saw1984} is the independence and identical distribution of the variables $\vert R_j^{(1)}\vert,\hdots,\vert R_j^{(l)}\vert$. Therefore, it is essential that the estimator function $\hat{X}$ is independent of the data $x^{(1)},\hdots,x^{(l)}$. To achieve this, we train the estimator function $\hat{X}$ using a dataset that is independent of the data $x^{(1)},\hdots,x^{(l)}$, used for estimating $R^{(1)},\hdots, R^{(l)}$. However, the mean and variance of $\vert R^{(1)}\vert$ and hence of $\vert R^{(i)}\vert$ depend on the variance of the noise $\varepsilon^{(1)}$, i.e. $\sigma^2$. Thus, different noise levels $\sigma$ require a new estimation of the mean and variance of $\vert R^{(1)}\vert$. Throughout this paper, we assume the noise level to be fixed and known. The latter assumption is motivated by two factors. First, the size of the confidence intervals relies on $\sigma$. Given that the primary focus of this paper is to determine the size of the confidence intervals based on the remainder term $R^{(1)}$, we seek to mitigate other influencing factors such as noise level estimation. Second, in domains like medical imaging, there is substantial knowledge about the noise level. For instance, the noise level in MRI can be directly measured from the scanner \cite{aja2016statistical}. If the noise level is unknown, there are methods to estimate it. In the debiased LASSO literature, the most used method is the scaled LASSO \cite{sun2012scaled}. Other methods for sparse regression, either in the LASSO context or more general for high-dimensional models, are \cite{reid2016study, kennedy2020greedy, giraud2012high, dicker2014variance, liu2020estimation, yu2019estimating}.

\textbf{Relaxation of assumptions in practice:} In practice, it is often the case, that $\vert R^{(i)}_1\vert ,\hdots, \vert R^{(i)}_N\vert$ are identical distributed resulting in $\mu_1 = \hdots = \mu_N$ and $(\sigma_R^2)_1 = \hdots = (\sigma_R^2)_N$.  Although the proof requires independence of the $\vert R_j^{(i)}\vert$, there are cases when it might suffice to relax this assumption by estimating the mean and variance pixel-wise uniformly, i.e., \begin{equation}\label{eq:uniform_estimator}
    \hat{S} = \frac{1}{l\cdot N}\sum\limits_{i=1}^l\sum\limits_{j=1}^N R_j^{(i)} \qquad \text{and}\qquad \hat{\sigma}_R^2 = \frac{1}{l\cdot N -1} \sum\limits_{i=1}^l\sum\limits_{j=1}^N (R_j^{(i)} - \hat{S})^2.
\end{equation}
In addition to saving computational resources, accuracy improves due to the higher number of samples. Furthermore, instead of solving the optimization problem \eqref{eq:min_lengh_CI} for every $j\in\{1,\hdots, N\}$, it might be a good idea to choose $\gamma_1=\hdots=\gamma_N$ as the minimizer of 
\begin{equation}\label{eq:one_gamma}
    \min\limits_{\gamma_j\in\left(0,1-\frac{1}{l\alpha}\right)} \frac{\sigma\sum\limits_{j=1}^N(M \hat{\Sigma}M^*)_{jj}^{1/2}}{\sqrt{m}N}\sqrt{\log\left(\frac{1}{\gamma_j\alpha}\right)} + c_l\left((1-\gamma_j)\alpha\right)\cdot \frac{1}{N}\sum\limits_{j=1}^N(\hat{\sigma}_R)_j.
\end{equation}
Then, one $\gamma$ can be used for computing the potentially different radii $r_j(\alpha)$.

\section{Proofs}\label{sec:proof_main}

\begin{proof}[Proof of Theorem \ref{thm:main_stat_result}]
    The statement $x^{(l+1)}_j \in C_j(\alpha)$ is equivalent to $\vert (\hat{x}^u)^{(l+1)}_j - x^{(l+1)}_j\vert \leq r_j(\alpha)$. To prove \eqref{eq:prob_CI}, we show that
    \begin{equation*}
        \mathbb{P}\left(\vert (\hat{x}^u)^{(l+1)}_j - x^{(l+1)}_j\vert\geq r_j(\alpha)\right)\leq \alpha
    \end{equation*}
    In the next step, we write the radius $r(\alpha)$ as the sum $r(\alpha) = r^W(\alpha) + r^R(\alpha)$. According to the decomposition $(\hat{x}^u)^{(l+1)}_j - x^{(l+1)}_j = W_j + R_j$ we obtain for fixed $j\in\{1,\hdots,N\}$
    \begin{align*}
        &\mathbb{P}\left(\vert (\hat{x}^u_j)^{(l+1)} - x_j^{(l+1)}\vert \geq r^W_j(\alpha) + r^R_j(\alpha)\right) = \mathbb{P}\left(\vert W_j + R_j\vert \geq r^W_j(\alpha) + r^R_j(\alpha)\right)\\
        \leq &\mathbb{P}\left(\vert W_j\vert +\vert R_j\vert \geq r_j^W(\alpha) + r_j^R(\alpha)\right)        
        \leq \mathbb{P}\left( \vert W_j\vert \geq r^W_j(\alpha) \right) + \mathbb{P}\left(\vert R_j\vert \geq r^R_j(\alpha) \right)
    \end{align*}
    where the last step follows from the pigeonhole principle. To estimate the first summand, we set $r^W_j(\alpha) := \frac{\sigma(M\hat{\Sigma}M^*)_{jj}^{1/2}}{\sqrt{m}}\sqrt{\log\left(\frac{1}{\gamma_j\alpha}\right)}$. Since $\vert W_j\vert \sim \operatorname{Rice}\left(0,\frac{\sigma(M\hat{\Sigma}M^*)_{jj}^{1/2}}{\sqrt{2m}}\right)$ we obtain
    \begin{align*}
        \mathbb{P}\left( \vert W_j\vert \geq r^W_j(\alpha)\right)
        &= \frac{2m}{\sigma^2\hat{\Sigma}_{jj}}\int\limits_{r^W_j(\alpha)}^{\infty} x \exp\left( - \frac{x^2m}{\sigma^2(M\hat{\Sigma}M^*)_{jj}}\right) dx
        = \int\limits_{\frac{(r^W_j(\alpha))^2m}{\sigma^2(M\hat{\Sigma}M^*)_{jj}}} \exp(-u) du \\
        &= \exp\left( -\frac{(r^W_j(\alpha))^2m}{\sigma^2(M\hat{\Sigma}M^*)_{jj}} \right) = \exp\left( -\log(1/\gamma_j\alpha) \right) = \gamma_j\alpha.
    \end{align*}
    For estimating the term $\mathbb{P}\left(\vert R_j\vert \geq r^R_j(\alpha) \right)$, we set $r^R_j(\alpha) = c_l(\alpha)\cdot (\hat{\sigma}_R)_j + \hat{S}_j$.
    This choice leads to
    \begin{align*}
        \mathbb{P}\left(\vert R_j\vert \geq r^R_j(\alpha) \right) &= \mathbb{P}\left(\vert R_j\vert -  \hat{S}_j \geq r^R_j(\alpha) -  \hat{S}_j \right)
        \leq \mathbb{P}\left( \left\vert \vert R_j\vert -  \hat{S}_j \right\vert \geq r^R_j(\alpha) -  \hat{S}_j \right) \\
        & = \mathbb{P}\left( \frac{\left\vert \vert R_j\vert -  \hat{S}_j \right\vert}{(\hat{\sigma}_R)_j} \geq \frac{r^R_j(\alpha) -  \hat{S}_j}{(\hat{\sigma}_R)_j} \right)
        = \mathbb{P}\left( \frac{\left\vert \vert R_j\vert - \hat{S}_j \right\vert}{(\hat{\sigma}_R)_j} \geq c_l(\alpha) \right).
    \end{align*}
    Now, we apply an empirical version of Chebyshev's inequality \cite{Saw1984,Stellato.2017}. This leads to
    \begin{align*}
        \mathbb{P}\left( \frac{\left\vert \vert R_j\vert - \hat{S}_j \right\vert}{(\hat{\sigma}_R)_j} \geq c_l(\alpha) \right)
        &\leq \min \left\{ 1, \frac{1}{l+1} \left\lfloor \frac{(l+1)(l^2-1 + l c_l(\alpha)^2)}{l^2 c_l(\alpha)^2} \right\rfloor \right\} \\
        &\leq \min \left\{ 1, \frac{l^2-1 + l c_l(\alpha)^2}{l^2 c_l(\alpha)^2} \right\}
        = \min \left\{ 1, \frac{l^2-1 + \frac{l^2-1}{l(1-\gamma_j)\alpha - 1 } }{ \frac{l(l^2-1)}{l(1-\gamma_j)\alpha -1} } \right\} \\
        &= \min \left\{ 1, \frac{1 + \frac{1}{l(1-\gamma_j)\alpha - 1 } }{ \frac{l}{l(1-\gamma_j)\alpha -1} } \right\}
        = \min \left\{ 1, (1-\gamma_j)\alpha \right\} = (1-\gamma_j)\alpha,
    \end{align*}
    where we used in the last step, that $(1-\gamma_j)\alpha<\alpha<1$. To summarize,
    \begin{align*}
        \mathbb{P}\left(\vert (\hat{x}^u)^{(l+1)}_j - x^{(l+1)}_j\vert\geq r_j(\alpha)\right)
        & \leq \mathbb{P}\left( \vert W_j\vert \geq r_j^W(\alpha)\right) + \mathbb{P}\left(\vert R_j\vert \geq r_j^R(\alpha)\right) \\
        & \leq \gamma_j\alpha + (1-\gamma_j)\alpha = \alpha.
    \end{align*}
\end{proof}

\begin{proof}[Proof of Theorem \ref{thm:remainder_dist_gaussian}]
    Since $W\sim\mathcal{CN}(0,\frac{\sigma^2}{m}M\hat{\Sigma}M^*)$ and $R\sim\mathcal{CN}(0,\frac{1}{m}\Sigma_R)$ the estimation error $\hat{x}^u-x^0=W+R$ follows again a multivariate normal distribution with zero mean and covariance matrix $\frac{1}{m}(\sigma^2M\hat{\Sigma}M^* + \Sigma_R)$. By exploiting the Gaussian distribution, we obtain
    \begin{align*}
        \mathbb{P}\left( \vert W_j + R_j\vert >r^G_j(\alpha)\right)
        &= \frac{2m}{\sigma^2(M\hat{\Sigma}M^*)_{jj} + (\Sigma_R)_{jj}}\int\limits_{r^G_j(\alpha)}^{\infty} x \exp\left( - \frac{x^2m}{\sigma^2(M\hat{\Sigma}M^*)_{jj} + (\Sigma_R)_{jj}}\right) dx\\
        &= \int\limits_{\frac{r^G_j(\alpha)^2m}{\sigma^2(M\hat{\Sigma}M^*)_{jj} + (\Sigma_R)_{jj}}} \exp(-u) du
        = \exp\left( -\frac{r^G_j(\alpha)^2m}{\sigma^2(M\hat{\Sigma}M^*)_{jj}+(\Sigma_R)_{jj}} \right)
    \end{align*}
    Thus, we have
    \begin{equation*}
        \mathbb{P}(\vert \hat{x}^u_j-x_j^*\vert >r^G_j(\alpha))\leq \exp\left( -\frac{r^G_j(\alpha)^2m}{\sigma^2(M\hat{\Sigma}M^*)_{jj} + (\Sigma_R)_{jj}} \right),
    \end{equation*}
    which needs to be equal to $\alpha>0$. Therefore,
    \begin{equation*}
        r^G_j(\alpha) = \frac{(\sigma(M\hat{\Sigma}M_{jj}+(\Sigma_R)_{jj})^{1/2}}{\sqrt{m}}\sqrt{\log\left(\frac{1}{\alpha}\right)}.
    \end{equation*}
\end{proof}

\section{Further Numerical Evaluation}\label{sec:further_numerical}
To confirm our theoretical findings claiming that the incorporation of the bias component renders the confidence intervals more robust, we present additional numerical experiments here.

\begin{figure}[tb]
  \centering
  \begin{subfigure}[T]{0.16\linewidth}
    \includegraphics[width=\textwidth]{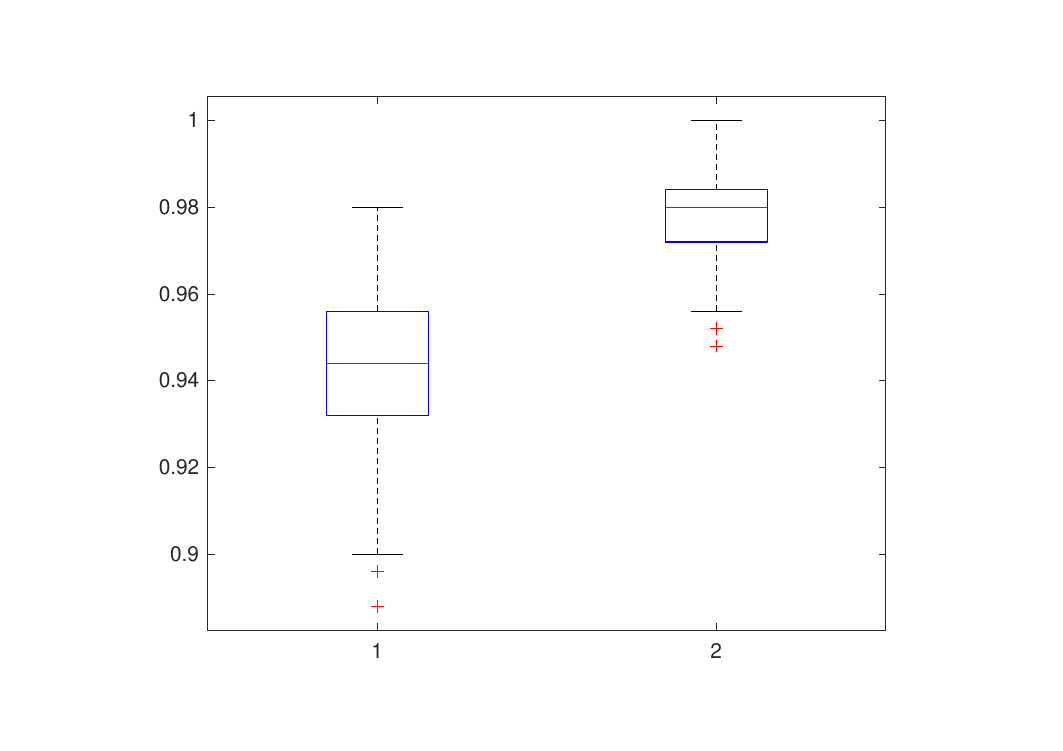}
    \caption{}
    \label{subfig:1}
  \end{subfigure}
  \hfill
  \begin{subfigure}[T]{0.16\linewidth}
    \includegraphics[width=\textwidth]{Gaussian_exp/2ada.pdf}
    \caption{}
    \label{subfig:2}
  \end{subfigure}
  \hfill
   \begin{subfigure}[T]{0.16\linewidth}
     \includegraphics[width=\textwidth]{Gaussian_exp/3ada.pdf}
    \caption{}
    \label{subfig:3}
  \end{subfigure}
  \begin{subfigure}[T]{0.16\linewidth}
     \includegraphics[width=\textwidth]{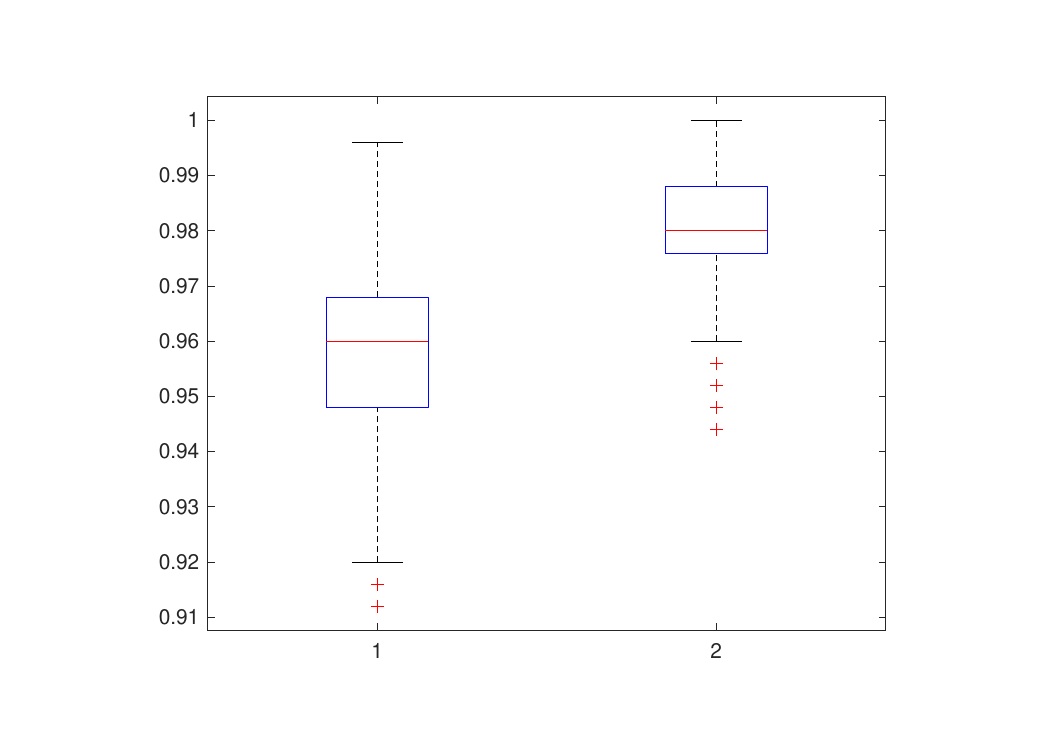}
    \caption{}
    \label{subfig:4}
  \end{subfigure}
  \begin{subfigure}[T]{0.16\linewidth}
    \includegraphics[width=\textwidth]{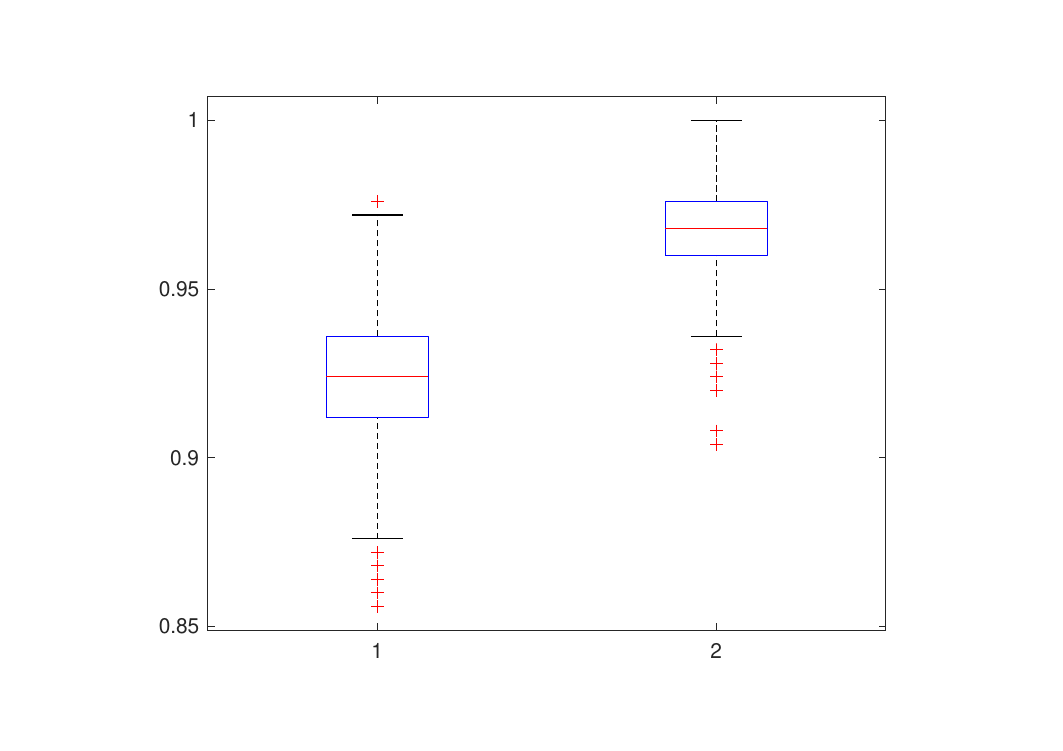}
    \caption{}
    \label{subfig:5}
  \end{subfigure}
  \hfill
  \begin{subfigure}[T]{0.16\linewidth}
    \includegraphics[width=\textwidth]{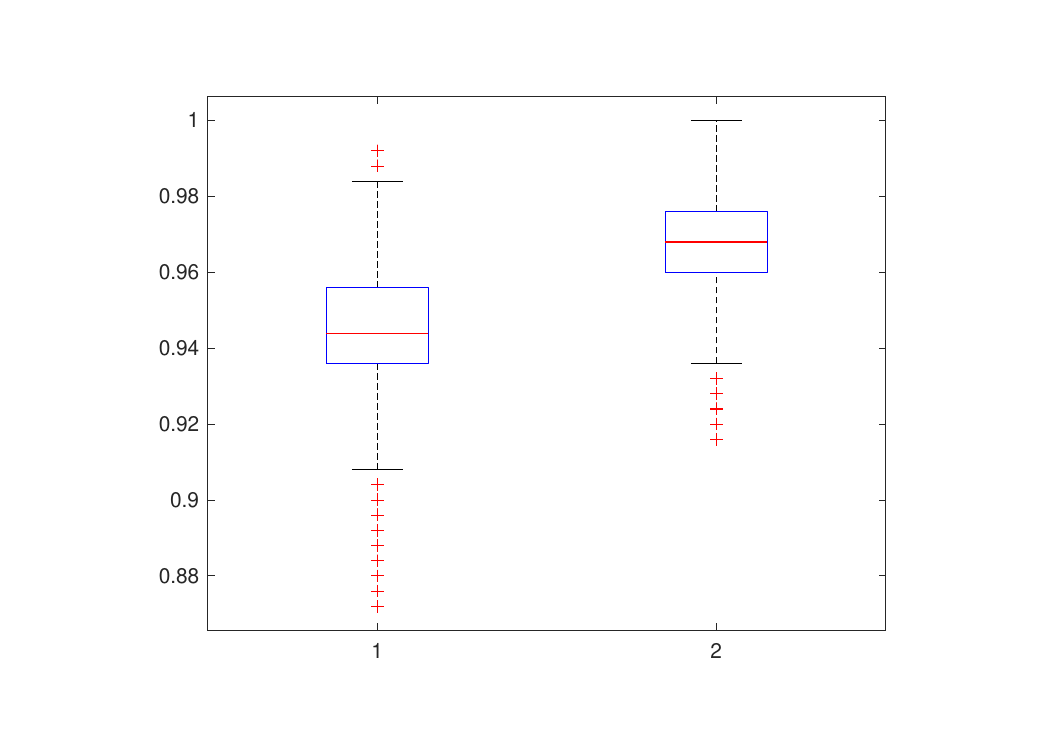}
    \caption{}
    \label{subfig:6}
  \end{subfigure}
  \hfill
   \begin{subfigure}[T]{0.16\linewidth}
     \includegraphics[width=\textwidth]{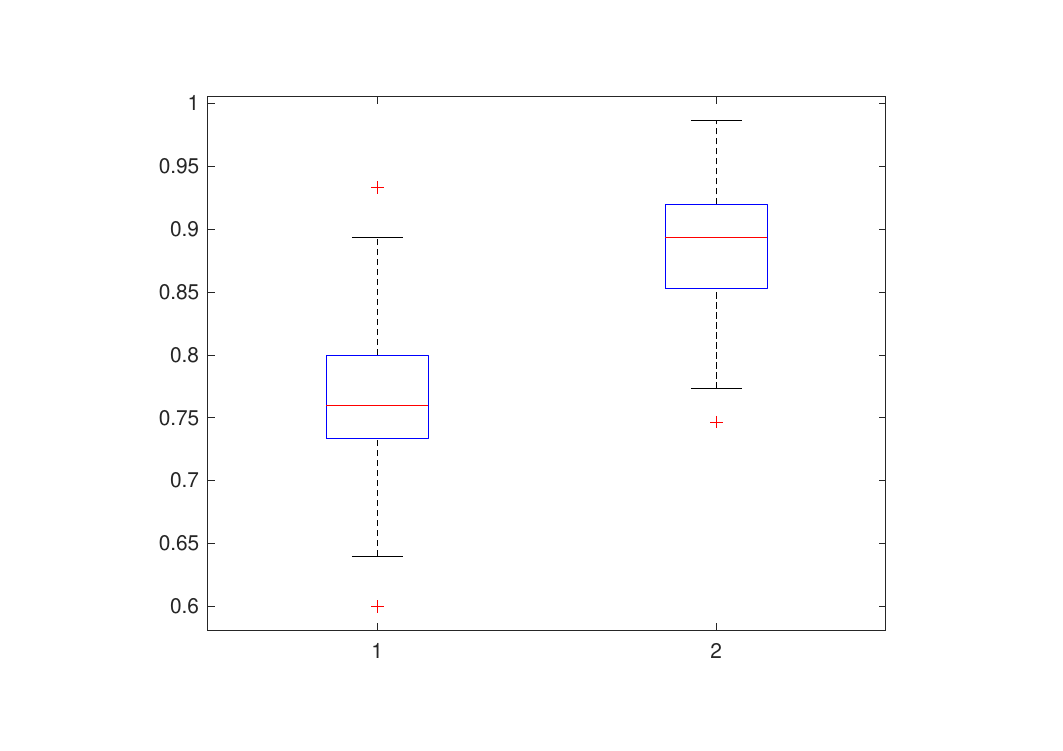}
    \caption{}
    \label{subfig:1s}
  \end{subfigure}
  \begin{subfigure}[T]{0.16\linewidth}
     \includegraphics[width=\textwidth]{Gaussian_exp/2sada.pdf}
    \caption{}
    \label{subfig:2s}
  \end{subfigure}
  \hfill
   \begin{subfigure}[T]{0.16\linewidth}
    \includegraphics[width=\textwidth]{Gaussian_exp/3sada.pdf}
    \caption{}
    \label{subfig:3s}
  \end{subfigure}
    \hfill
   \begin{subfigure}[T]{0.16\linewidth}
    \includegraphics[width=\textwidth]{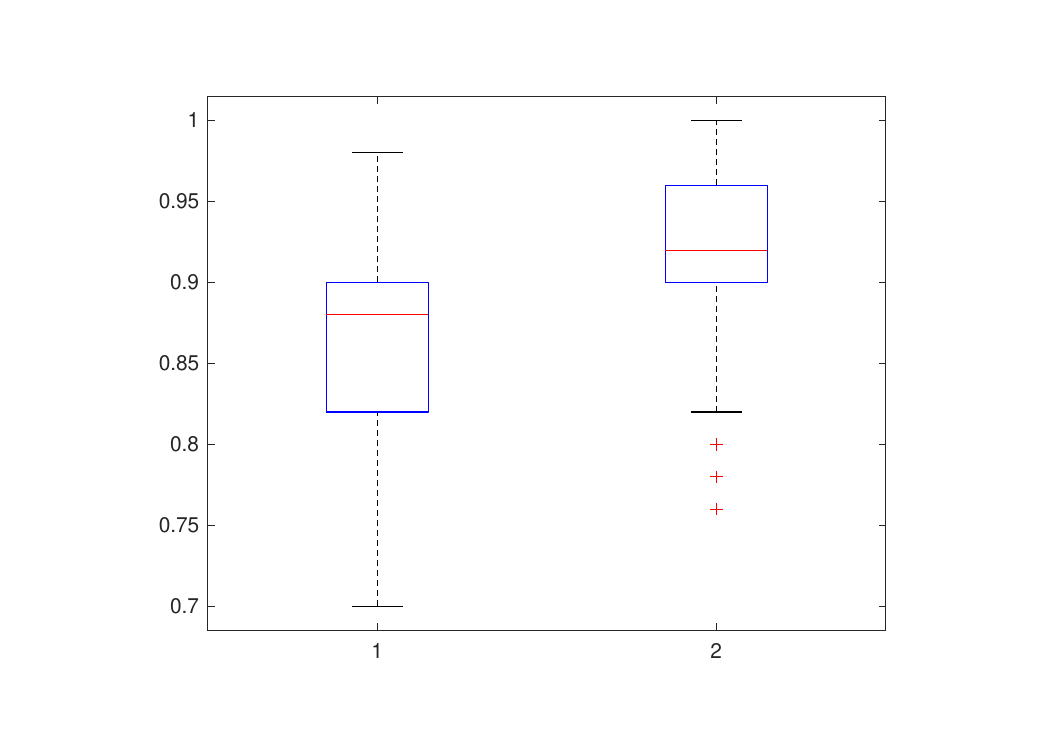}
    \caption{}
    \label{subfig:4s}
  \end{subfigure}
     \begin{subfigure}[T]{0.16\linewidth}
    \includegraphics[width=\textwidth]{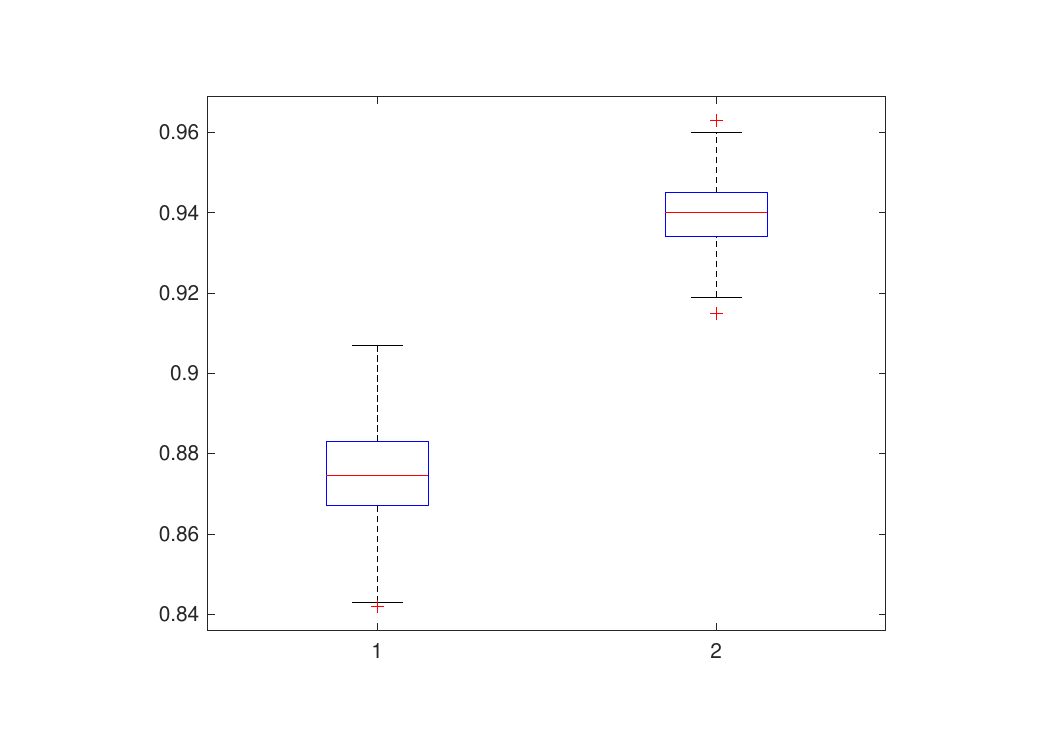}
    \caption{}
    \label{subfig:5s}
  \end{subfigure}
    \hfill
   \begin{subfigure}[T]{0.16\linewidth}
    \includegraphics[width=\textwidth]{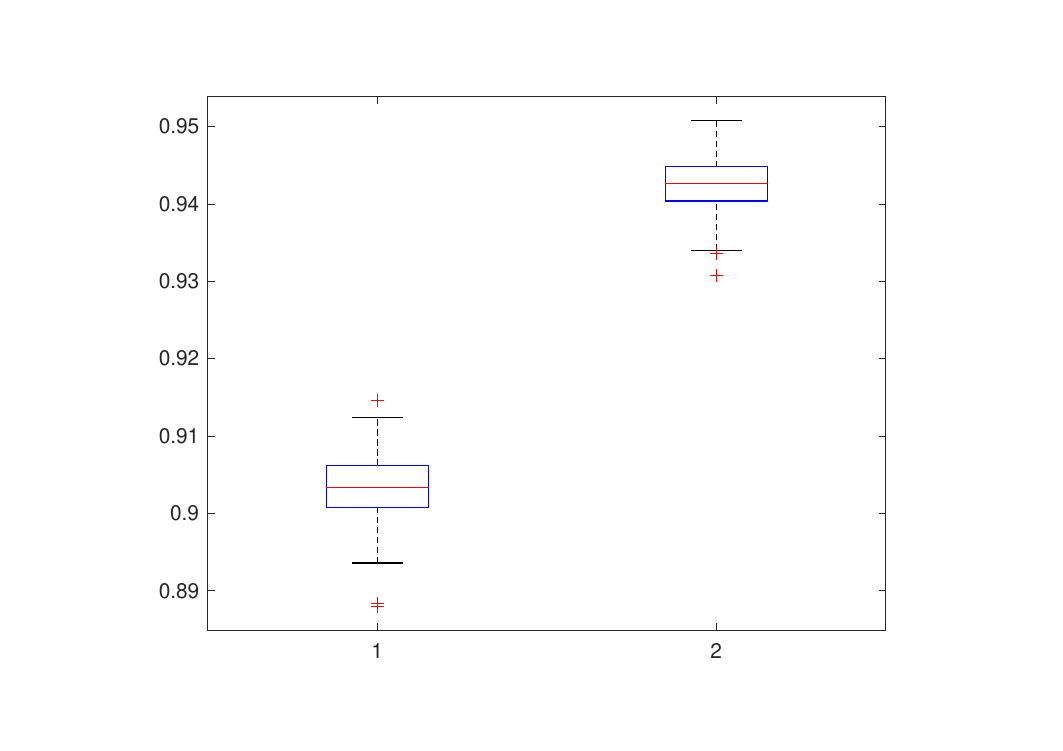}
    \caption{}
    \label{subfig:6s}
  \end{subfigure}
  \caption{Box plots for hit rates of sparse regression experiments. The settings are those described in Table \ref{tab:results_gauss_regr}. The first row presents the hit rates over all components, and the second the hit rates of the support, e.g., \ref{subfig:1} and \ref{subfig:1s} correspond to the first column of the table, \ref{subfig:2} and \ref{subfig:2s} to the second one and so forth. In each plot, the left box represents the asymptotic hit rates, and the right one the Gaussian-adjusted hit rates.}
  \label{fig:boxplots setting table}
\end{figure}

\paragraph{UQ for Classical Model-Based Regression}
For the experiments described here, we use Tfocs \cite{becker2011templates}. Analogous to the experiment described in Section \ref{subsec:classic regression}, we run further experiments in the classical sparse regression setting when the measurement matrix is a Gaussian and subsampled Fourier matrix. The different settings, including the results, can be found in Table \ref{tab:results_gauss_regr}. The results show that the Gaussian adjustment of our proposed method significantly increases the hit rates, especially on the support, while moderately increasing the confidence interval length. Our data-driven adjustment achieves even better hit rates, but the confidence intervals are larger. Although in well-posed settings, like the second column of Table \ref{tab:results_gauss_regr}, the hit rates $h^W(0.05)$ based on asymptotic confidence intervals lead \emph{overall} to almost $95\%$, however on the \emph{support}, which are the crucial features, the asymptotic hit rates fail. In particular, our corrections are essential in ill-posed regression problems as the third Gaussian column. The hit rates for the asymptotic CIs and the corrected ones with Gaussian adjustment are visualized in more detail in Figure \ref{fig:boxplots setting table}.

\begin{table}
\centering
\begin{tabular}{|c|ccc|ccc|}
\hline
                        & \multicolumn{3}{c|}{Gaussian}                                      & \multicolumn{3}{c|}{Fourier}                                       \\ \hline
feature dimension           & \multicolumn{1}{c|}{1000}   & \multicolumn{2}{c|}{10000}           & \multicolumn{1}{c|}{1000}   & \multicolumn{1}{c|}{10000}  & 100000 \\ \hline
undersampling               & \multicolumn{1}{c|}{50\%}   & \multicolumn{1}{c|}{40\%}   & 60\%   & \multicolumn{1}{c|}{40\%}   & \multicolumn{1}{c|}{60\%}   & 50\%   \\ \hline
sparsity                    & \multicolumn{1}{c|}{7.5\%}  & \multicolumn{1}{c|}{2\%}    & 10\%   & \multicolumn{1}{c|}{5\%}  & \multicolumn{1}{c|}{10\%}    & 5\%    \\ \hline
relative noise              & \multicolumn{1}{c|}{15\%}   & \multicolumn{1}{c|}{10\%}   & 20\%   & \multicolumn{1}{c|}{15\%}   & \multicolumn{1}{c|}{5\%}   & 10\%   \\ 
\specialrule{.1em}{.05em}{.05em}
R/W $\ell_2$-ratio          & \multicolumn{1}{c|}{0.7062} & \multicolumn{1}{c|}{0.5117} & 0.9993 & \multicolumn{1}{c|}{0.5446} & \multicolumn{1}{c|}{0.5924} & 0.4701 \\ \hline
R/W $\ell_\infty$-ratio     & \multicolumn{1}{c|}{0.8191} & \multicolumn{1}{c|}{0.5161} & 1.1581 & \multicolumn{1}{c|}{0.5752} & \multicolumn{1}{c|}{0.6088} & 0.4794 \\
\specialrule{.1em}{.05em}{.05em}
average asympt. radius $r^W(0.05)$ & \multicolumn{1}{c|}{0.0116} & \multicolumn{1}{c|}{0.0027} & 0.0049 & \multicolumn{1}{c|}{0.0130} & \multicolumn{1}{c|}{0.0011} & 0.0008 \\ \hline
av. radius $r^G(0.05)$ (Thm.\ref{thm:remainder_dist_gaussian})      & \multicolumn{1}{c|}{0.0142} & \multicolumn{1}{c|}{0.0031} & 0.0069 & \multicolumn{1}{c|}{0.0148} & \multicolumn{1}{c|}{0.0013} & 0.0009 \\ \hline
av. radius $r(0.05)$ (Thm.\ref{thm:main_stat_result}) & \multicolumn{1}{c|}{0.0304} & \multicolumn{1}{c|}{0.0060} & 0.0155 & \multicolumn{1}{c|}{0.0295} & \multicolumn{1}{c|}{0.0026} & 0.0016 \\ 
\specialrule{.1em}{.05em}{.05em}
$h^W(0.05)$                 & \multicolumn{1}{c|}{0.9437} & \multicolumn{1}{c|}{0.9353} & 0.8692 & \multicolumn{1}{c|}{0.9582} & \multicolumn{1}{c|}{0.9246} & 0.9444 \\ \hline
$h^G(0.05)$                 & \multicolumn{1}{c|}{0.9787} & \multicolumn{1}{c|}{0.9684} & 0.9691 & \multicolumn{1}{c|}{0.9799} & \multicolumn{1}{c|}{0.9665} & 0.9687 \\ \hline
$h(0.05)$                   & \multicolumn{1}{c|}{1} & \multicolumn{1}{c|}{1} & 1 & \multicolumn{1}{c|}{1} & \multicolumn{1}{c|}{1} & 0.9999 \\ 
\specialrule{.1em}{.05em}{.05em}
$h^W_S(0.05)$               & \multicolumn{1}{c|}{0.7661} & \multicolumn{1}{c|}{0.8941} & 0.6783 & \multicolumn{1}{c|}{0.8629} & \multicolumn{1}{c|}{0.8745} & 0.9041 \\ \hline
$h^G_S(0.05)$                 & \multicolumn{1}{c|}{0.8852} & \multicolumn{1}{c|}{0.9421} & 0.8948 & \multicolumn{1}{c|}{0.9226} & \multicolumn{1}{c|}{0.9396} & 0.9425 \\ \hline
$h_S(0.05)$                 & \multicolumn{1}{c|}{0.9999} & \multicolumn{1}{c|}{1} & 1 & \multicolumn{1}{c|}{0.9999} & \multicolumn{1}{c|}{1} & 0.9998 \\ \hline
\end{tabular}
\caption{Experiments for sparse regression for Gaussian and Fourier matrix. Every experiment uses 500 estimation and 250 evaluation data.}
\label{tab:results_gauss_regr}
\end{table}
 
\begin{figure}[tb]
  \centering
  \begin{subfigure}[T]{0.16\linewidth}
    \includegraphics[width=\textwidth]{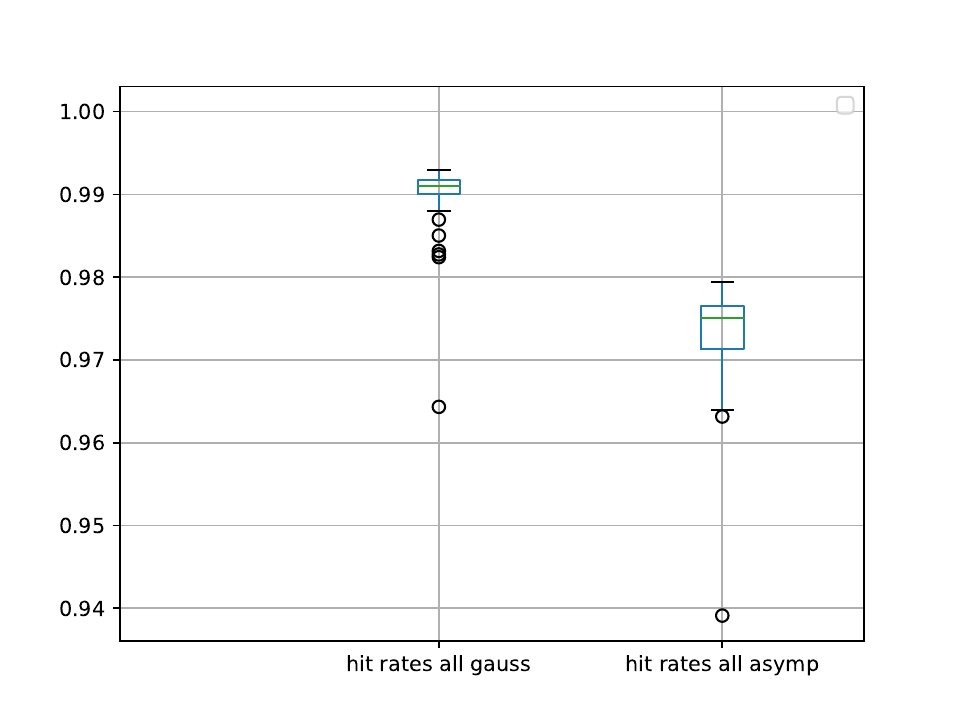}
    \caption{}
    \label{subfig:1itnet}
  \end{subfigure}
  \hfill
  \begin{subfigure}[T]{0.16\linewidth}
    \includegraphics[width=\textwidth]{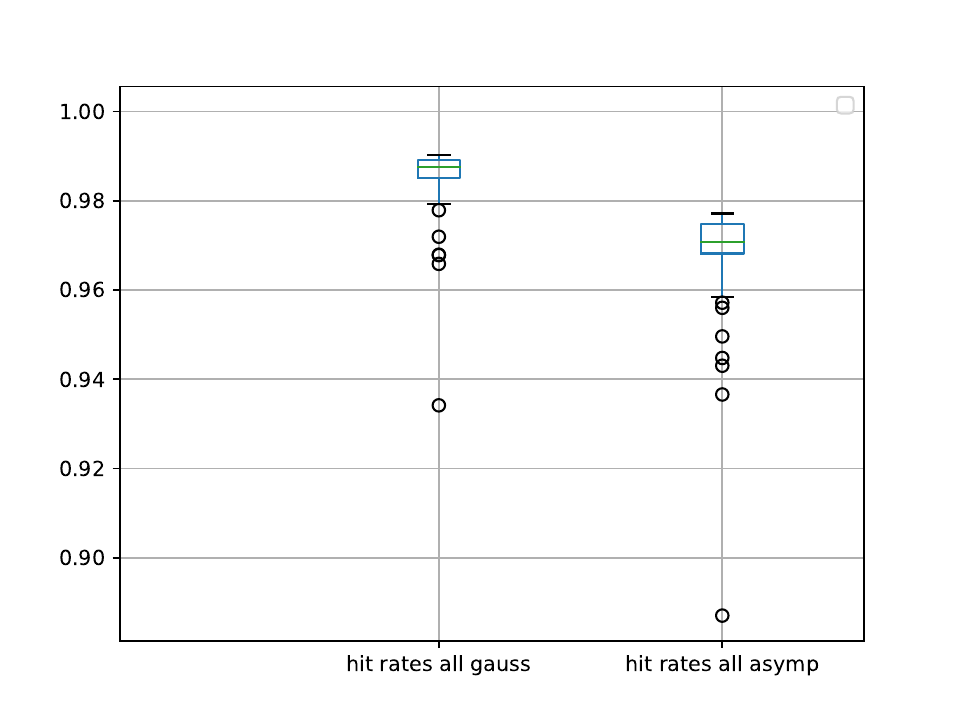}
    \caption{}
    \label{subfig:2itnet}
  \end{subfigure}
  \hfill
   \begin{subfigure}[T]{0.16\linewidth}
     \includegraphics[width=\textwidth]{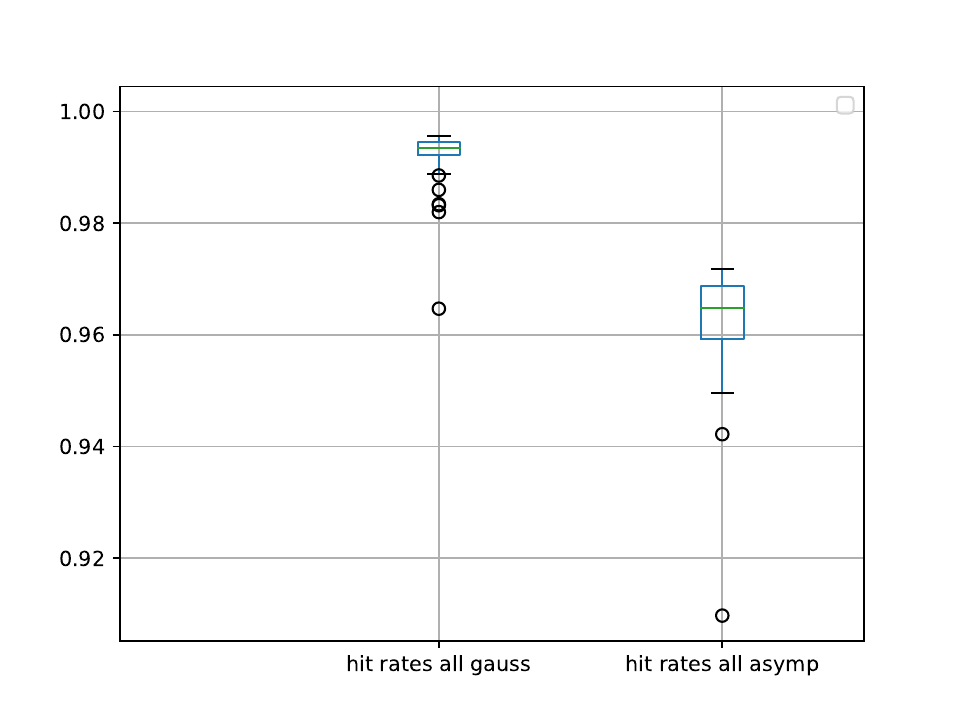}
    \caption{}
    \label{subfig:3itnet}
  \end{subfigure}
  \begin{subfigure}[T]{0.16\linewidth}
     \includegraphics[width=\textwidth]{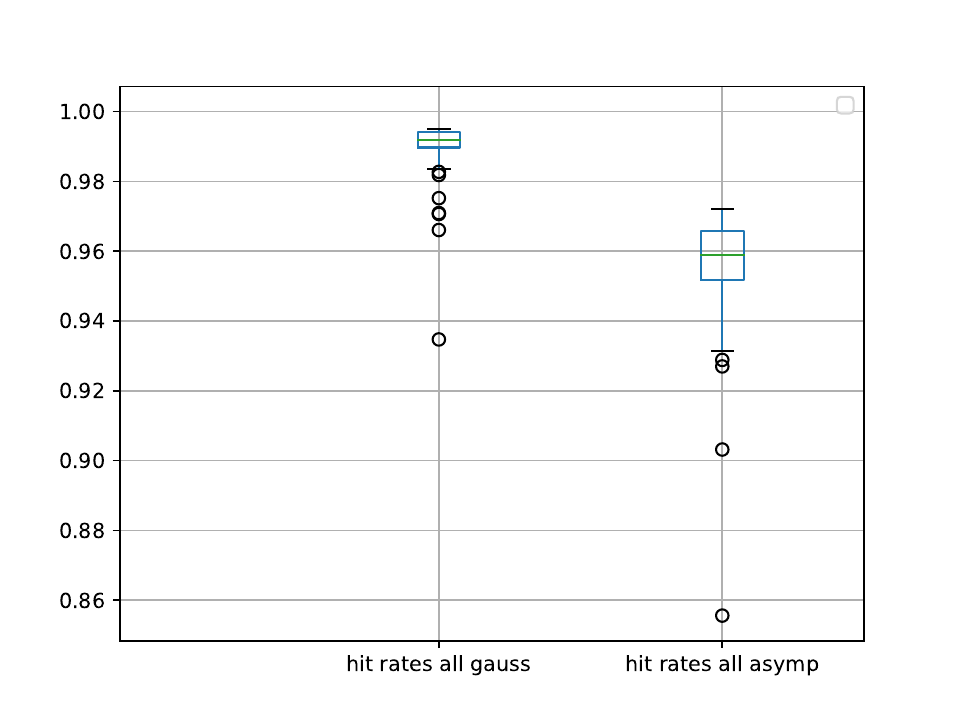}
    \caption{}
    \label{subfig:4itnet}
  \end{subfigure}
  \begin{subfigure}[T]{0.16\linewidth}
    \includegraphics[width=\textwidth]{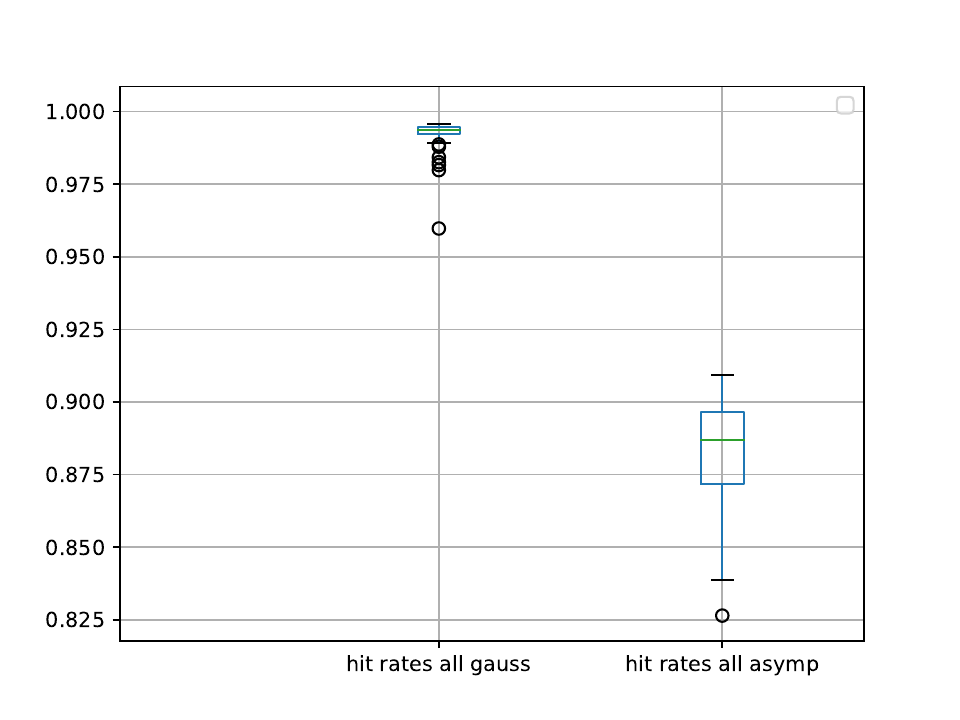}
    \caption{}
    \label{subfig:5itnet}
  \end{subfigure}
  \hfill
  \begin{subfigure}[T]{0.16\linewidth}
    \includegraphics[width=\textwidth]{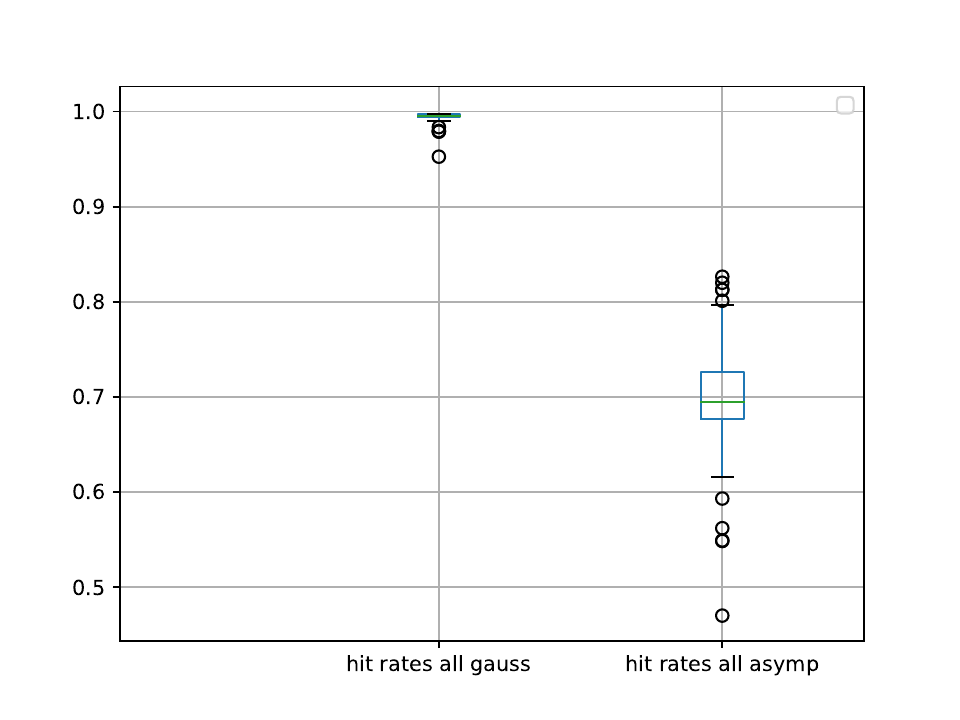}
    \caption{}
    \label{subfig:6itnet}
  \end{subfigure}
  \hfill
   \begin{subfigure}[T]{0.16\linewidth}
     \includegraphics[width=\textwidth]{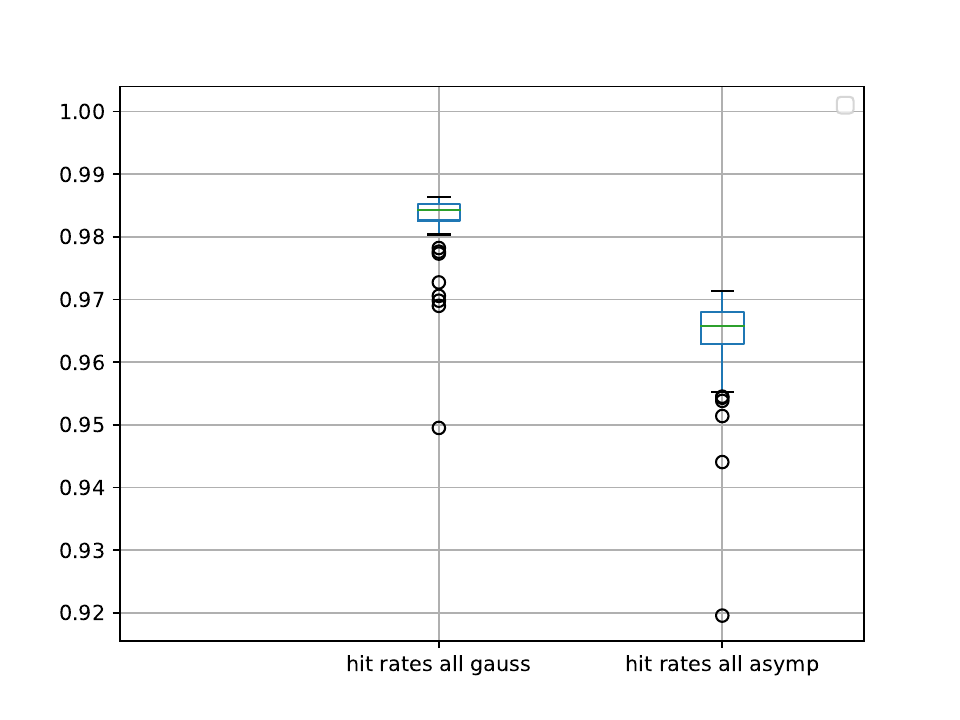}
    \caption{}
    \label{subfig:1unet}
  \end{subfigure}
  \begin{subfigure}[T]{0.16\linewidth}
     \includegraphics[width=\textwidth]{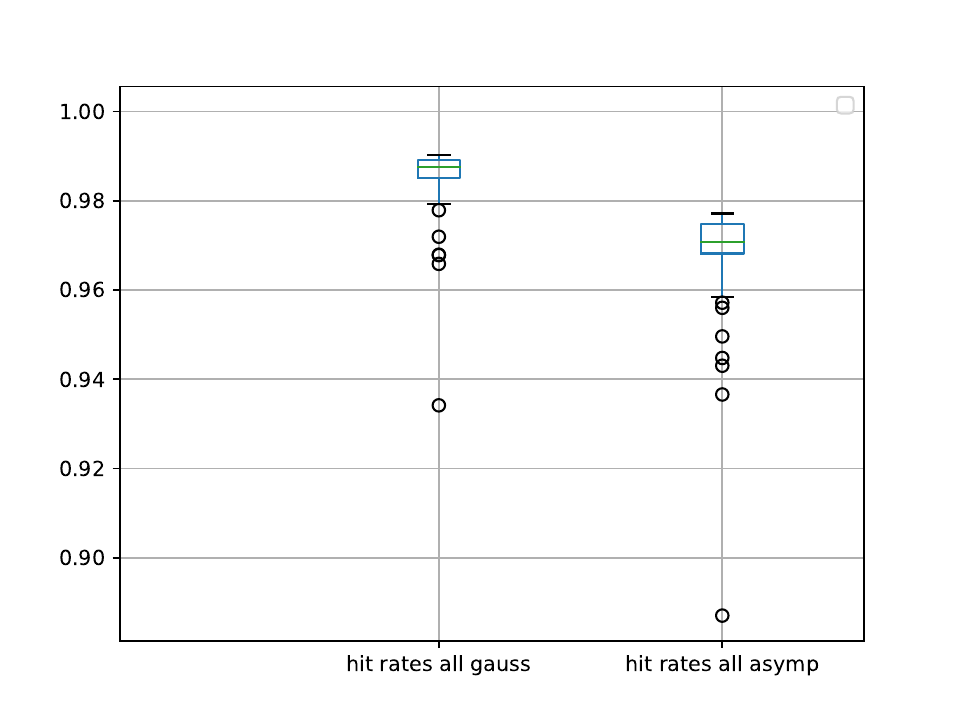}
    \caption{}
    \label{subfig:2unet}
  \end{subfigure}
  \hfill
   \begin{subfigure}[T]{0.16\linewidth}
    \includegraphics[width=\textwidth]{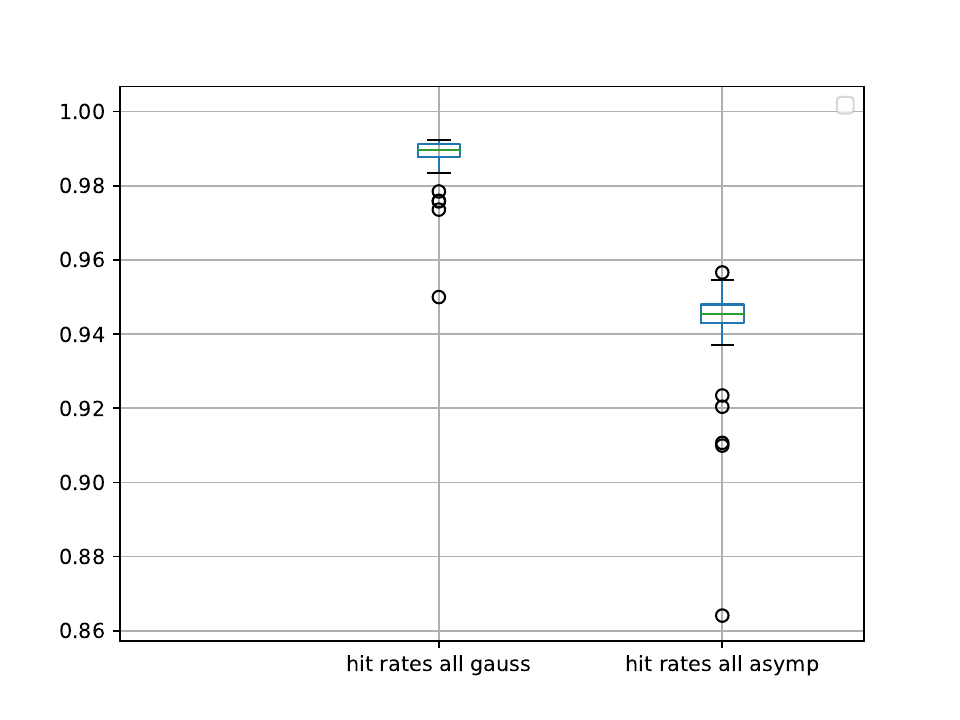}
    \caption{}
    \label{subfig:3unet}
  \end{subfigure}
    \hfill
   \begin{subfigure}[T]{0.16\linewidth}
    \includegraphics[width=\textwidth]{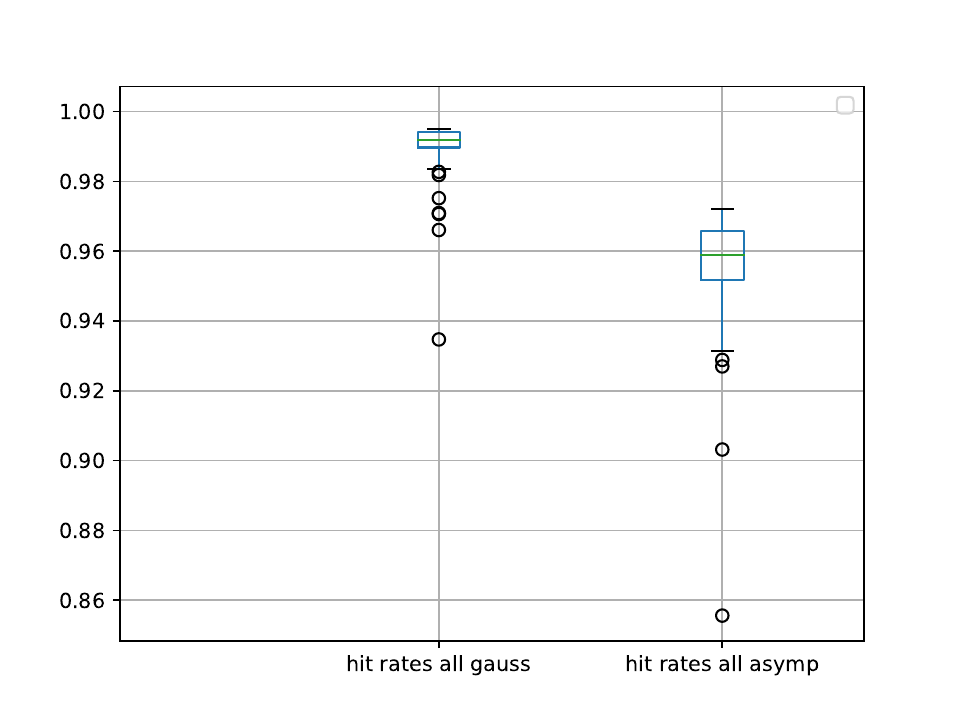}
    \caption{}
    \label{subfig:4unet}
  \end{subfigure}
     \begin{subfigure}[T]{0.16\linewidth}
    \includegraphics[width=\textwidth]{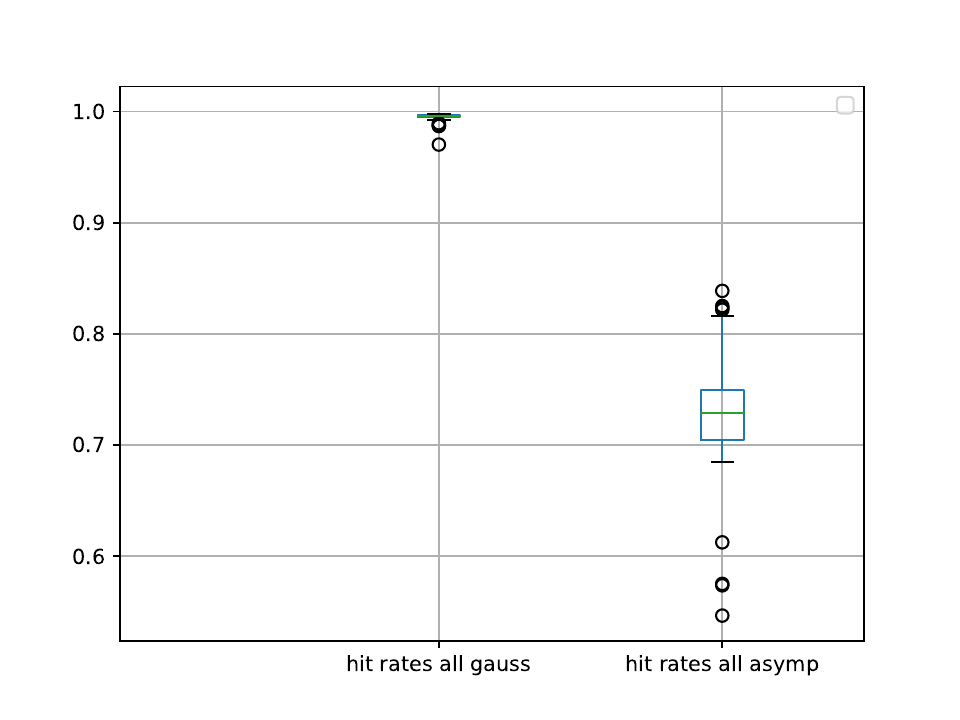}
    \caption{}
    \label{subfig:5unet}
  \end{subfigure}
    \hfill
   \begin{subfigure}[T]{0.16\linewidth}
    \includegraphics[width=\textwidth]{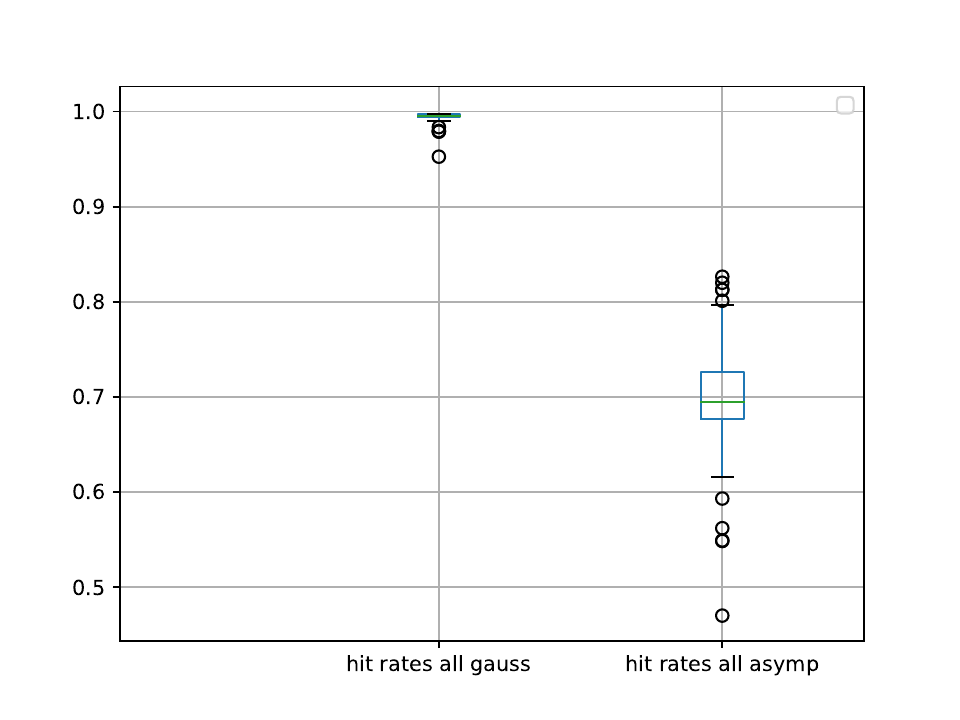}
    \caption{}
    \label{subfig:6unet}
  \end{subfigure}
  \caption{Box plots for hit rates of neural network experiments. The settings are those described in Table \ref{tab:results_itnet} and Table \ref{tab:results_unet}. The first row presents the hit rates for the different It-nets for $95\%$ confident intervals and the second the hit rates for the U-Nets for $95\%$ confident intervals, e.g., \ref{subfig:1itnet} and \ref{subfig:1unet} correspond to the first columns of the tables. In each plot, the left box represents the asymptotic hit rates, the right the Gaussian-adjusted ones.}
  \label{fig:boxplots setting table networks}
\end{figure}

\paragraph{UQ for MRI Reconstruction with Neural Networks}
In this section, we present more experiments for UQ for MRI reconstruction with neural networks. Our experimental settings, as well as our code for this experiment, are based on the paper and code \footnote{https://github.com/jmaces/robust-nets} by \cite{genzel2022near}. The dataset used for conducting the experiments is the fastMRI single-coil knee dataset. For documentation, see \cite{zbontar2019fastmri, fastMRIdataset}. \\
Table \ref{tab:results_itnet} represents the results obtained by learning the reconstruction function $\hat X$ using the It-net with $8$ layers, with $60 \%, 40 \%$ and $30 \%$ radial undersampling and for noise levels obtained by adding complex gaussian noise with standard deviation $\sigma=60$ and $\sigma=84$, respectively.  Similarly, Table \ref{tab:results_unet} shows the results obtained by the U-Net. In Figure \ref{fig:boxplots setting table networks}, the asymptotic hit rates and the Gaussian adjusted ones for the $95 \%$ confidence level are compared in a box plot for each experiment.

All It-Net and U-Nets are trained with a combination of the MS-SSIM-loss \cite{wang2003multiscale}, the $\ell_1$-loss and the Adam optimizer with a learning rate of $5e^{-5}$, epsilon of $1e^{-4}$ and weight decay parameter $1e^{-5}$. The It-Nets were trained for $15$ epochs, and the U-Nets were trained for $20$ epochs, both with batch size $40$. Every U-Net has $2$ input and output channels, $24$ base channels, and encodes the image to a size of $20 \times 20$ and at most $384$ channels. The It-Net employs the U-Net in each layer as a residual network and has a data consistency part around each U-Net in every layer. 

Comparing the tables, the It-Net has, in general, better hit rates as well as a better R/W ratio than the U-Net due to its more accurate reconstruction. Further, the hit rates for all the pixels are higher than those obtained only for the support. For achieving reliable results for safety-critical applications, obtaining hit rates higher than the confidence level is crucial, especially on the support, i.e., on the non-zeros pixels. Otherwise, one might achieve a certain confidence level overall but cannot trust the pixels of interest.

The experiments were conducted using Pytorch $1.9$ on a desktop with AMD EPYC 7F52 16-Core CPUs and NVIDIA A100 PCIe030 030 40GB GPUs. The code for the experiments can be found in the supplementary material. The execution time of the code is around $5$ hours for each It-Net, $2$ hours for each U-Net, and around $30$ minutes for the rest of each experiment. So, in total, this gives us a time of $48$ hours for the MRI reconstruction experiments. The execution time for the classical model-based regression experiments takes $5$ to $30$ minutes each; therefore, in total, it is less than $3$ hours.

\begin{figure}
    \centering
    \begin{subfigure}[T]{0.33\linewidth}
    \rotatebox[origin=c]{180}
      {\includegraphics[width=\linewidth]{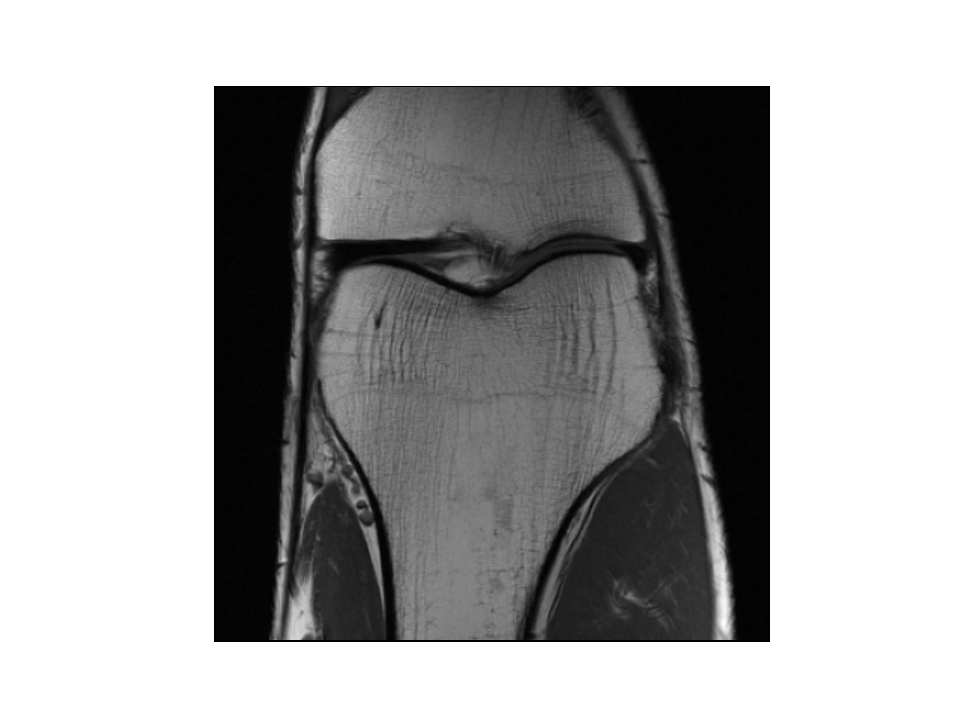}}
    \caption{}
    \label{subfig:knee}
    \end{subfigure}
    \begin{subfigure}[T]{0.33\linewidth}
    \includegraphics[width=\linewidth]{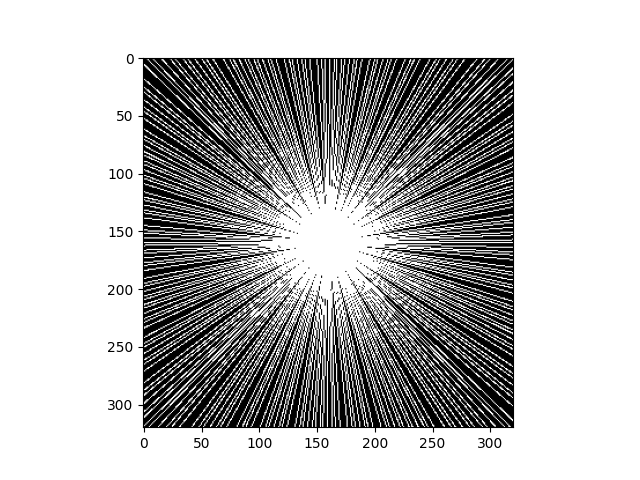}
    \caption{}
    \label{subfig:mask}
    \end{subfigure}
    \hspace{-1cm}
    \begin{subfigure}[T]{0.33\linewidth}
    \includegraphics[width=\linewidth]{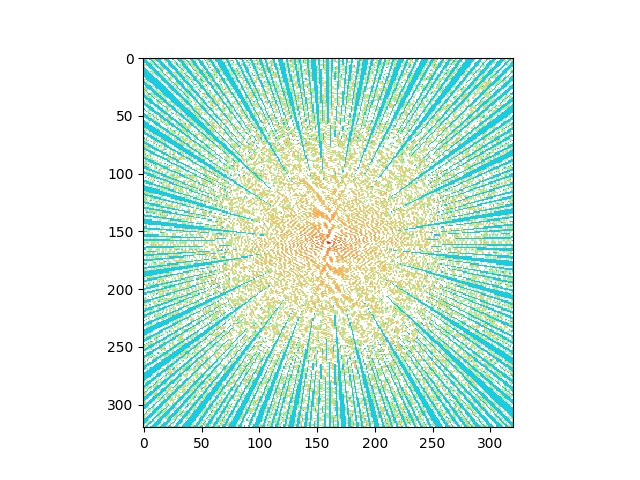}
    \caption{}
    \label{subfig:k-space}
    \end{subfigure}
    \caption{Knee MRI groundtruth image from fastMRI dataset \ref{subfig:knee} \cite{zbontar2019fastmri, fastMRIdataset}, radial sampling mask \ref{subfig:mask} and undersampled k-space data \ref{subfig:k-space}.}
    \label{fig:imagemaskkspace}
\end{figure}

\begin{table}
 \centering
\begin{tabular}{|c|cc|cc|cc|}
\hline
undersampling & \multicolumn{2}{c|}{60\%}            & \multicolumn{2}{c|}{40\%}            & \multicolumn{2}{c|}{30\%}            \\ \specialrule{.1em}{.05em}{.05em}
noise level   & \multicolumn{1}{c|}{15\%}   & 10\%   & \multicolumn{1}{c|}{12\%}   & 8\%    & \multicolumn{1}{c|}{10\%}   & 7\%    \\ \hline
R/W $\ell_2$-ratio   & \multicolumn{1}{c|}{0.3558} & 0.3800 & \multicolumn{1}{c|}{0.6332} & 0.6922 & \multicolumn{1}{c|}{0.8759} & 0.5759 \\ 
\hline
R/W $\ell_{\infty}$-ratio     & \multicolumn{1}{c|}{0.3842} & 0.4929 & \multicolumn{1}{c|}{0.6268} & 0.6924  & \multicolumn{1}{c|}{1.0614 } & 0.6282 \\ \specialrule{.1em}{.05em}{.05em} 
$h(0.05)$     & \multicolumn{1}{c|}{1.0} &  1.0 & \multicolumn{1}{c|}{1.0} & 1.0 & \multicolumn{1}{c|}{1.0} & 1.0 \\ \hline
$h^G(0.05)$     & \multicolumn{1}{c|}{0.9904} &  0.9899 & \multicolumn{1}{c|}{0.9925} & 0.9919 & \multicolumn{1}{c|}{0.9926} & 0.9893 \\ \hline
$h^W(0.05)$   & \multicolumn{1}{c|}{0.9737} & 0.9787 & \multicolumn{1}{c|}{0.9632} & 0.9744 & \multicolumn{1}{c|}{0.8825} & 0.9806 \\ \hline
$h(0.1)$      & \multicolumn{1}{c|}{0.9999} & 0.9999 & \multicolumn{1}{c|}{1.0} & 1.0 & \multicolumn{1}{c|}{1.0} & 1.0 \\ \hline
$h^G(0.1)$     & \multicolumn{1}{c|}{0.9754} & 0.9752  & \multicolumn{1}{c|}{0.9800} & 0.9793 & \multicolumn{1}{c|}{0.98} & 0.9741 \\ \hline
$h^W(0.1)$    & \multicolumn{1}{c|}{0.9403} & 0.9502 & \multicolumn{1}{c|}{0.9196} & 0.9409 & \multicolumn{1}{c|}{0.8094} & 0.9581\\  \specialrule{.1em}{.05em}{.05em}
$h_S(0.05)$   & \multicolumn{1}{c|}{1.0} & 1.0 & \multicolumn{1}{c|}{1.0} & 1.0 & \multicolumn{1}{c|}{1.0} & 1.0 \\ \hline
$h_S^G(0.05)$     & \multicolumn{1}{c|}{0.965} & 0.99 & \multicolumn{1}{c|}{0.995} &  0.9995 & \multicolumn{1}{c|}{0.985} & 0.985 \\ \hline
$h_S^W(0.05)$ & \multicolumn{1}{c|}{0.94} & 0.91  & \multicolumn{1}{c|}{0.955} & 0.915 & \multicolumn{1}{c|}{0.975} & 0.965 \\ \hline
$h_S(0.1)$    & \multicolumn{1}{c|}{1.0} & 1.0 & \multicolumn{1}{c|}{1.0} & 1.0 & \multicolumn{1}{c|}{1.0} & 1.0 \\ \hline
$h_S^G(0.1)$     & \multicolumn{1}{c|}{0.955} &  0.98 & \multicolumn{1}{c|}{0.985} & 0.995 & \multicolumn{1}{c|}{0.98} & 0.975\\ \hline
$h_S^W(0.1)$  & \multicolumn{1}{c|}{0.89} & 0.87 & \multicolumn{1}{c|}{0.92} & 0.845 & \multicolumn{1}{c|}{0.955} & 0.94 \\ \hline
\end{tabular}
     \caption{Experiments for It-Net with 8 iterations. Results of hit rates averaged over $k=100$ samples.}
      \label{tab:results_itnet}
\end{table}

\begin{table}
\centering
\begin{tabular}{|c|cc|cc|cc|}
\hline
undersampling & \multicolumn{2}{c|}{60\%}            & \multicolumn{2}{c|}{40\%}            & \multicolumn{2}{c|}{30\%}            \\ 
\specialrule{.1em}{.05em}{.05em} 
noise level   & \multicolumn{1}{c|}{15\%}   & 10\%   & \multicolumn{1}{c|}{12\%}   & 8\%    & \multicolumn{1}{c|}{10\%}   & 7\%    \\ \hline
R/W $\ell_2$-ratio     & \multicolumn{1}{c|}{0.2747} & 0.3976 & \multicolumn{1}{c|}{0.6182} & 0.6671 & \multicolumn{1}{c|}{1.2641} & 1.3399 \\ 
\hline
R/W $\ell_{\infty}$-ratio     & \multicolumn{1}{c|}{0.3923} & 0.4292 & \multicolumn{1}{c|}{0.5681} & 0.7082 & \multicolumn{1}{c|}{1.1668} & 1.2501\\ \specialrule{.1em}{.05em}{.05em} 
$h(0.05)$     & \multicolumn{1}{c|}{1.0} & 1.0 & \multicolumn{1}{c|}{1.0} & 1.0 & \multicolumn{1}{c|}{1.0} & 1.0 \\ \hline
$h^G(0.05)$     & \multicolumn{1}{c|}{0.9831} & 0.9857  & \multicolumn{1}{c|}{0.9885} & 0.9903  & \multicolumn{1}{c|}{0.9952} &  0.9943 \\ \hline
$h^W(0.05)$   & \multicolumn{1}{c|}{0.9642} & 0.9687 & \multicolumn{1}{c|}{0.9439} & 0.9562 & \multicolumn{1}{c|}{0.7277} & 0.7002 \\ \hline
$h(0.1)$      & \multicolumn{1}{c|}{0.9994} & 0.9999 & \multicolumn{1}{c|}{1.0} & 1.0 & \multicolumn{1}{c|}{1.0} & 1.0 \\ \hline 
$h^G(0.1)$     & \multicolumn{1}{c|}{0.9623} & 0.9674  & \multicolumn{1}{c|}{0.9723} & 0.9763 & \multicolumn{1}{c|}{0.9867} & 0.9853 \\ \hline
$h^W(0.1)$    & \multicolumn{1}{c|}{0.9230} & 0.9310 & \multicolumn{1}{c|}{0.8869} & 0.9091 & \multicolumn{1}{c|}{0.6088} & 0.5585 \\ \specialrule{.1em}{.05em}{.05em} 
$h_S(0.05)$   & \multicolumn{1}{c|}{1.0} & 1.0 & \multicolumn{1}{c|}{1.0} & 1.0 & \multicolumn{1}{c|}{1.0} & 1.0    \\ \hline
$h_S^G(0.05)$     & \multicolumn{1}{c|}{0.99} &  0.995 & \multicolumn{1}{c|}{0.995} & 0.99 & \multicolumn{1}{c|}{0.99} & 0.995 \\ \hline
$h_S^W(0.05)$ & \multicolumn{1}{c|}{0.945} & 0.93 & \multicolumn{1}{c|}{0.935} & 0.955& \multicolumn{1}{c|}{0.925} & 0.695      \\ \hline
$h_S(0.1)$    & \multicolumn{1}{c|}{1.0} & 1.0 & \multicolumn{1}{c|}{1.0} & 1.0 & \multicolumn{1}{c|}{1.0} & 1.0 \\ \hline
$h_S^G(0.1)$     & \multicolumn{1}{c|}{0.985} & 0.965  & \multicolumn{1}{c|}{0.985} & 0.975 & \multicolumn{1}{c|}{0.98} & 0.995 \\ \hline
$h_S^W(0.1)$  & \multicolumn{1}{c|}{0.895} & 0.895 & \multicolumn{1}{c|}{0.91} & 0.895  & \multicolumn{1}{c|}{0.875} & 0.605 \\ \hline
\end{tabular}
\caption{Experiments for U-Nets. Results of hit rates averaged over $k=100$ samples.}
\label{tab:results_unet}
\end{table}

\section{Distribution Visualization of Remainder Term}\label{sec:gaussian_remainder_density}

\begin{figure}[H]
  \centering
  \begin{subfigure}[T]{0.32\linewidth}
    \includegraphics[width=\textwidth]{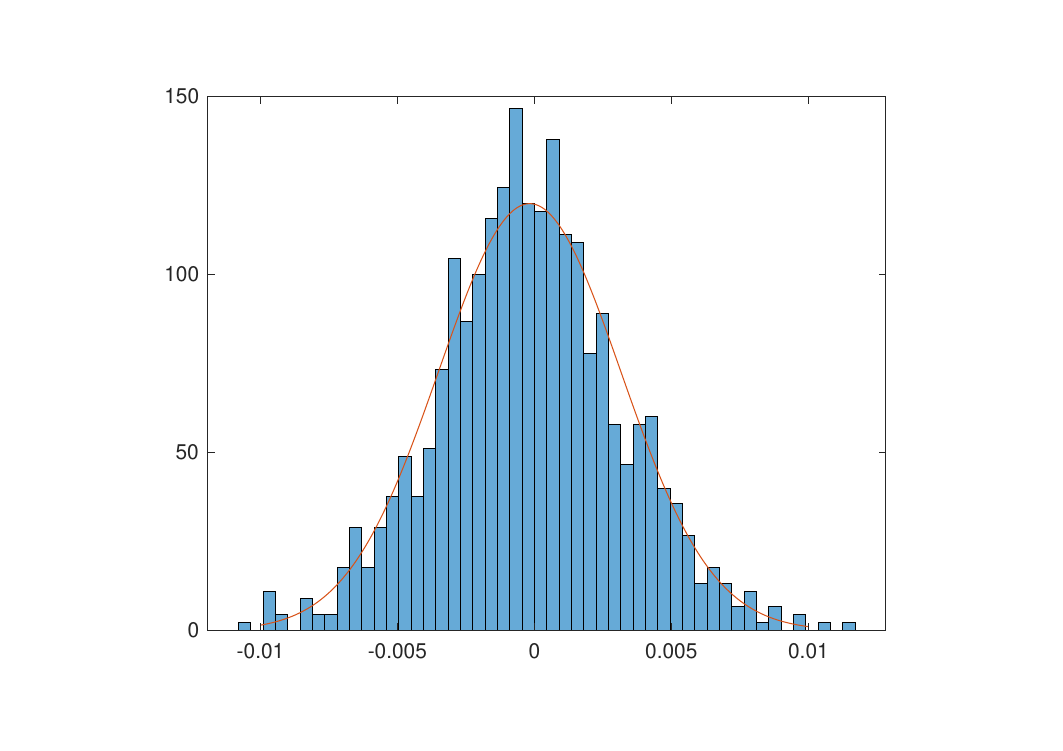}
    \caption{}
    \label{subfig:1density}
  \end{subfigure}
  \hfill
  \begin{subfigure}[T]{0.32\linewidth}
    \includegraphics[width=\textwidth]{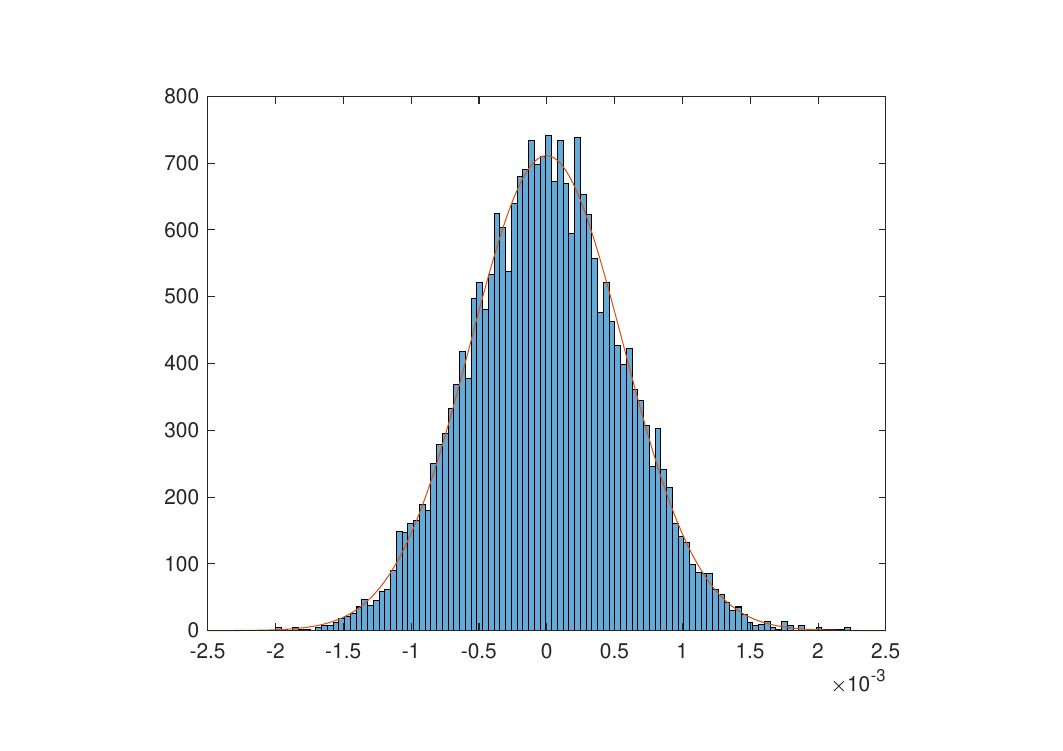}
    \caption{}
    \label{subfig:2density}
  \end{subfigure}
  \hfill
   \begin{subfigure}[T]{0.32\linewidth}
     \includegraphics[width=\textwidth]{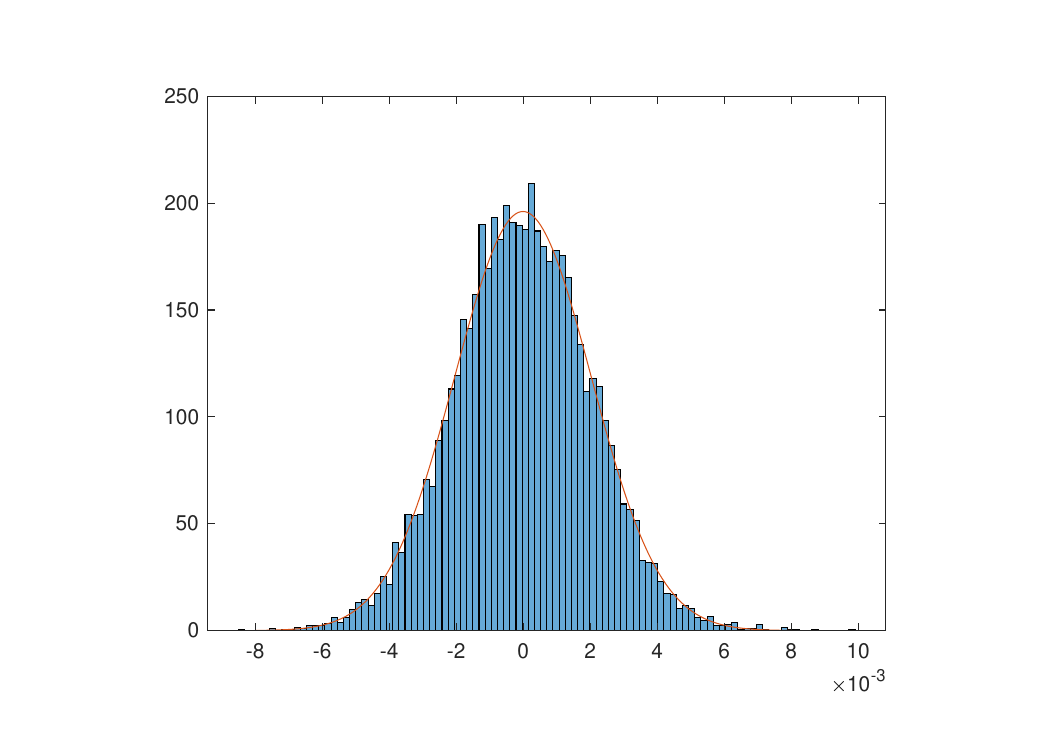}
    \caption{}
    \label{subfig:3density}
  \end{subfigure}
  \begin{subfigure}[T]{0.32\linewidth}
     \includegraphics[width=\textwidth]{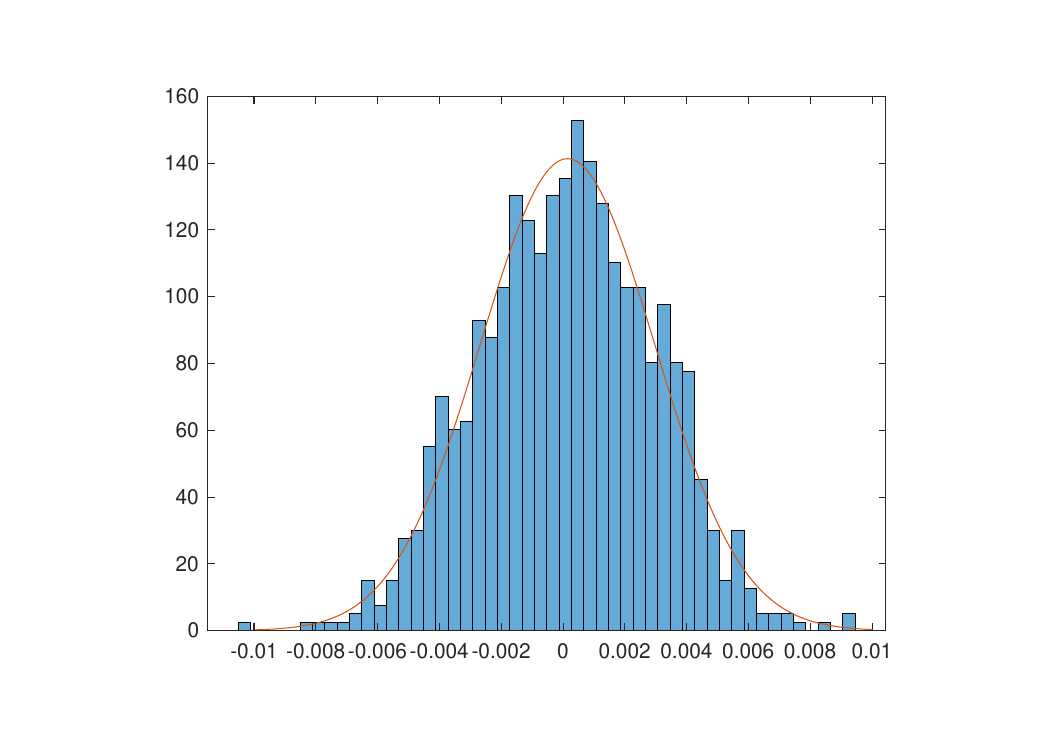}
    \caption{}
    \label{subfig:4density}
  \end{subfigure}
  \begin{subfigure}[T]{0.32\linewidth}
    \includegraphics[width=\textwidth]{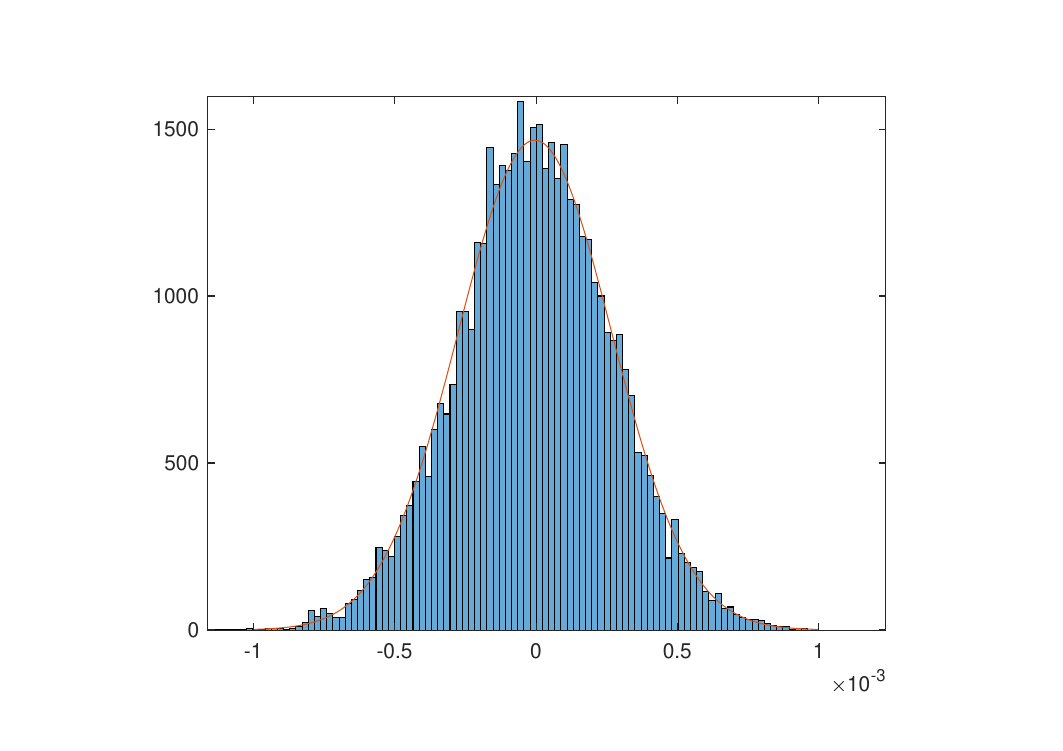}
    \caption{}
    \label{subfig:5density}
  \end{subfigure}
  \hfill
  \begin{subfigure}[T]{0.32\linewidth}
    \includegraphics[width=\textwidth]{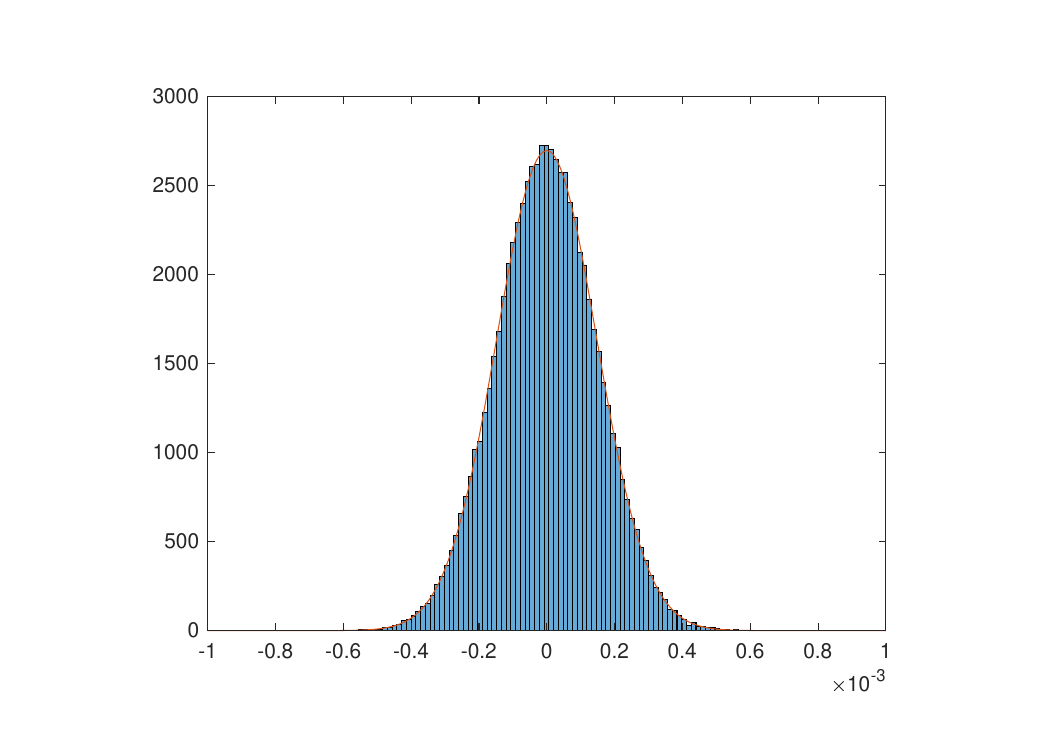}
    \caption{}
    \label{subfig:6density}
  \end{subfigure}
  \caption{Histograms showing the empirical distribution of the remainder component. The real part of $R$ is plotted as a histogram and the red line corresponds to a Gaussian fit. One realization of the remainder term is visualized for each of the experiments described in Table \ref{tab:results_gauss_regr}. (\ref{subfig:1density} corresponds to the first column, \ref{subfig:2density} to the second column and so on. The plots for the imaginary part look similar.)}
  \label{fig:densities_sparse_regr}
\end{figure}

In Figure \ref{fig:densities_sparse_regr}, we present a series of histograms illustrating the empirical distribution of the remainder term's real part across all experimental settings in sparse regression conducted in this paper. These histograms provide evidence that the remainder term can be approximated by a Gaussian distribution, with the approximation becoming increasingly precise as the dimensionality increases. Across low-dimensional scenarios, the empirical distributions exhibit some deviations from the Gaussian form, but these discrepancies diminish as the dimensionality grows larger. In high-dimensional regimes, the empirical distributions demonstrate an exceptional degree of convergence to the Gaussian approximation. This close alignment lends strong support to the validity of the key assumption of Theorem \ref{thm:remainder_dist_gaussian}, allowing a Gaussian adjustment to the confidence intervals. \\

\begin{figure}[H]
  \centering
  \begin{subfigure}[T]{0.32\linewidth}
    \includegraphics[width=\textwidth]{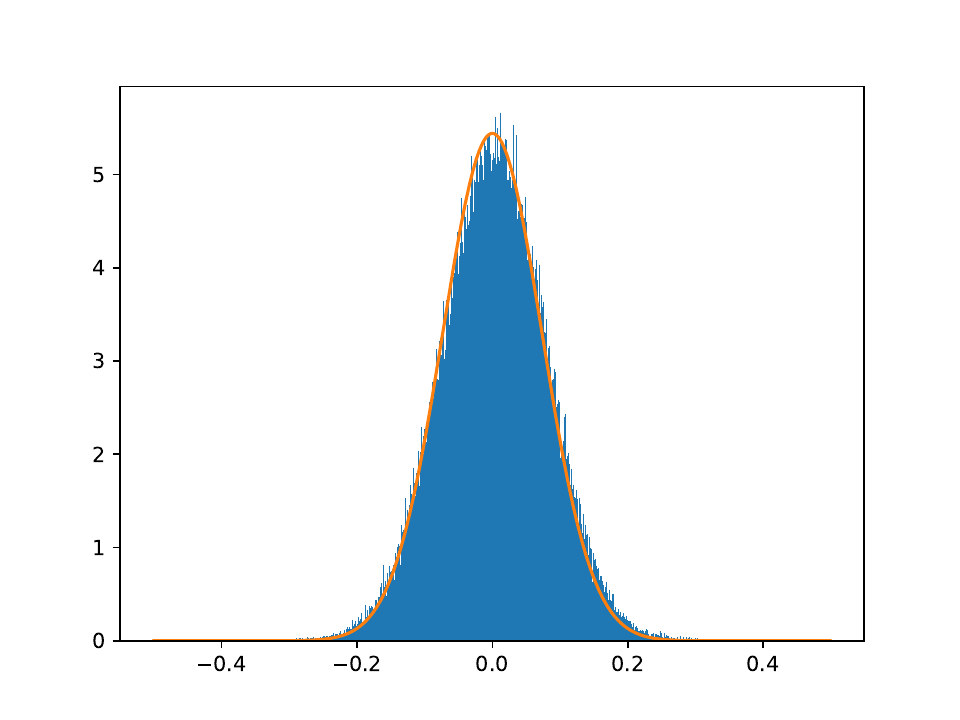}
    \caption{}
    \label{subfig:1densityunet}
  \end{subfigure}
  \hfill
  \begin{subfigure}[T]{0.32\linewidth}
    \includegraphics[width=\textwidth]{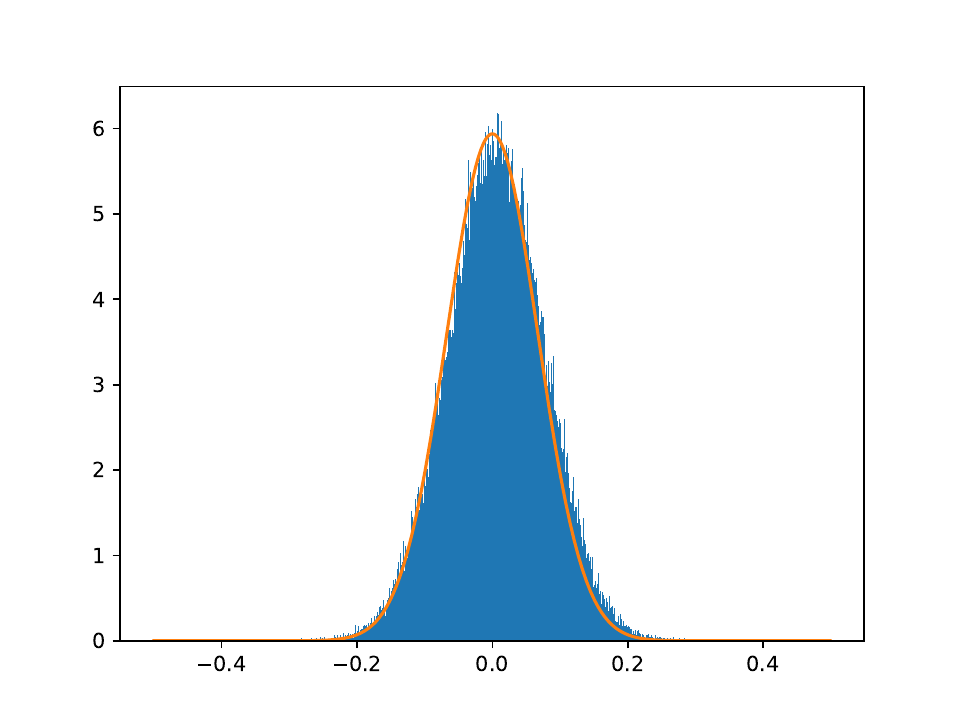}
    \caption{}
    \label{subfig:2densityunet}
  \end{subfigure}
  \hfill
   \begin{subfigure}[T]{0.32\linewidth}
     \includegraphics[width=\textwidth]{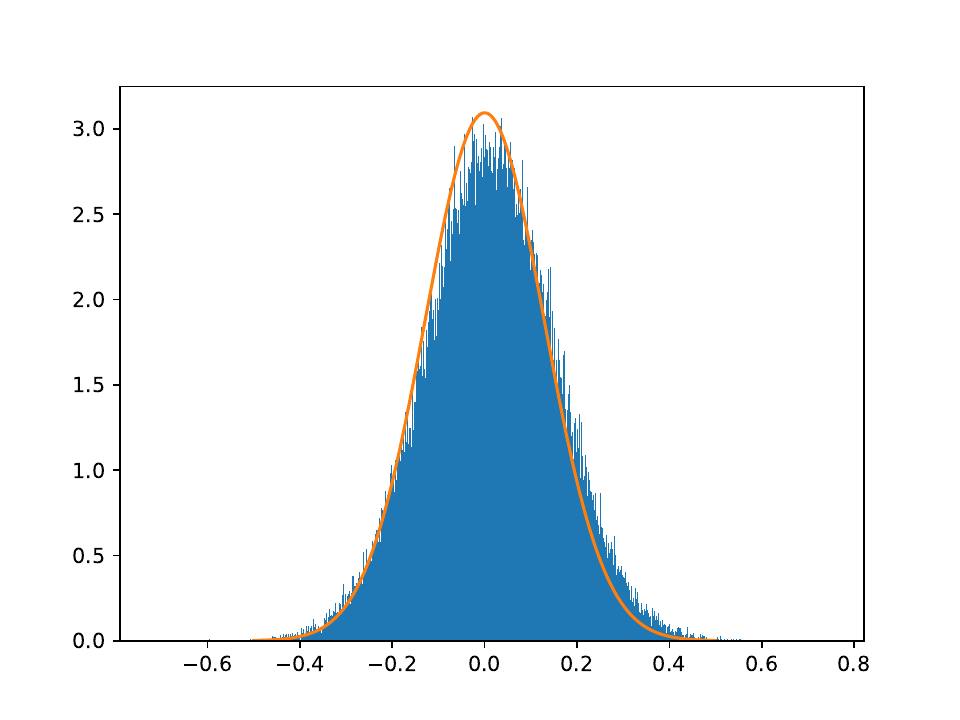}
    \caption{}
    \label{subfig:3densityunet}
  \end{subfigure}
  \begin{subfigure}[T]{0.32\linewidth}
     \includegraphics[width=\textwidth]{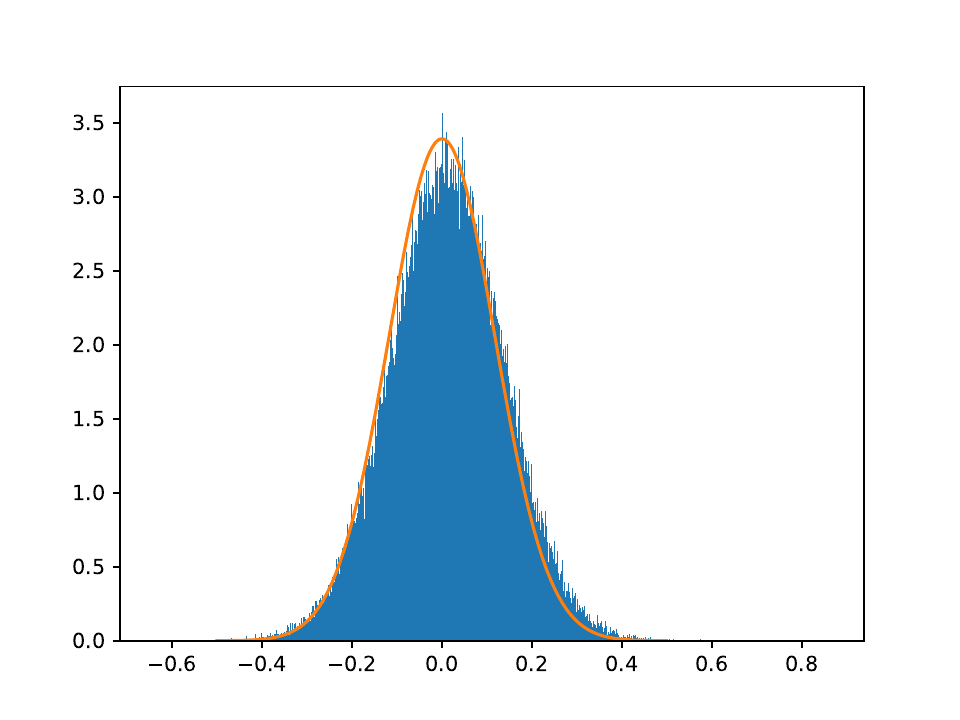}
    \caption{}
    \label{subfig:4densityunet}
  \end{subfigure}
  \begin{subfigure}[T]{0.32\linewidth}
    \includegraphics[width=\textwidth]{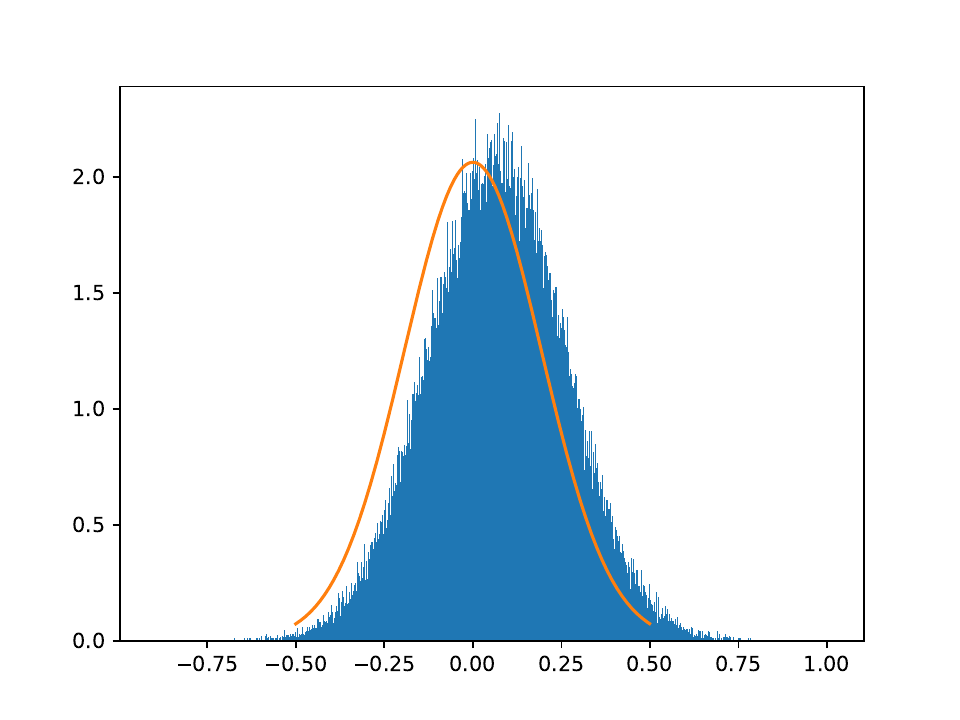}
    \caption{}
    \label{subfig:5densityunet}
  \end{subfigure}
  \hfill
  \begin{subfigure}[T]{0.32\linewidth}
    \includegraphics[width=\textwidth]{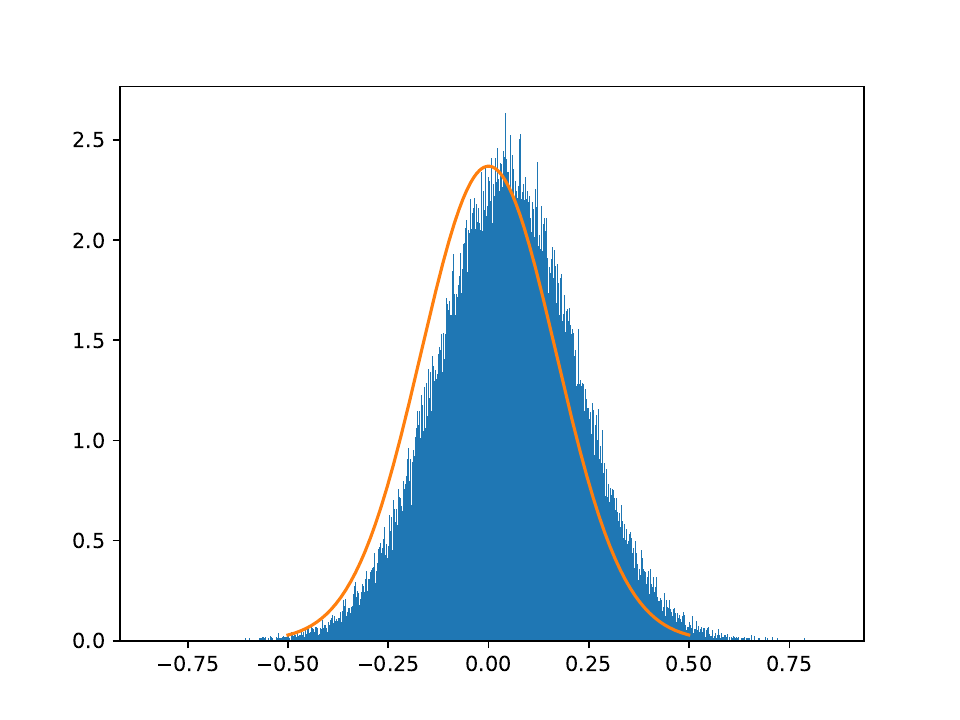}
    \caption{}
    \label{subfig:6densityunet}
  \end{subfigure}
  \caption{Histograms showing the empirical distribution of the remainder component. The real part of $R$ is plotted as a histogram and the red line corresponds to a Gaussian fit. One realization of the remainder term is visualized for each of the experiments described in Table \ref{tab:results_unet}. (\ref{subfig:1densityunet} corresponds to the first column, \ref{subfig:2densityunet} to the second column and so on.)}
  \label{fig:densities_unet}
\end{figure}

\begin{figure}[H]
  \centering
  \begin{subfigure}[T]{0.32\linewidth}
    \includegraphics[width=\textwidth]{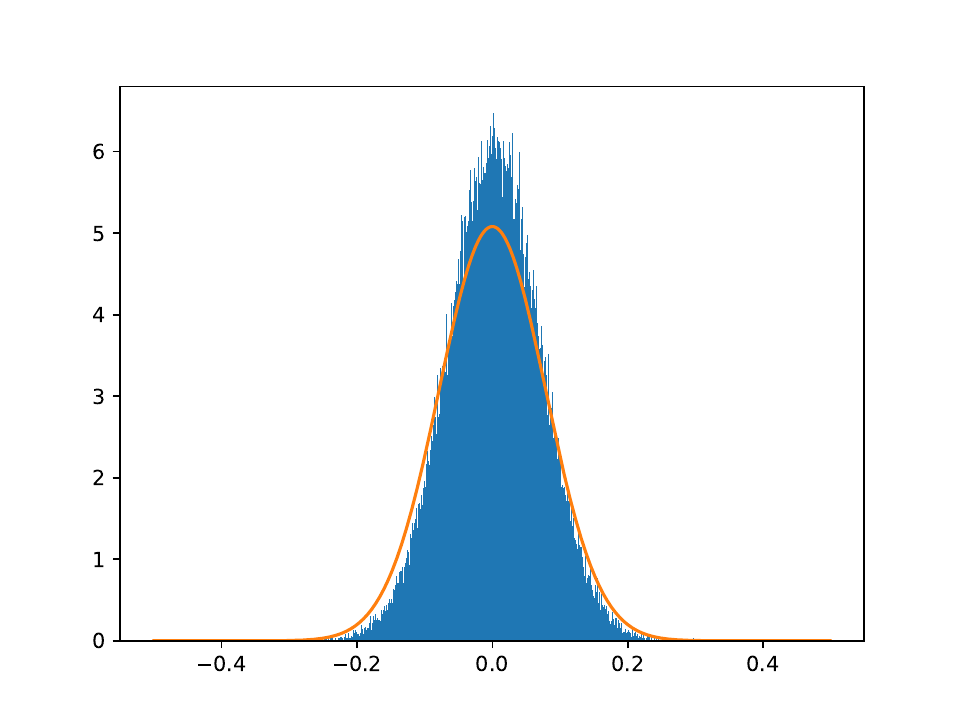}
    \caption{}
    \label{subfig:1densityitnet}
  \end{subfigure}
  \hfill
  \begin{subfigure}[T]{0.32\linewidth}
    \includegraphics[width=\textwidth]{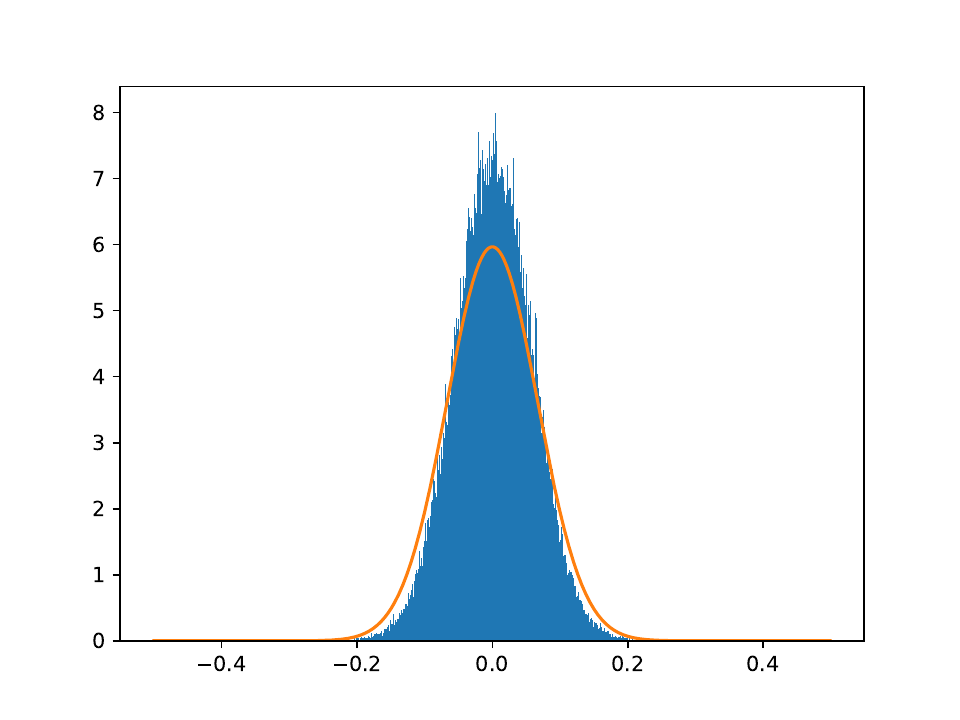}
    \caption{}
    \label{subfig:2densityitnet}
  \end{subfigure}
  \hfill
   \begin{subfigure}[T]{0.32\linewidth}
     \includegraphics[width=\textwidth]{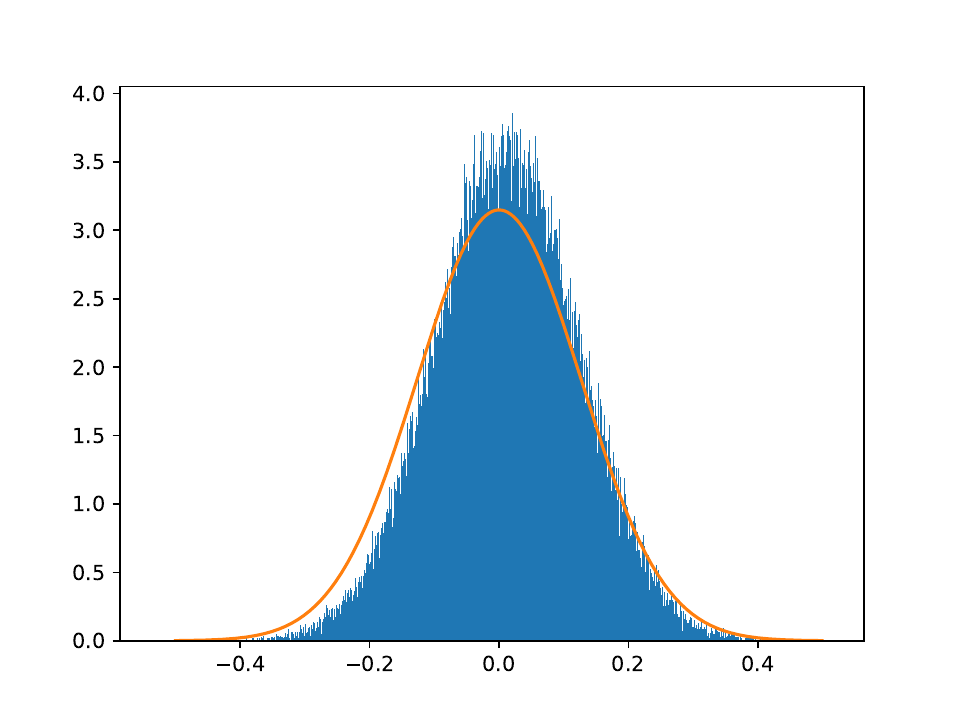}
    \caption{}
    \label{subfig:3densityitnet}
  \end{subfigure}
  \begin{subfigure}[T]{0.32\linewidth}
     \includegraphics[width=\textwidth]{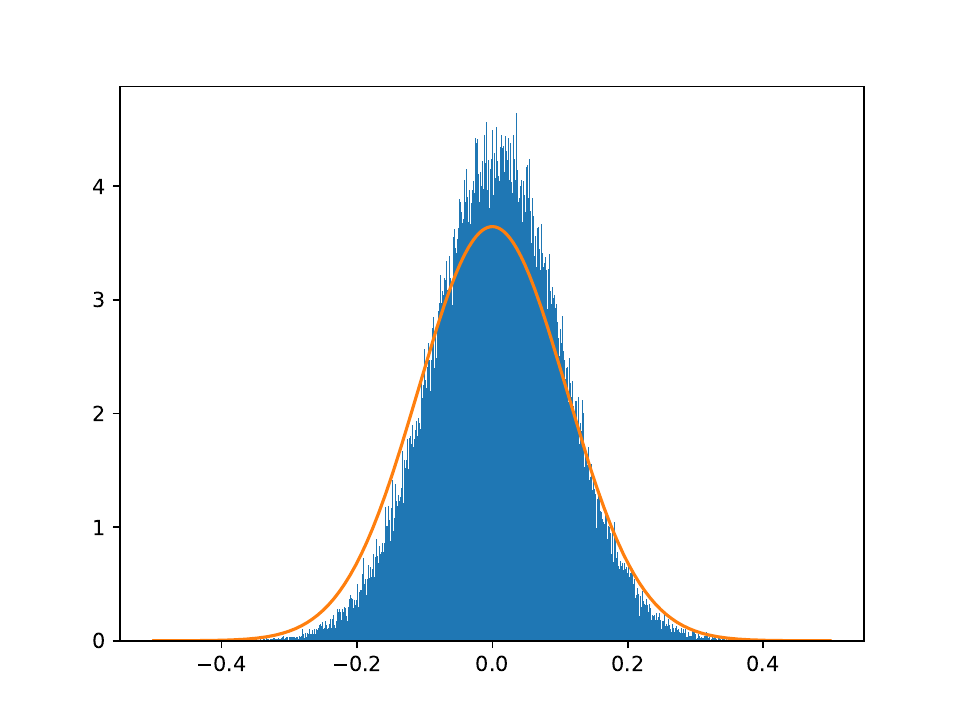}
    \caption{}
    \label{subfig:4densityitnet}
  \end{subfigure}
  \begin{subfigure}[T]{0.32\linewidth}
    \includegraphics[width=\textwidth]{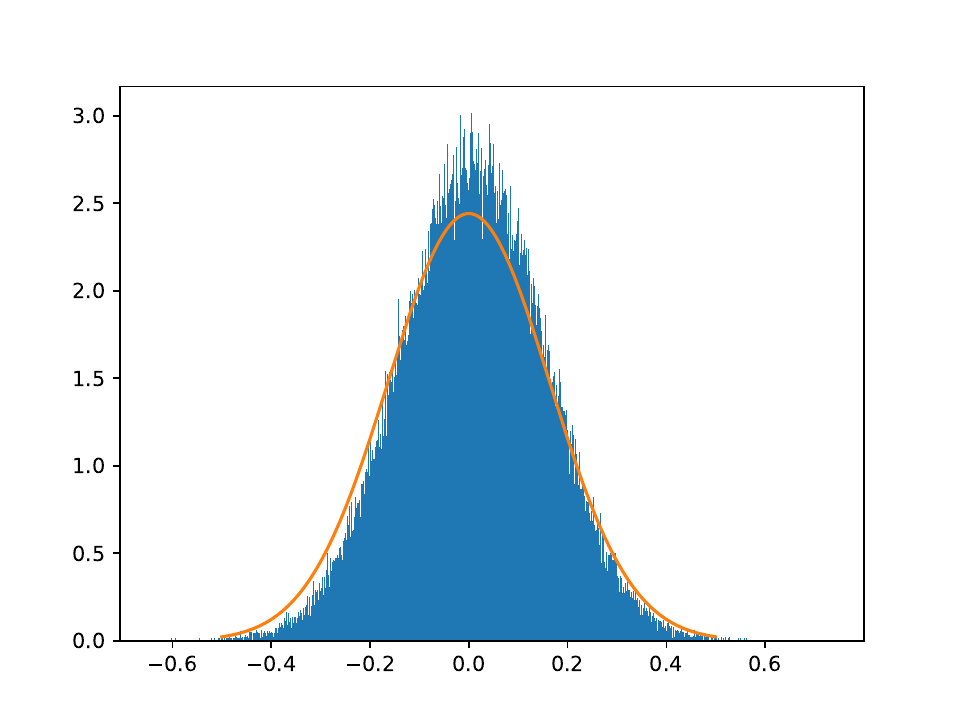}
    \caption{}
    \label{subfig:5densityitnet}
  \end{subfigure}
  \hfill
  \begin{subfigure}[T]{0.32\linewidth}
    \includegraphics[width=\textwidth]{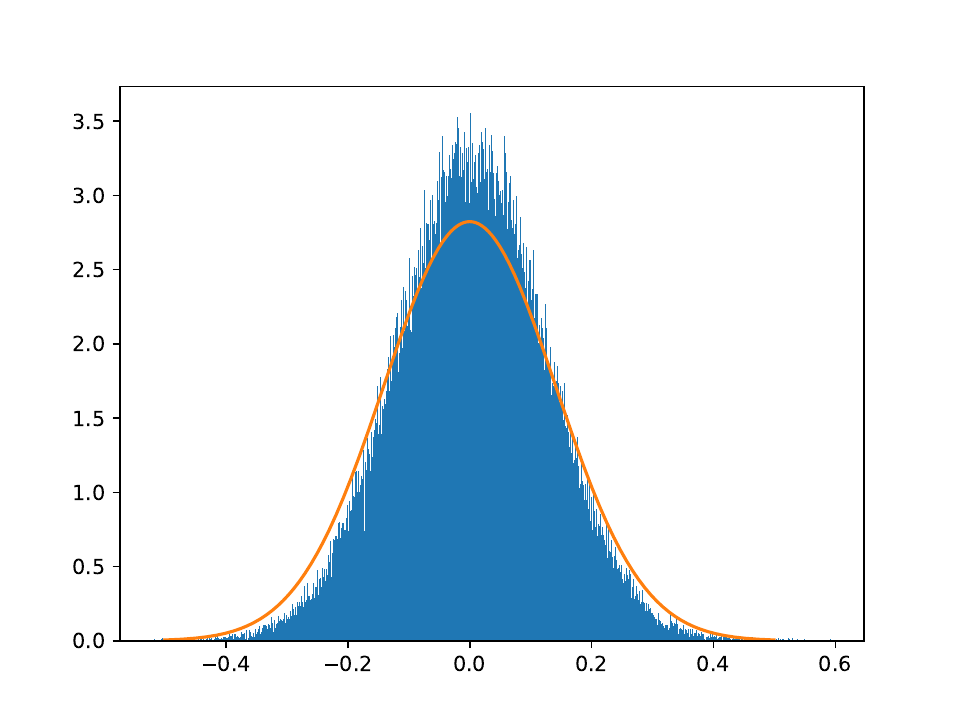}
    \caption{}
    \label{subfig:6densityitnet}
  \end{subfigure}
  \caption{Histograms showing the empirical distribution of the remainder component. The real part of $R$ is plotted as a histogram and the red line corresponds to a Gaussian fit. One realization of the remainder term is visualized for each of the experiments described in Table \ref{tab:results_itnet}. (\ref{subfig:1densityitnet} corresponds to the first column, \ref{subfig:2densityitnet} to the second column and so on.
  )}
  \label{fig:densities_itnet}
\end{figure}

In Figure \ref{fig:densities_unet} we present a series of histograms representing the empirical distribution of the remainder term's real part for the six different experimental settings, for the U-Net, conducted in this paper. Figure \ref{fig:densities_itnet} represents the histograms for the It-Net experiments. In most scenarios, the real part of the remainder term is Gaussian distributed with mean $0$. The only exceptions are Figures \ref{subfig:5densityunet} and \ref{subfig:6densityunet}, which correspond to the U-Net experiments with $30 \%$ undersampling. 

\end{document}